\documentclass[Afour,sageh,times]{sagej}

\usepackage{moreverb,url}
\usepackage[T1]{fontenc} % package to render some rare letters that appear in bib files
\usepackage{textcomp}

\usepackage{hyperref}
\hypersetup{colorlinks,bookmarksopen,bookmarksnumbered,citecolor=red,urlcolor=red}

\newcommand\BibTeX{{\rmfamily B\kern-.05em \textsc{i\kern-.025em b}\kern-.08em
T\kern-.1667em\lower.7ex\hbox{E}\kern-.125emX}}

\def\volumeyear{2026}

\usepackage{times}

\usepackage{enumitem}
\usepackage{multicol}

\usepackage{bm}
\usepackage{amsmath}
\usepackage{amssymb}
\usepackage{mathtools} % Provides \coloneqq
\usepackage{amsfonts}
\usepackage{comment}
\usepackage{graphicx}
\usepackage[dvipsnames]{xcolor}

\usepackage{gensymb}
\usepackage[linesnumbered,ruled,vlined]{algorithm2e}
\newcommand{\code}[1]{\textcolor{RoyalBlue}{$\rhd$ \emph{#1}}}

\let\cite\citep

\makeatletter
\renewcommand{\algocf@Vline}[1]{%
\strut\par\nointerlineskip
\algocf@push{\skiprule}%
\hbox{\vrule
  \vtop{\algocf@push{\skiptext}%
    \vtop{\algocf@addskiptotal #1}\Hlne}}%
\algocf@pop{\skiprule}%
\nointerlineskip % <— removed the \vskip\skiphlne
}%
\makeatother

\SetKwInput{Require}{Require}
\usepackage{booktabs}
\usepackage{makecell}  % for \Xhline
\usepackage{tabularx} % For controlling table width
\usepackage{cases}  % provides the numcases environment

\usepackage{pifont} % for cross symbol

\usepackage{multirow}

\usepackage{caption}
\usepackage{svg}  % for svg type image

\usepackage{amssymb}
\usepackage{booktabs} % for pandas-generated table
\usepackage{dblfloatfix}

\begin{document}

% \title{Beyond Behavior Cloning: Robustness through Interactive Imitation and Contrastive Learning}

% \title{From Point Labels to Action Sets: A Set-Based Framework for Interactive Policy Learning}

% \title{Reinterpreting Corrective Feedback as Action Sets for Interactive Policy Learning}

% \title{From Action Labels to Sets: Reformulating Action Supervision for Interactive Learning from Corrections}
\title{From Action Labels to Sets: Rethinking Action Supervision for Imitation Learning from Corrective Feedback}

%\title{From Action Labels to Sets: Reformulating Action Supervision through Corrective Imitation Learning}
% \title{From Action Labels to Sets: Reformulating Action Supervision for Learning from Corrective Feedback}

% \author{
%     \IEEEauthorblockN{Zhaoting Li, Rodrigo P{\'e}rez-Dattari, Robert Babuska, Cosimo Della Santina, Jens Kober} \\
%     \IEEEauthorblockA{Delft University of Technology, \{ z.li-23, r.j.perezdattari, r.babuska, c.dellasantina, j.kober \}@tudelft.nl}
%     \href{https://clic-webpage.github.io }{https://clic-webpage.github.io}
% }

\author{Zhaoting Li\affilnum{1}, Rodrigo P{\'e}rez-Dattari\affilnum{2}, Robert Babuska\affilnum{1}, Cosimo Della Santina\affilnum{1} and Jens Kober\affilnum{3}}

\affiliation{\affilnum{1}Delft University of Technology, The Netherlands \\
\affilnum{2}KTH, Sweden \\
\affilnum{3}University of Stuttgart, Germany}

\corrauth{Zhaoting Li, Delft University of Technology,
The Netherlands.}

\email{z.li-23@tudelft.nl}

% \begin{abstract} Behavior cloning (BC) learns policies by treating human demonstrations as pointwise action labels. This supervision can lead to overfitting under noisy data, particularly when expressive models are used (e.g., energy-based models). To address this, we propose a human-in-the-loop online learning alternative that reformulates supervision as set-valued action targets derived from corrective feedback. We introduce \textit{Contrastive policy Learning from Interactive Corrections (CLIC)}. CLIC leverages human corrections to estimate a set of desired actions and optimizes the policy to select actions from this set. Extensive simulation and real-robot experiments show that CLIC yields more stable learning under noisy or inconsistent feedback, supports complex multi-modal behaviors, and outperforms existing state-of-the-art methods across a range of feedback settings.
% Our implementation is publicly available at \href{https://github.com/clic-webpage/CLIC}{https://github.com/clic-webpage/CLIC}.
% \end{abstract}

\begin{abstract} Behavior cloning (BC) optimizes policies by treating human demonstrations as pointwise action labels. While effective with accurate action labels, this formulation is brittle in practice: when human-provided actions are imperfect, treating each label as an exact target can steer the policy away from the underlying desired behavior, particularly when expressive models are used (e.g., energy-based models). As a result, we propose a human-in-the-loop alternative that replaces pointwise supervision with set-valued action targets. We introduce \textit{Contrastive policy Learning from Interactive Corrections (CLIC)}. CLIC leverages human corrections to construct and refine sets of desired actions, and optimizes a policy to place probability mass over these sets rather than over a single action target. This formulation naturally accommodates both absolute and relative corrections and can represent complex multi-modal behaviors. Extensive simulation and real-robot experiments show that the proposed approach leads to effective policy learning across diverse settings: CLIC remains competitive with the state of the art under accurate data while being substantially more robust under noisy, relative, and partial feedback. Our implementation is publicly available at \href{https://clic-webpage.github.io/}{https://clic-webpage.github.io/}.
\end{abstract}

\keywords{
Interactive Imitation Learning, Human-in-the-Loop Learning, Learning from Demonstration, 
% Implicit Behavior Cloning, 
Corrective Feedback, Contrastive Learning
}

% \keywords{
% Interactive Imitation Learning, Corrective feedback, Contrastive Learning, Learning from Demonstration, Energy-based Models
% }
 \maketitle
% \maketitle
% \twocolumn[{%
% 	\renewcommand\twocolumn[1][]{#1}%
%         \maketitle
%         \vspace{-6mm}
% 	\begin{center}
% 		\includegraphics[width=0.99\textwidth]{figs/Fig_3_cover_Figure_3.pdf}

%  \captionof{figure}{\small
%  A: Our method operates in an Interactive Imitation Learning framework. Example rollouts of this framework are shown in Fig. \ref{fig:CLIC_overview}.
%  During each step, the robot's policy outputs a robot action $\bm a^r$. A human teacher provides corrective feedback occasionally if the robot action is suboptimal.
%  In (a1), each feedback stored in the data buffer $\mathcal{ D}$ defines a desired action set—a set of actions that includes the optimal action. 
%  These individual spaces are then aggregated to form an overall desired action set $\hat{\mathcal{A}}^{\mathcal{D}}$, which refines the estimate of optimal actions.
%  In (a2), the policy, modeled as an energy-based model (EBM), is trained to generate actions within $\hat{\mathcal{A}}^{\mathcal{D}}$.
% B: Examples of the learned EBMs in a 2D action space. Implicit BC \cite{2022_implicit_BC} overfits each action label, while our method estimates the optimal action without overfitting.
% \label{fig:framework}}
% 	\end{center}
% }]

\begin{figure*}
    \centering
   	\includegraphics[width=0.99\textwidth]{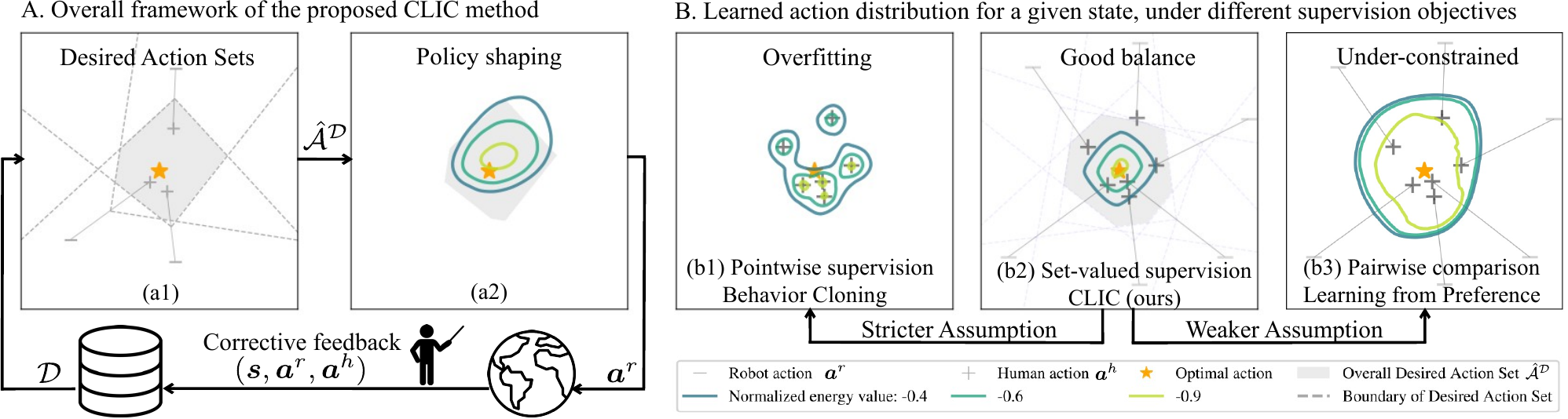}

 \caption{\small
\textbf{A:} Our method operates in an Interactive Imitation Learning framework. Example rollouts of this framework are shown in Fig. \ref{fig:CLIC_overview}.
At each step, the robot's policy outputs an action $\bm a^r$, and a human teacher provides corrective feedback occasionally if the robot action is suboptimal.
In (a1), each feedback stored in the data buffer $\mathcal{ D}$ defines a desired action set---actions consistent with the feedback given prior knowledge of its type.
These sets are then aggregated to form an overall desired action set $\hat{\mathcal{A}}^{\mathcal{D}}$.
In (a2), the policy, represented as an energy-based model (EBM), is trained to generate actions within $\hat{\mathcal{A}}^{\mathcal{D}}$.
\textbf{B:} Examples of a learned action distribution in a 2D action space.  Pointwise BC \cite{2022_implicit_BC} can overfit individual action labels, while pairwise comparison may provide supervision that is too weak to sufficiently constrain the action space. CLIC uses set-valued supervision, which offers a balance between pointwise targets and pairwise comparisons, recovering the underlying optimal action without overfitting to individual labels or under-constraining the policy.
\label{fig:framework}}
\end{figure*}

% \begin{keywords}
%     Interactive Imitation Learning, Corrective feedback, Contrastive Learning, Learning from Demonstration, Energy-based Models
% \end{keywords}

\section{Introduction}
\label{sec:introduction}

Behavior Cloning (BC) enables robots to acquire complex skills by imitating human demonstrations \cite{2024_IJRR_survery_feedback_types, 2018_review_IL, 2020_Survey_LfD, 2023_Survey_LfD}. 
It casts policy learning as supervised learning by treating the demonstrated action at each state as a pointwise training target.
This formulation also naturally extends to interactive settings under \emph{absolute corrective feedback}, where a human intervenes by taking control of the robot in states where the learning policy performs poorly, and uses these corrective actions as training targets to iteratively refine the policy~\cite{2022_implicit_BC, 2019_HG_DAgger, 2022_IIL_survey}. 
In practice, however, human-provided actions---whether collected offline or through interactive corrections---can be suboptimal or noisy. 
This is problematic because expressive policy classes, derived from, e.g., energy-based models or score functions, are prone to overfitting such suboptimal data rather than recovering the underlying optimal behavior \cite{2017_DNN_noise_memorization, 2017_NIPS_understanding_noise_generalization}. 
Yet, most existing methods still rely on pointwise BC objectives to learn from such data \cite{2023_IIFL_implicit_interactive_BC, 2022_implicit_BC, 2019_HG_DAgger, 2011_DAgger}. Fig. \ref{fig:framework}b1 shows an energy-based model overfitting to noisy action labels, forming sharp local minima around each training target. 
%With expressive policy classes, this encourages the model to fit individual corrections rather than recover the underlying optimal behavior \cite{2017_DNN_noise_memorization, 2017_NIPS_understanding_noise_generalization}. 

% We focus on \emph{corrective feedback}, which can be either \emph{absolute} or \emph{relative} \cite{2019_Carlos_IJRR, 2022_IIL_survey, 2024_IJRR_survery_feedback_types}. Absolute feedback treats action labels in the same way as pointwise BC \cite{2022_implicit_BC}, with the only difference that data is collected and trained on interactively \cite{2019_HG_DAgger}. In contrast, 

% This limitation is more pronounced when feedback is relative rather than absolute \cite{2019_Carlos_IJRR, 2022_IIL_survey, 2024_IJRR_survery_feedback_types}. 
%A more challenging form of corrective feedback is \emph{relative correction}, which indicates the direction of the optimal action relative to the robot's current action, without specifying the magnitude \cite{2019_Carlos_IJRR, 2018_D_COACH}. Such methods use this feedback to construct surrogate action targets and update the policy with a pointwise BC objective. This, however, introduces an additional challenge: as the policy improves, past \emph{on-policy} corrections become suboptimal or contradictory, destabilizing training \cite{2018_D_COACH}. 
%Taken together, these challenges suggest that the pointwise BC loss relies on a strict assumption that feedback provides an exact and reliable action target.  When this assumption is violated by noisy or relative feedback, the policy becomes prone to overfitting and instability.

In contrast, work focusing on learning from \emph{preference feedback} relaxes the pointwise BC assumption by replacing exact action labels with pairwise comparisons between actions or trajectories \cite{2017_survey_preference}.
Pairwise ranking allows the learner to reason about relative quality and can enable policies to improve over suboptimal demonstrations \cite{2019_IRL_ranked_demonstration}.
Early approaches learn a reward model from preferences and then optimize a policy via reinforcement learning, forming a two-stage pipeline that increases modeling complexity and optimization instability \cite{2023_Contrastive_Prefence_Learning, 2023_DPO}.
Recent methods directly update policies from pairwise comparisons \cite{2023direct_preference_policy_optimization, 2025_PPL_intervention_preference,  2023_Contrastive_Prefence_Learning, 2024_Step_DPO, 2023_slic}.
% However, such pairwise ranking losses enforce only that one action is preferred over another, without directly constraining the probability density of other actions.
% In high-dimensional continuous action spaces, especially with expressive policies, pairwise comparison can under-constrain the policy and allow the policy to output suboptimal actions  \cite{ chen_deformpam_2025, 2017_survey_preference}.
However, such pairwise ranking losses constrain only the relative ordering between two actions, leaving the rest of the action space largely unconstrained. As a result, a policy can satisfy the same preference while still assigning substantial probability mass to suboptimal actions, especially in high-dimensional continuous spaces \cite{2017_survey_preference}. This issue is again particularly pronounced for expressive policy classes, which can represent many such weakly constrained solutions \cite{chen_deformpam_2025}.
 % However, such pairwise ranking losses constrain only the relative ordering between two actions: they enforce that one action should be preferred over another, but leave the rest of the action space largely unconstrained.
 % Consequently, many action distributions can satisfy the same pairwise preference while still assigning substantial probability mass to suboptimal actions, especially in high-dimensional continuous spaces \cite{2017_survey_preference}.
 % This limitation is particularly problematic for expressive policy classes, since they can represent many such weakly constrained solutions \cite{ chen_deformpam_2025}.
This limitation is illustrated in Fig.~\ref{fig:framework}b3, where a policy satisfies the pairwise comparison constraint while still assigning substantial probability mass to suboptimal regions.
This leaves a gap for a supervision objective that is more robust than pointwise BC and more informative than pairwise preference learning.

In this work, we propose a set-based formulation that bridges these two extremes.
Rather than treating human corrections as pointwise targets or pairwise comparisons, we interpret them as specifying a \emph{desired action set}.
That is, for each correction at a given state, we define a subset of the action space that includes actions consistent with the feedback while excluding inconsistent ones.
We then design a training objective that encourages the policy to place probability mass on this desired action set. 
A single correction, however, is generally insufficient for the policy to recover the optimal action at a given state, since the corresponding desired action set may still contain suboptimal actions. This motivates the interactive nature of our framework, allowing such sets to be iteratively refined as more feedback is provided. We therefore propose \emph{Contrastive policy Learning from Interactive Corrections} (CLIC) to iteratively refine the policy by aggregating new corrections and updating the policy to align with the resulting desired action sets. 
As illustrated in Fig.~\ref{fig:framework}b2, the policy is progressively shaped towards action sets rather than directly matching individual action labels.

In addition to absolute corrections (i.e., interventions), we study CLIC in the context of \emph{relative corrections}. This form of feedback indicates the direction of the optimal action relative to the policy's current action, without specifying its magnitude, and has been shown to require less cognitive load than demonstrations while achieving competitive results \cite{2019_Carlos_IJRR, 2022_IIL_survey, 2018_D_COACH}. Typically, these methods use this corrective feedback to construct surrogate action targets. Then, the policy is updated via a pointwise BC objective toward actions that improve upon the current policy, but are not necessarily optimal. This introduces an additional challenge: as the policy improves, past \emph{on-policy} corrections become suboptimal or even contradictory, potentially destabilizing training \cite{2018_D_COACH}. 
Crucially, our framework addresses this issue by treating each relative correction as specifying a desired action set rather than an exact action target. This desired action set constrains a region of actions consistent with the indicated direction, without committing to a specific magnitude.

The main contributions of this work are as follows. First, we propose a set-valued supervision perspective for corrective feedback, where supervision is represented as desired action sets that bridge the gap between pointwise behavior cloning and pairwise preference learning. Second, we derive a practical policy-shaping framework, CLIC, that constructs desired action sets from feedback and trains policies to align their probability mass with these sets. Most of our study is carried out for implicit policy representations. However, for completeness, we further analyze CLIC in the context of explicit policy representations. Third, we provide extensive evaluations in simulation and on real robots showing that the proposed formulation improves robustness to noisy feedback, captures multi-modal behaviors, and unifies learning from both absolute and relative corrections within a single framework.

\section{Related Work}
\label{sec:related_work}
We review prior work on policy learning from different forms of human feedback, focusing on demonstrations, preferences, and relative corrections. Within each category, we discuss how each feedback modality is translated into a supervision objective for policy learning.

%  The energy-based model can be trained in many different approaches, depending on the assumption of the type of data (demonstration, correction, reward, etc).
% Besides, other types of models, such as diffusion, have been applied to policy representation, which outperforms its EBM counterpart (IBC). 
% In this section, we briefly overview the related works and the IBC's training instability issue. At the end, we draw connection between IBC and our method, which utlize the corrective feedback for EBM traning. 

% new sections: learning under pointwise supervision
% learning under pairwise supervision

% Learning under pointwise supervision from Demonstrations/Absolute corrections
% pointwise label-> explicit policy -> implicit policy -> noisy data 

% Learning under pairwise ranking (however, should mention in trjectory level or state-action level..)-> preference over trajectory -> perference over actions -> noisy demo

% Learning from corrections or interventions (can be pointwise or pairwise)

\subsection{Learning from Demonstration and Absolute Corrections}
Learning from demonstration aims to teach robot behavior models from demonstration data, which provides the robot with examples of desired actions. 
Behavior cloning (BC) formulates this problem as supervised learning by treating each demonstrated action as a pointwise action label \cite{2018_review_IL, 2023_Survey_LfD}.
Classical approaches typically rely on \textit{explicit policies}, which directly predict an action or parameterize a simple action distribution for each state. While effective in many settings, such policies often struggle to represent complex action distributions, particularly when multiple actions are optimal for the same state, as commonly arises in manipulation tasks with multi-modal solutions \cite{2023_diffusionpolicy}.
To address this limitation, recent work has explored more expressive policy classes, such as \textit{implicit policies} parametrized by deep generative models, including EBMs \cite{2019_EBM_Du_Yilun, 2021_how_to_train_EBM},  diffusion models \cite{2020_diffusion, 2015_diffusion}, which have been introduced to better capture such multi-modal data distributions \cite{2024_survey_deep_generative_model_in_robotics}. 

Among these approaches, Implicit Behavior Cloning (IBC) learns an EBM over the state-action space using a contrastive objective defined on demonstrated actions and sampled counterexamples \cite{2022_implicit_BC}.  
However, training EBMs depends critically on the sampled counterexamples used in the contrastive objective, which can affect training stability \cite{2020_flow_constrastive_estimation_EBM, 2020_hard_negative_mixing_contrastive, 2021_how_to_train_EBM, 2022_arxiv_IBC_gaps}.
Diffusion models, which learn to denoise noise-corrupted data \cite{2020_diffusion, 2015_diffusion}, have also been utilized to represent robot policies, resulting in diffusion policies \cite{2023_diffusionpolicy, 2023_score_diffusion_policy}. These diffusion policies effectively learn the gradient (score) of the EBM \cite{2020_Score_based_diffusion} and offer improved training stability \cite{2023_diffusionpolicy}.
Such implicit policies have achieved strong performance on long-horizon and multi-modal tasks \cite{2024_EBM_planning_air_hockey_application, 2024_survey_deep_generative_model_in_robotics, 2024_IBC_RL_planning}, but like other offline imitation learning methods, they remain vulnerable to covariate shift when the learned policy visits states not covered by the demonstration data \cite{2011_DAgger}. A common way to address this issue is to move from offline demonstrations to interactive imitation learning, where the teacher provides \textit{interventions} or \textit{absolute corrections} on states visited by the learner \cite{2022_IIL_survey, 2020_RSS_expert_interventio_learning}. Following this direction, recent work has extended implicit policies to online IIL settings \cite{2023_IIFL_implicit_interactive_BC, 2024_Diffusion_dagger}.
% These implicit policies have achieved strong performance on long-horizon and multi-modal tasks \cite{2024_EBM_planning_air_hockey_application, 2024_survey_deep_generative_model_in_robotics, 2024_IBC_RL_planning}, but still suffer from distributional shift \cite{2011_DAgger, 2020_RSS_expert_interventio_learning}.  To address this issue, these implicit policies have been threfore extended into an online interactive imitation learning (IIL) framework \cite{2023_IIFL_implicit_interactive_BC, 2024_Diffusion_dagger}.  

% However, as mentioned in the introduction section, the powerful encoding capability of these models can also cause overfitting behavior with the pointwise BC objective, especially when the label deviates from the optimal action. 
However, these methods still rely on a pointwise BC objective that treats each demonstrated or corrected action as an exact target. As discussed in the introduction, this is problematic when the provided action deviates from the optimal one, particularly for implicit policies. In contrast, our method retains the expressive benefits of implicit policies while extending pointwise supervision to set-valued supervision derived from corrective feedback.

% Despite these advances, most existing methods still rely on a pointwise behavior cloning objective that treats each demonstrated action as an exact target. This can be restrictive when demonstrations or corrections are imperfect, as the pointwise supervision does not represent uncertainty around the  action label.

% \cite{2024_IBC_RL_planning}

\subsection{Learning from Preference Feedback}
Preference-based feedback typically involves comparing different robot trajectory segments.
From such comparisons, a reward or objective model is usually learned and then used to optimize a policy \cite{2017_DRL_preference, 2021_Pebble_preference_RL, 2015_IJRR_Preference_traj_ranking, 2020_openai_RLHF, 2024_comparative_language}.
More recent approaches improve efficiency by learning policies directly from pairwise comparisons, either through contrastive objectives \cite{2023direct_preference_policy_optimization, 2023_Contrastive_Prefence_Learning,  zhao2022calibrating, 2023_slic} or direct preference optimization \cite{2023_DPO}.
Preference-based formulations have also been used to extract useful supervision from suboptimal demonstrations \cite{2019_IRL_ranked_demonstration}.

% Trajectory-level preference feedback is often less data-efficient for robotics because it provides only indirect supervision for state-action learning \cite{2022_IIL_survey}.
Similar pairwise objectives have also been applied directly in the action space. For example, 
\citet{2025_PPL_intervention_preference} use a pairwise comparison objective to train an explicit policy.
Proxy Value Propagation (PVP) \cite{2023_NIPS_PVP} learns an energy-based policy by assigning lower energy to human actions than to recorded robot actions.
% Compared with the pointwise BC objective, pairwise supervision relaxes the need for exact action targets, but it only constrains which action should be preferred. 
Compared with the pointwise BC objective, pairwise supervision relaxes the requirement for exact action targets, requiring only that one action be preferred over another.
However, such pairwise comparisons leave the rest of the continuous action space largely unconstrained, since many action distributions can satisfy the same pairwise ordering.
Our method instead uses corrective feedback to define desired action sets, providing a richer supervision signal for policy learning in continuous action spaces.

% While there are works to make it data efficient via active learning \cite{2024batch_acitve_learning_Preference}  or utilizing prior knowledge of state \cite{2024hindsight_preference}, our paper focuses on human feedback in the state-action space.

\subsection{Learning from Relative Corrections}
% Relative corrective feedback provides incremental information on how to improve an action, balancing information richness and simplicity for the teacher \cite{2022_IIL_survey}. 
% % The data consists of pre-correction and post-correction actions.
% This correction feedback can be transferred into preference data with trajectory pairs and
% the objective function can be learned from the preference data, as in \cite{2015_IJRR_Preference_traj_ranking, 2017_pHRI_correction_learning_objective, 2018_uncertainty_correction_objective}.
% Alternatively, \cite{2022_TRO_correction_objective_function} proposed directly inferring the objective function without preference transformation.
% However, these objective functions are linear combinations of features, which may struggle with complex tasks.
% Another line of work is the COACH-based framework (Corrective Advice Communicated by Humans), which directly learns a policy from relative corrections
% \cite{2018_D_COACH, 2019_Carlos_COACH, 2021_BDCOACH}.
% This framework has been extended to 
%  utilize the feedback from the state space instead of the action space 
% \cite{2020_COACH_state_Space} and combined with reinforcement learning to increase the RL efficiency\cite{2019_Carlos_IJRR}.
% However, as we mentioned in Section~\ref{sec:introduction}, the previous COACH-based framework fails to utilize the history data, making it inefficient compared with demonstration-based methods. 
% Instead, our CLIC method can utilize the history data as it will not harm the policy under our assumption of desired action set. 

Relative corrections provide incremental information on how to improve an action, balancing information richness and simplicity for the teacher \cite{2022_IIL_survey}. 
% The data consists of pre-correction and post-correction actions.
This correction feedback can be transferred into preference data with trajectory pairs and
objective functions can be learned from the preference data, as in \citet{ 2017_pHRI_correction_learning_objective, 2018_uncertainty_correction_objective, 2015_IJRR_Preference_traj_ranking}.
Alternatively, \citet{2022_TRO_correction_objective_function} proposed directly inferring objective functions without preference transformation.
However, these objective functions are linear combinations of features, which may struggle with complex robotic tasks.

Another line of work is the COACH-based framework (Corrective Advice Communicated by Humans), which directly learns a policy from relative corrections
\cite{2019_Carlos_COACH, 2021_BDCOACH, 2018_D_COACH}.
This framework has been extended to 
 utilize feedback from the state space instead of the action space 
\cite{2020_COACH_state_Space} and combined with reinforcement learning to increase the RL efficiency \cite{2019_Carlos_IJRR}.
However, COACH-based methods rely on the over-optimistic assumption that the action labels derived from relative corrections are optimal, allowing the policy to be optimized via the pointwise BC objective \cite{2019_Carlos_IJRR, 2019_Carlos_COACH, 2019_Rodrigo_D_COACH}. 
This assumption becomes a critical limitation when feedback is aggregated into a replay buffer.  
As the robot's policy continuously improves, previous feedback may no longer be valid, causing incorrect policy updates \cite{2021_BDCOACH}. 
As a result, this limits the usefulness of replay buffers and often requires keeping only recent feedback. 
In contrast, our method does not convert relative corrections into exact action targets. Instead, it interprets each correction as inducing a desired action set and aggregates these sets across interactions. This allows earlier corrections to remain useful even when they no longer define an optimal point target.

\section{Preliminaries}
\label{sec:Preliminaries}

\subsection{Interactive Imitation Learning Formulation}
\label{sec:Preliminaries:IIL}
We consider a typical Interactive Imitation Learning (IIL) problem, where a human instructor, known as the \emph{teacher}, aims to improve the behavior of the learning agent, referred to as the \emph{learner}, by providing feedback on the learner's actions \cite{2022_IIL_survey}.  
The following sections formalize this problem by detailing the framework, the teacher's feedback, the policy representation, and the learning objective.
\subsubsection{Markov Decision Process in IIL}
In IIL, a Markov Decision Process (MDP) is used to model the decision-making of the learner taught by the teacher. 
The MDP considered here is a 4-tuple \((\mathcal  S, \mathcal A, T, H)\), where \(\mathcal  S\) represents the set of all possible states in the environment, \(\mathcal  A\) denotes the set of actions the agent can take, \(T(\bm s' |\bm s, \bm a)\) is the transition probability, 
and \( \bm h = {H}(\bm s, \bm a)\) denotes the teacher feedback.
The human teacher has the ability to assess whether the learner's current action $\bm a$ is suboptimal, denoted by the function $G(\bm s, \bm a) \in \{0, 1\}$; 
the teacher can provide the 
  feedback signal $\bm h$
if $G(\bm s, \bm a) = 1$.
Concretely, we define the set of optimal actions as $\mathcal{A}^*_{\bm s} =\{\bm a \in \mathcal{A} | G(\bm s, \bm a) = 0\}$.
We denote optimal actions as $\bm a^* \in \mathcal{A}^*_{\bm s}$.
A task is \emph{unimodal} if 
$\mathcal{A}^*_{\bm s}$ 
 is connected (i.e., has exactly one connected component), and \emph{multi-modal} if 
$\mathcal{A}^*_{\bm s}$ 
 has at least two disjoint connected components.
\subsubsection{Teacher Corrective Feedback}
\label{sec:sub:sub:teacher_feedback}
The feedback $\bm h$ can be defined according to the feedback type.
In demonstration or intervention feedback, $\bm h$ represents the action the learner should execute at a given state. In relative corrective feedback, 
 $\bm h$ is a normalized vector indicating the direction in which
the learner's action should be modified, i.e., 
${\bm h \in \mathcal{H} = \{ \bm d \in \mathcal{A} \mid  ||\bm d|| = 1\}}$.

Let \(\bm a^r\) denote the robot action executed at state \(\bm s\), and let \(\bm a^h\) denote the action implied by the human's feedback. 
For absolute corrections, we have that $\bm a^h = \bm h$. 
For relative corrections, we have that $\bm a^h = \bm a^r + e \bm h$, where the magnitude hyperparameter $e$ is set to a small value.
For notational convenience, we represent each corrective feedback instance at state $\bm s$ by the action pair \((\bm a^r, \bm a^h)\). This action pair is used throughout the paper in place of the raw feedback signal \(\bm h\).
% Accordingly, we define the \textit{observed action pair} $(\bm a^r, \bm a^h)$, where $\bm a^r$ denotes the robot action and $\bm a^h$ denotes the human feedback action, referred to as human action for simplicity. 
% For absolute correction, we have that $\bm a^h = \bm h$. 
% In contrast, for relative correction, we have that $\bm a^h = \bm a^r + e \bm h$, where the magnitude hyperparameter $e$ is set to a small value.

\subsubsection{Policy}
\label{sec:pre:policy}
A policy in an MDP defines the agent's behavior at a given state, denoted by \(\pi\).
It specifies the action distribution conditioned on state \(\bm s\), with \(\pi: \mathcal S \times \mathcal A \rightarrow [0, 1]\). In this work, we parameterize the policy $\pi_\theta$ with a deep neural network (DNN) and $\bm \theta$ denotes the DNN's parameter vector.
Following implicit behavior cloning (IBC) \cite{2022_implicit_BC}, 
we model the policy through an energy-based model (EBM) \( E_\theta(s, a) \) that takes state \( s \) and action \( a \) as inputs and outputs a scalar energy value, leading to
\begin{equation}
    \pi_{\theta}(\bm a | \bm s) = \frac{1}{Z}e^{-E_{\theta}(\bm s, \bm a)},
    \label{eq:estiamtion_policy_EBM}
\end{equation}
% where $Z$ is a normalizing constant that can be approximated by a set of action samples $ \{ {\bm a}_j  \}_{j=1}^{N_{a}}$: $Z = \sum_{j=1}^{N_{\text{a}}} e^{-E_{\theta}(\bm s, \bm a_j)}  $.
% $N_a$ denotes the number of sampled actions.
% The $N_{\text{a}}$ samples $\{\bm a_j\}$ are obtained via Langevin MCMC sampling (see Eq.~\eqref{eq:mcmc_sampling}).
where $Z$ is a normalizing constant. In practice, it is approximated using a finite set of $N_a$ action samples $\{\bm a_j\}_{j=1}^{N_a}$, i.e., $Z \approx \sum_{j=1}^{N_a} e^{-E_{\theta}(\bm s, \bm a_j)}$. Here, $N_a$ denotes the number of sampled actions, and the samples are obtained via Langevin MCMC sampling (see Eq.~\eqref{eq:mcmc_sampling} in Appendix \ref{Appendix:IBC}).

For unimodal tasks, we additionally consider an explicit Gaussian policy with fixed covariance \(\bm \Sigma\) and mean \(\bm \mu_{\theta}(\bm s)\),
\begin{equation}
    \pi_{\theta}(\bm a | \bm s) = \mathcal{N}\!\left(\bm \mu_{\theta}(\bm s), \bm \Sigma\right).
\end{equation}

\subsubsection{Objective}
\label{sec:sub:objective}
In IIL, an observable surrogate loss \(\ell_{\pi}(\bm{s})\) is typically employed. This loss measures the alignment of the learner's policy \(\pi\) with the teacher's feedback.
The learner's policy $\pi^{l*}$ is then optimized by solving the following equation:
\begin{equation}
    \pi^{l*} = \underset{\pi\in\Pi}{\arg\min} \mathbb{E}_{\bm s\sim d_{\pi}(\bm s)} \left[ \ell_{\pi}( \bm s) \right],
    \label{eq:IIL_formulation}
\end{equation}
where $d_{\pi}(\bm s)$ is the state distribution induced by the policy $\pi$.
In practice, the expected value of the surrogate loss in Eq.~\eqref{eq:IIL_formulation} is approximated using the data collected by a policy that interacts with the environment and the teacher.
% The resulting data buffer is $\mathcal{D} = \{[\bm s_t, \bm a^r_t, \bm a^h_t], t = 1, \dots \}$. 

\subsection{Pointwise Behavior Cloning Objective}
\label{sec:pre:bc}
To optimize the learner's policy in Eq.~\eqref{eq:IIL_formulation}, a common approach is to use the pointwise behavior cloning (BC) objective, which increases the likelihood of the teacher action $\bm a^h$ at state $\bm s$:
\begin{equation}
    \ell_{\pi_\theta}(\bm{s}) = - \log \pi_{\bm \theta} (\bm a^h | \bm s).
    \label{eq:BC_Loss}
\end{equation}
This objective treats the teacher action as an exact target for the given state. Such pointwise supervision can be overly strict when the teacher action is noisy, suboptimal, or derived from a relative correction, since the policy is then encouraged to match an action that may not be truly optimal.

Implicit behavior cloning \cite{2022_implicit_BC} adopts this objective while representing the policy with an EBM. Under this parameterization, Eq.~\eqref{eq:BC_Loss} can be reformulated as the InfoNCE loss \cite{2018_InfoNCE_representation_learning, 2024_revisting_IBC}:
\begin{align}
\label{eq:ibc_info_NCE}
\ell_{\text{InfoNCE}}(\bm s, \bm a^h \!\!,\mathbb{A}^{neg})
\!\! = \!\! - \! \log \! \left[ \! \frac{e^{-E_\theta(\bm s, \bm a^h)}}{e^{-\!E_\theta(\bm s, \bm a^h  \!)} \!\! + \!\!\sum_{j=1}^{N_{a}}  \!e^{-\!E_\theta(\bm s, {\bm a}_j \!)}} \!\right] \!\!,\!\!
\end{align}
where  $ \mathbb{A}^{neg} = \{ {\bm a}_j  \}_{j=1}^{N_{a}}$ is treated as the set of \emph{negative samples} in the contrastive formulation of the pointwise BC objective.
These samples are drawn from the current EBM policy, as described in Section \ref{sec:pre:policy}. 
% The policy is encouraged to assign lower likelihood to these negative samples.
The resulting loss lowers the relative likelihood of these negative samples with respect to the teacher action.
This implies that actions that are in fact valid may also be treated as negative examples and consequently overly penalized.

These limitations of pointwise supervision motivate moving toward a more flexible supervision objective.

% One core assumption of IBC is that the teacher action is optimal and all other actions are not \cite{2022_implicit_BC}.
% However, actions considered as negative may still be valid and should not be overly penalized.
% This makes selecting appropriate negative samples challenging and introduces instability during the IBC's training process.
% Besides, the surrogate loss in Eq.~\eqref{eq:BC_Loss} also limits IBC being applied when $\bm a^h$ is noisy or comes from relative corrections.
% To address these issues, we must therefore develop a new surrogate loss to align the learner's policy with various teacher feedback and avoid overfitting.

% \begin{figure}[b!]
% 	\centering
% 	\includegraphics[width=0.485\textwidth]{figs/overview_CLIC-crop.pdf}
% 	\caption{
% Overview of CLIC. This figure illustrates three iterations of the CLIC framework in Fig.~\ref{fig:framework}A.
% The gray-shaded area represents desired action sets induced by corrective feedback, while the contour map shows the energy landscape of the EBM policy.
% Taking the second iteration as an example: the teacher provides corrections (c1) on the robot's action sampled from its policy (p1), resulting in a new desired action set and refining the overall desired action set (a2).
% The robot policy is then updated in (p2) to place more probability mass within this refined set.
% As iterations progress, the overall desired action set is gradually refined with new corrective feedback, and the robot policy is updated accordingly.
%  }
% \label{fig:CLIC_overview}
% \end{figure}

\section{Method Overview}
% \subsection{Overview of Our Method}

We propose Contrastive policy Learning from Interactive Corrections (CLIC) to address the limitations of pointwise behavior cloning objectives described in Section \ref{sec:pre:bc}. 
To this end, we build on the IIL formulation in Section~\ref{sec:Preliminaries:IIL} and design a surrogate loss $\ell_{\pi}$ that enables effective policy learning from corrective feedback.

\begin{figure}[t!]
	\centering
	\includegraphics[width=0.485\textwidth]{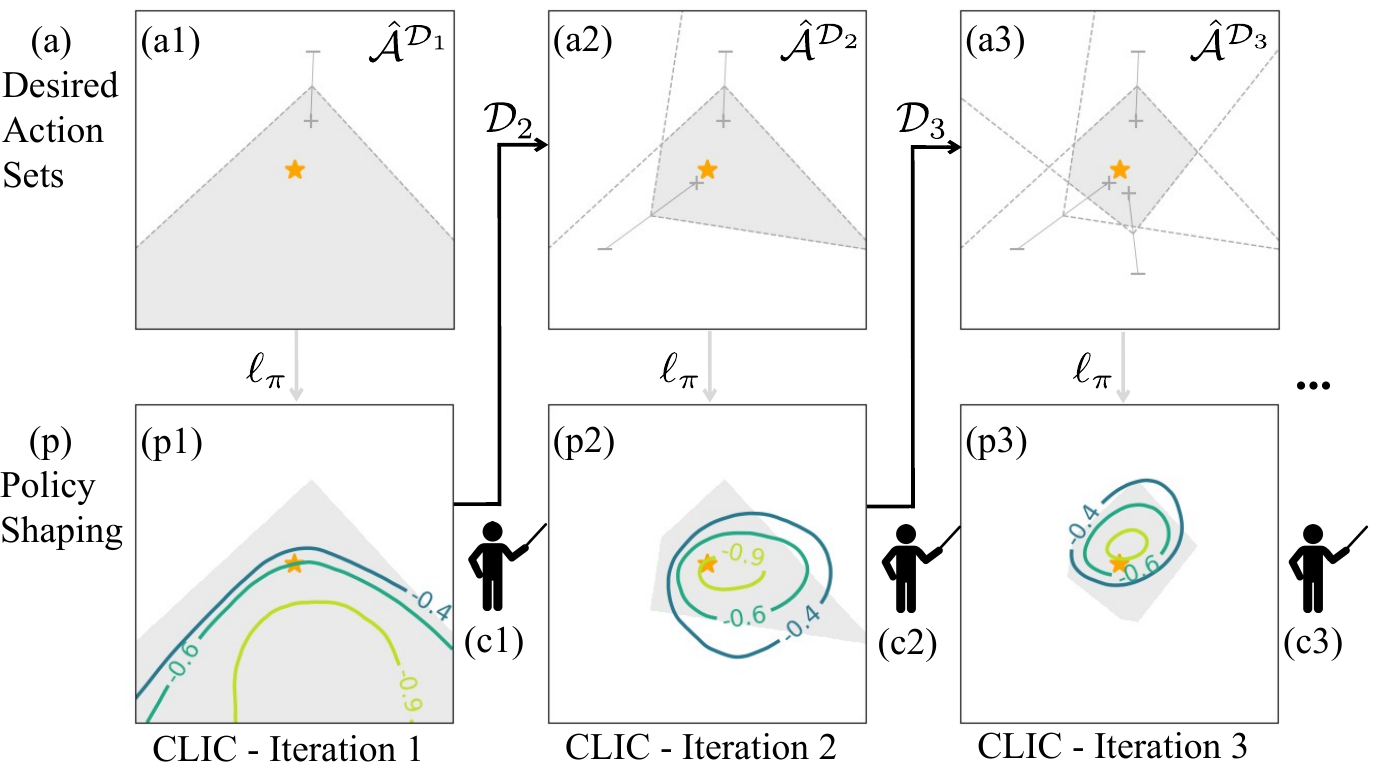}
	\caption{
Overview of CLIC. This figure illustrates three iterations of the CLIC framework in Fig.~\ref{fig:framework}A.
The gray-shaded area represents desired action sets induced by corrective feedback, while the contour map shows the energy landscape of the EBM policy.
Taking the second iteration as an example: the teacher provides corrections (c1) on the robot's action sampled from its policy (p1), resulting in a new desired action set and refining the overall desired action set (a2).
The robot policy is then updated in (p2) to place more probability mass within this refined set.
As iterations progress, the overall desired action set is gradually refined with new corrective feedback, and the robot policy is updated accordingly.
 }
\label{fig:CLIC_overview}
\end{figure}

The key idea of CLIC is to replace pointwise action targets with set-valued supervision. For each corrective feedback, CLIC first constructs a desired action set. This set, rather than a single action label $\bm a^h$, represents a region of desired actions that are consistent with this corrective feedback. We then use this set to define a surrogate loss that incentivizes the learner’s policy toward this space. As illustrated in Fig.~\ref{fig:CLIC_overview}, this process is iterative. The aggregation of multiple corrections for a given state leads to a progressively refined overall desired action set, and the policy adapts accordingly. Section \ref{sec:Desired_action_space} describes the construction and aggregation of the desired action sets, while Section \ref{sec:Policy_shaping} introduces the surrogate loss used to shape the policy under this set-valued supervision.

\section{Desired action sets: Formulation and Aggregation}
\label{sec:Desired_action_space}

This section formalizes desired action sets constructed from corrective feedback and shows how their aggregation converges to optimal actions.
We first outline their common properties (Section \ref{sec:sub:general_def_desiredA}). 
Next, we describe two specific types of desired action sets: polytopes (Section \ref{sec:sub:desired_action_space_relative}) and balls (Section \ref{sec:sub:desired_action_space_absolute}). 
Finally, in Section \ref{sec:overall_desiredA_space}, we introduce the CLIC algorithm for aggregating multiple corrections and illustrate convergence conditions for the overall desired action set.

\subsection{General Definition of a Desired Action Set}
\label{sec:sub:general_def_desiredA}
We view corrective feedback  as providing set-valued information about the action space rather than as specifying a pointwise action label.
Specifically, the feedback rules out the robot’s executed action and provides information about the set of actions that would be consistent with the teacher’s action. Accordingly, for a corrective feedback  \((\bm{a}^r, \bm{a}^h)\) at state $\bm s$, we define a \textit{desired action set} \(\hat{\mathcal{A}}(\bm{a}^r, \bm{a}^h)\) such that the following properties hold:
\begin{enumerate}
    \item {Consistency with the correction:}
    \[
    \bm{a}^h \in \hat{\mathcal{A}}(\bm{a}^r, \bm{a}^h) \quad \text{and} \quad \bm{a}^r \notin \hat{\mathcal{A}}(\bm{a}^r, \bm{a}^h).
    \]
   \item {Set-valued supervision:}
Let $\mathcal{A}^*_{\bm s}$ denote the optimal-action set at $\bm s$. We assume
\[
\hat{\mathcal{A}}(\bm a^r,\bm a^h)\cap \mathcal{A}^*_{\bm s} \neq \varnothing,
\]
and allow $\hat{\mathcal{A}}(\bm a^r,\bm a^h)$ to contain actions outside $\mathcal{A}^*_{\bm s}$.
\end{enumerate}
\vspace{-2mm}
This formulation makes a weaker assumption than point-label BC about how specifically feedback identifies the desired action: instead of requiring the correction to identify a unique optimal action, it only requires the induced set to remain compatible with at least one optimal action. Meanwhile, it is stronger than the pairwise comparison objective, because it constrains a region of the action space rather than merely preferring one action over another. Standard BC objective appears as a special case when the desired action set collapses to a single point, \(\hat{\mathcal A}(\bm a^r, \bm a^h)=\{\bm a^h\}\).
Beyond this case, we provide two concrete definitions of $\hat{\mathcal{A}}(\bm{a}^r, \bm{a}^h)$ in the following sections.

% Compared to point-label BC, this set-valued supervision is weaker, while remaining more informative than pairwise comparisons because it constrains a region of the action space.
% The assumption $\hat{\mathcal{A}}(\bm a^r,\bm a^h)\cap \mathcal{A}^*_{\bm s} \neq \varnothing$ is also weaker than the standard BC assumption $\bm a^h \in \mathcal{A}^*_{\bm s}$.
% BC is recovered as the limiting case with $\hat{\mathcal{A}}(\bm a^r,\bm a^h)=\{\bm a^h\}$. 
% % Our framework therefore generalizes BC while retaining informative supervision.
% Beyond this case, we provide two concrete definitions of $\hat{\mathcal{A}}(\bm{a}^r, \bm{a}^h)$ in the following sections.
% In practice, $\hat{\mathcal{A}}(\bm a^r,\bm a^h)$ is instantiated by a parameterized construction; we assume its hyperparameters can be chosen so that $\hat{\mathcal{A}}(\bm a^r,\bm a^h)\cap \mathcal{A}^*_{\bm s}\neq\varnothing$.

\subsection{Polytope Desired Action Set}
\label{sec:sub:desired_action_space_relative}
\begin{figure}[t!]
	\centering
	\includegraphics[width=0.48\textwidth]{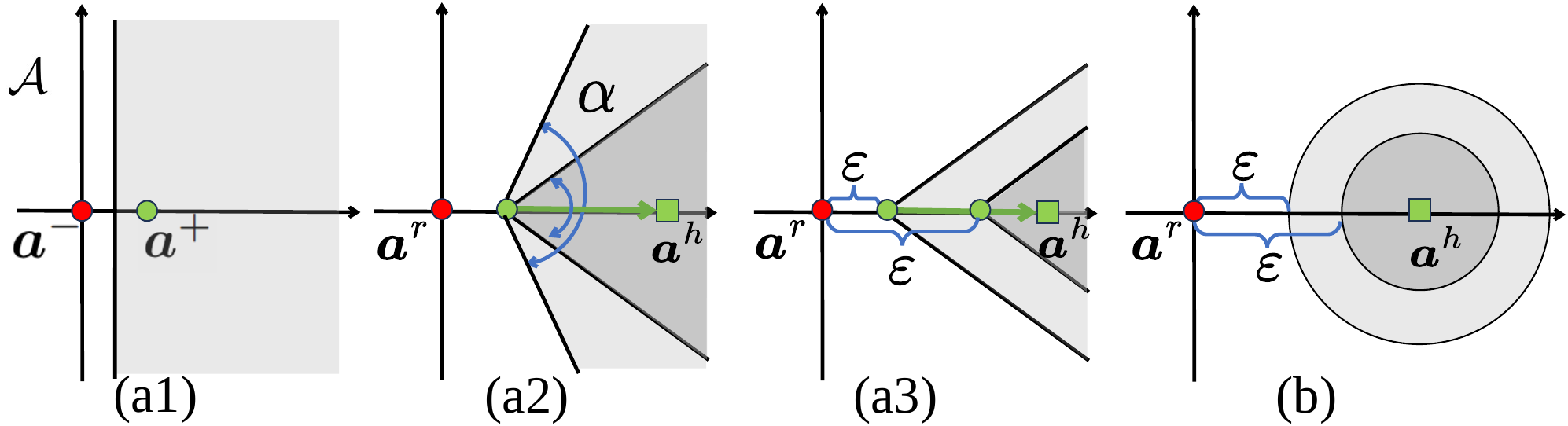}
	\caption{
 2D examples of desired action sets, shown as gray-shaded regions. (a1) Half-space desired set. (a2) Polytope desired action set with different $\alpha$. (a3) Polytope desired set with different $\varepsilon$. (b) Circular desired action set with different $\varepsilon$.}
\label{fig:Desired_action_space_twoTypes}
\end{figure}

Here, we detail how to construct a polytope desired action set from one corrective feedback.
The basic building block is a 
half-space desired set: 
a region defined by a linear hyperplane that partitions the entire action space into two.
For one corrective feedback, multiple half-space desired sets can be defined.
The polytope desired action set can then be obtained by intersecting these half-spaces.

\subsubsection{Half-space Desired Action Set}
\label{sec:sub:desired_halfspace}
A half-space $\mathcal{A}^H$ can be constructed from a pair of actions $(\bm  a^{-}, \bm a^{+})$:
\begin{equation}
    \mathcal{A}^H{(\bm  a^{-}, \bm a^{+})} = \{ \bm a \in \mathcal{A} | \mathbb{D}(\bm a, \bm a^-) \geq \mathbb{D}(\bm a, \bm a^+) \}.
    \label{eq:desired_half_space}
\end{equation}
where $\mathbb{D}(\bm a_1, \bm a_2) = \|\bm a_1 - \bm a_2 \|$.
We refer to $(\bm  a^{-}, \bm a^{+})$ as a \emph{contrastive action pair}.
Geometrically, $\mathcal{A}^H(\bm a^{-}, \bm a^{+})$ is the region of actions that are closer to $\bm a^{+}$ than to $\bm a^{-}$, as shown in Fig.~\ref{fig:Desired_action_space_twoTypes}(a1).
In particular, when we choose $(\bm a^{-}, \bm a^{+}) = (\bm a^r, \bm a^h)$, the resulting set $\mathcal{A}^H(\bm a^r, \bm a^h)$ satisfies the general definition of a desired action set in Section~\ref{sec:sub:general_def_desiredA}. We therefore refer to $\mathcal{A}^H(\bm a^r, \bm a^h)$ as the \emph{half-space desired action set}.
However, this set can be uninformative, as it classifies exactly half of $\mathcal{A}$ as desired.
To obtain a more informative $\hat{\mathcal{A}}{(\bm  a^{r}, \bm a^{h})}$, we therefore introduce the polytope desired action set.

   \subsubsection{Intersection of  Half-space Desired Action Sets}
   \label{sec:sub:intersection_polytope}
Given a corrective feedback $(\bm a^r, \bm a^h)$, 
we generate $N_I$ contrastive action pairs $(\bm a^{-}_i, \bm a_i^{+}), i=1, \dots, N_I,$ via data augmentation.
We define the intersection of the  half-space desired sets enforced by each contrastive action pair as
\begin{equation}
\label{eq:def_desired_action_space_From_Halfspaces}
\hat{\mathcal{A}}^H(\bm a^r, \bm a^h) =  \bigcap\nolimits_{i=0}^{N_I} \mathcal{A}^H{ (\bm a^{-}_i, \bm a_i^{+})}. 
\end{equation}
Here, $\{\bm a^{-}_i, \bm a^{+}_i\}_i^{N_I} = \mathrm{DataAug}(\bm a^r, \bm a^h; \varepsilon, \alpha)$ (see Appendix \ref{appendix:sub:CLIC_one_corrective_feedback} for details). 
The resulting geometry is a polytope, which approaches a cone as $N_I \rightarrow \infty$.
We refer to $\hat{\mathcal{A}}^H(\bm a^r, \bm a^h)$ as the \emph{polytope desired action set}.
Compared with a single half-space, the intersection reduces the volume of the action space classified as desired, thereby providing a more informative desired action set.
The magnitude certainty parameter $\varepsilon \in [0, 1)$ controls the cone's apex, located at $(1- \varepsilon)\bm a^r + \varepsilon\bm a^h$, while the directional uncertainty parameter $\alpha \in (0^\circ, 180^\circ]$ controls the cone's opening angle. 
Examples of $\hat{\mathcal{A}}^H(\bm a^r, \bm a^h)$ for varying $\alpha$ and $\varepsilon$ are shown in Fig.~\ref{fig:Desired_action_space_twoTypes}(a2)-(a3).

\subsection{Circular Desired Action Set}
\label{sec:sub:desired_action_space_absolute}

The polytope construction in Section \ref{sec:sub:desired_action_space_relative} is applicable to both relative and absolute corrections.
However, when the human feedback is known to be an absolute correction, it provides stronger information than a purely directional correction and can be used to define a more informative desired set.
We therefore define, for absolute corrections,  the \emph{circular desired action set} as
\begin{equation}
    \!\!\hat{\mathcal{A}}^C{(\bm a^r, \bm a^h)} \!=\! \{ \bm a \in \mathcal{A} |(1-\varepsilon)  \mathbb{D}(\bm a^r, \bm a^h) \geq \mathbb{D}(\bm a, \bm a^h) \},\!\!  
    \label{eq:desired_circular_space}
\end{equation}
which is the ball centered at $\bm a^h$ with radius $(1-\varepsilon) \cdot \mathbb{D}(\bm a^r, \bm a^h)$.
Intuitively, this set captures the idea that, once the teacher provides an absolute correction, other desired actions should remain in a neighborhood of that correction rather than only satisfying a directional constraint.
The hyperparameter $\varepsilon \in [0, 1)$ adjusts the radius of the ball. 
For $\varepsilon \rightarrow 1$, the ball reduces to the single action point $\bm a^h$, recovering pointwise supervision as a limiting case. 
Examples are shown in Fig.~\ref{fig:Desired_action_space_twoTypes}(b).

% We refer to CLIC using circular desired action sets $\hat{\mathcal{A}}^C{(\bm a^r, \bm a^h)}$ as \emph{CLIC-Circular}, and CLIC with polytope desired action sets $\hat{\mathcal{A}}^H{(\bm a^r, \bm a^h)}$ as \emph{CLIC-Half}.

\subsection{CLIC Algorithm and Overall Desired Action Set}
\label{sec:overall_desiredA_space}

The desired action set $\hat{\mathcal{A}}(\bm a^r, \bm a^h)$, resulting from a single corrective feedback, is intentionally weaker than the BC's pointwise supervision in Section~\ref{sec:sub:general_def_desiredA}: it contains at least one optimal action but may also include suboptimal actions. 
Therefore, a single feedback signal is generally insufficient to determine the optimal action. This motivates aggregating multiple corrective feedback signals, so that the desired region can be progressively refined as more information is collected.

\subsubsection{Overall Desired Action Set}
During interactive learning, the corrective feedback provided by the teacher is aggregated into a data buffer
$
\mathcal{D} = \{[\bm s_t, \bm a^r_t, \bm a^h_t], t = 1, \dots \}.$
The desired action sets from these feedback signals can collectively define an \textit{overall desired action set} for each state. 
Formally,
we denote the feedback received at state \(\bm{s}\) by \(\mathcal{D}_{\bm s} = \{[\bm{s}, \bm{a}^r_i, \bm{a}^h_i], i = 1, \dots, k\}  \subset  \mathcal{D}\), where \(k\) indicates the number of feedback inputs.
We denote the overall desired action set enforced by $\mathcal{D}_{\bm s}$ as $\hat{ \mathcal{A}}^{\mathcal{D}_{\bm s}}_k$.
 For the unimodal case, we define
$\hat{ \mathcal{A}}^{\mathcal{D}_{\bm s}}_k = \cap_i^k \hat{\mathcal{A}}(\bm a^r_i, \bm a^h_i)$.
This intersection refines the overall desired region as more feedback is obtained, as illustrated in Fig.~\ref{fig:Fig12_illustration_convergence_including_circle_2}.

\subsubsection{Policy as an Approximation of the Overall Desired Action Set}
Rather than explicitly maintaining $\hat{ \mathcal{A}}^{\mathcal{D}_{\bm s}}_k$ for every state, 
% CLIC uses
% we realize the set-valued supervision through a surrogate loss.
we formulate a surrogate loss to encode set-valued supervision.
This loss encourages the learner's policy to increase the probability of selecting actions within the desired action sets:
\begin{equation}
    l_{\pi_{\bm \theta}}= - \log \pi_{\bm \theta}(\bm{a} \in \hat{\mathcal{A}}(\bm{a}^r, \bm{a}^h) | \bm{s}).
    \label{eq:loss_desired_action_space_high_level}
\end{equation}
This loss is presented in detail in Section \ref{sec:Policy_shaping}. 
We then use $l_{\pi_{\bm \theta}}$ to update $\pi_{\bm \theta}$ via sampled batches from the data buffer, as in line \ref{alg:line:update_policy_once} of Algorithm \ref{alg:CLIC_algorithm}. 
This update process progressively shapes the policy distribution $\pi_{\bm \theta}(\bm a | \bm s)$, causing it to concentrate its probability mass within the accumulating desired regions. As a result, the policy serves as an approximation of the overall desired action set $\hat{\mathcal{A}}^{\mathcal{D}_{\bm s}}_k$.

\begin{algorithm*}[!t]
\caption{CLIC: Contrastive policy Learning from Interactive Corrections}\label{alg:CLIC_algorithm}
\begin{multicols}{2}
\DontPrintSemicolon
\SetInd{0.1em}{0.6em} % change the indentation, see: https://tex.stackexchange.com/questions/459919/how-to-change-the-indent-length-for-blocks-in-algorithm2e
% Define the functions using the algorithm2e syntax
\SetKwFunction{FDesiredA}{DesiredActionSet}
\SetKwFunction{Flearning}{Learning}
\SetKwFunction{Fimplicit}{ImplicitPolicyShaping}
\SetKwFunction{FPolicyshaping}{PolicyShaping}
\SetKwProg{Fn}{Function}{:}{}

% \textbf{Notations}\;
% \hspace*{0.2mm} $\mathcal D$: data buffer\;
%  \hspace*{0.2mm} $\mathcal B$: batch sampled from data buffer $\mathcal D$:\;
%  \hspace*{0.2mm} $E_{\bm \theta}$: Energy-based Model\;
% \hspace*{0.2mm} $\mathbb{A}$: action samples used to approximate the KL loss\;
% \hspace*{0.2mm} $\hat{\mathcal A}_i$: shorthand for $\hat{\mathcal{A} }{(\bm a^r_i , \bm a^h_i)}$\;
% \hspace*{0.2mm} $p_{\!\hat{\mathcal A},i}$: shorthand for observation model $\!p \big(\!\bm a \! \in\! \!\hat{\mathcal{A} }{(\bm a^r_i , \bm a^h_i)}|\bm a , \! \bm s_i\!\big)\!$\;
% \hspace*{0.2mm} $b$: update frequency during each episode\;
% \hspace*{0.2mm} $N_{\text{training}}$: training steps at end of each episode\;
% \hspace*{0mm} $type_{\!\hat{\mathcal{A}}}\in \{\textit{Polytope},\textit{Circular}\}$: type of desired action spaces\;

\textbf{Notations}
\small
\begin{tabular}{@{}l@{\,:\,}l}
 $\mathcal{D}$ & Data buffer of corrective feedback $(\bm s, \bm a^r, \bm a^h)$ \\
 $\mathcal{B}$ & Batch sampled from data buffer $\mathcal{D}$ \\
 $\pi_{\bm \theta}$ & Policy parameterized by $\bm\theta$, via an energy model $E_{\bm \theta}$ \\
 $\mathbb{A}_i$ & Set of action samples for state $\bm s_i$ \\
 $\hat{\mathcal{A}}_i$ & Shorthand for $\hat{\mathcal{A} }{(\bm a^r_i , \bm a^h_i)}$ \\
 $p_{\hat{\mathcal{A}}_i}$ & Shorthand for observation model $\!p \big(\!\bm a \! \in\! \!\hat{\mathcal{A} }{(\bm a^r_i , \bm a^h_i)}|\bm a , \! \bm s_i\!\big)\!$ \\
 $b$ & In-episode update frequency \\
 $N_{\text{training}}$ & End-of-episode training steps \\
 $\eta$ & Learning rate \\
 $type_{\!\hat{\mathcal{A}}}$ & Type of the desired action set, $\{\textit{Polytope},\textit{Circular}\}$ \\
\end{tabular}
\normalsize

\vspace{2mm}
\code{Interactive Imitation Learning Loop (Fig. \ref{fig:framework}A and \ref{fig:CLIC_overview}) }\;
\For{episode = 1, 2, \dots}{
 \For{$t = 1, 2, \dots$}{
 Observe $\bm s_t$, execute $\bm a^r_t$\;
 Receive feedback $\bm a^ h_t $, if $\bm a^r_t$ is suboptimal \label{alg:line:receive_feedback}\;
Append $[\bm s_t, \bm a^r_t, \bm a^ h_t]$ to $\mathcal D$, if $\bm a^ h_t$ is provided\;
\If{$t \% b = 0$ or $\bm a^h_t$ is provided}{ \label{alg:line:update_feq_b}
Sample batch $\mathcal B$ from $\mathcal D$\;
\label{alg:line:sample_batch}
% $\mathcal B^{\mathcal{A}}, p_{\hat{\mathcal{A}}} \leftarrow$ \FDesiredA{$\mathcal{B}$, $type_{\mathcal{A}}$}\;
% $\pi_{\bm \theta} \leftarrow$ \FPolicyshaping{$\mathcal B$,  $\mathcal B^{\mathcal{A}}$, $\pi_{\bm \theta}$, $p_{\hat{\mathcal{A}}}$}\;\label{alg:line:update_policy_once}
$\big\{\, p_{\hat{\mathcal A}_i}\big\}_{i\in\mathcal B} \!\!\leftarrow$ \!\!\FDesiredA{$\mathcal B$, $type_{\!\hat{\mathcal{A}}}$}\;\label{alg:line:desired_action_space}
      $\bm\theta \leftarrow$ \FPolicyshaping{$\mathcal B$,  $\{p_{\hat{\mathcal A_i}}\}$, ${\bm\theta}$}\;\label{alg:line:update_policy_once}
}
 }
 
 Update policy $\pi_{\bm \theta}$ as in lines \ref{alg:line:sample_batch}-\ref{alg:line:update_policy_once} for $N_{\text{training}}$ steps \label{alg:line:end_of_episode}\;
}
\columnbreak

\code{Constructing Desired Action Sets (Fig. \ref{fig:Desired_action_space_twoTypes} and \ref{fig:Fig2_illustration_2d_measurement_model})}\;
\Fn{\FDesiredA{$\mathcal{B}$, $type_{\!\hat{\mathcal{A}}}$}}
{
\ForEach{$(\bm s_i,\bm a^r_i,\bm a^h_i)\in\mathcal B \textup{ in parallel}$}{
\uIf{$type_{\!\hat{\mathcal{A}}}$ = \text{Polytope}}{
Generate contrastive action pairs as Eq.~\eqref{eq:assumption_negative_correction} and \eqref{eq:implicit_action_pairs}\;
% Appendix \ref{appendix:sub:CLIC_one_corrective_feedback}\;
Create $\hat{\mathcal{A} }{(\bm a^r_i , \bm a^h_i)} $ via Eq.~\eqref{eq:def_desired_action_space_From_Halfspaces}\;
Define  $p \big(\!\bm a \! \in \!\hat{\mathcal{A} }{(\bm a^r_i, \bm a^h_i)} |\bm a, \bm s_i\big) $ via Eq.~\eqref{eq:observation_Model_half}\;
}
\ElseIf{$type_{\!\hat{\mathcal{A}}} = \text{Circular}$}{
   Create $\hat{\mathcal{A} }{(\bm a^r_i , \bm a^h_i)} $ via Eq.~\eqref{eq:desired_circular_space}\;
   Define  $p \big(\!\bm a \! \in \!\hat{\mathcal{A} }{(\bm a^r_i , \bm a^h_i)} |\bm a  , \bm s_i\big) $ via Eq.~\eqref{eq:observation_model_circular}\;
}
}
\Return{$\{p_{\hat{\mathcal A_i}}\}$}
}

\vspace{1mm}
\code{Policy shaping via Desired Action Sets (Fig. \ref{fig:Fig9_illustration_implicit_policy_loss_effects})}\;
\Fn{\FPolicyshaping{$\mathcal B$, $\{p_{\hat{\mathcal A_i}}\}$, ${\bm\theta}$}}{\label{alg:implicit_policy_shaping}
\ForEach{$(\bm s_i,\bm a^r_i,\bm a^h_i)\in\mathcal B \textup{ in parallel}$ }{
Draw $\mathbb A_i$ from $E_{\bm \theta}(\bm s_i, \cdot)$, as Eq.~\eqref{eq:sampled_actions_EBM}\label{alg:line:MCMC}\;
Calculate $p \big(\!\bm a \! \in \!\hat{\mathcal{A} }{(\bm a^r_i , \bm a^h_i)} | \bm a , \bm s_i\big) \!, \bm a\in \mathbb A_i\!$\label{alg:line:observation_model_cal}\;
Estimate $\pi_{\bm \theta}(\bm a | \bm s_i) \! \propto\! {\exp(\!-E_{\bm \theta}(\bm s_i, \bm a))}, \bm a \!\in \!\mathbb A_i$,as Eq.\eqref{eq:estiamtion_policy_EBM} \label{alg:line:estimate_policy}\;
Estimate target $\pi^{\text{target}}(\bm a| \bm s_i, \hat{\mathcal{A} }_i) , \bm a \in \mathbb A_i $,  via Eq.~\eqref{eq:policy-weighted-bayes-loss}\label{alg:line:estimate_target}\;
Accumulate $ \ell(\bm\theta) +\!=\mathbb{E}_{\mathcal{B}} \left[ \mathrm{KL}\!\left( \pi^{\text{target}}(\cdot| \bm s_i) \big\| \pi_{\bm \theta}(\cdot | \bm s_i) \right) \! \right] $ \label{alg:line:KL_loss}\;
}
$\bm \theta \leftarrow \bm \theta -\eta\nabla_{\bm\theta}\ell(\bm\theta) $\;
\Return{$\bm \theta$}
}
\vspace*{0.05mm}
\end{multicols}
\end{algorithm*}

\subsubsection{Algorithm of CLIC}
\label{sec:sub:sub:algorithm_CLIC}
\begin{figure}[t!]
	\centering
	\includegraphics[width=0.475\textwidth]{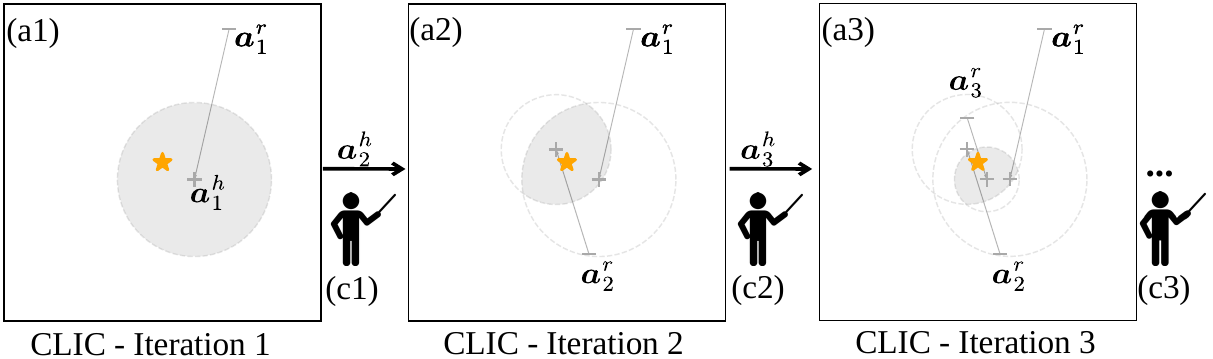}
    % \vspace*{-5mm}
	\caption{
Overall desired action set formed by aggregating multiple teacher feedback signals in a single state.
This set in (a) gets progressively refined as additional corrections (c) are provided.
The examples illustrate CLIC-Circular; corresponding examples for CLIC-Half are shown in Fig. \ref{fig:CLIC_overview}.
 % (a) Current overall desired space. (c) The teacher provides corrections on the robot action. 
%  (1) A human action $\bm a^h_1$ is provided, forming an initial overall desired action set $\hat{ \mathcal{A}}^{\mathcal{D}_{\bm s}}_1 = \hat{\mathcal{A}}{(\bm a^r_1, \bm a^h_1)} $. The robot policy is updated to generate actions within this space. 
%  (2) $\bm a^r_2$ is selected, and $\bm a^h_2$ is provided, shrinking the overall desired space to $\hat{ \mathcal{A}}^{\mathcal{D}_{\bm s}}_2$.
% % The policy is then updated accordingly.
%  (3) The process continues with further feedback (e.g., $\bm a^h_3$), leading to a smaller overall desired space $\hat{ \mathcal{A}}^{\mathcal{D}_{\bm s}}_3$.  
 }
\label{fig:Fig12_illustration_convergence_including_circle_2}
\end{figure}

CLIC follows the standard IIL loop and refines the above overall desired region through iterative policy improvement.
Algorithm~\ref{alg:CLIC_algorithm} summarizes the complete procedure, with the key steps shown from line \ref{alg:line:sample_batch} to line \ref{alg:line:update_policy_once}.
At line \ref{alg:line:desired_action_space}, desired action sets are generated for each feedback signal in the sampled batch.
Line \ref{alg:line:update_policy_once} then computes a surrogate loss based on these desired action sets to update the policy $\pi_{\bm \theta}$ (introduced in Section~\ref{sec:Policy_shaping}).
Through this update, the policy approximates the current overall desired set and is therefore encouraged to produce actions within this overall set in subsequent iterations.
If the policy still performs suboptimally, the teacher provides additional corrective feedback (line \ref{alg:line:receive_feedback}). 
This new feedback further refines the overall desired set, and the policy is updated again so that its probability mass increasingly lies within the refined set (Fig.~\ref{fig:Fig12_illustration_convergence_including_circle_2}).
This process iteratively improves the policy with aggregated corrections.
Depending on the desired action set used in line \ref{alg:line:desired_action_space}, we obtain two CLIC variants. 
We refer to CLIC with polytope desired action sets $\hat{\mathcal{A}}^H(\bm a^r,\bm a^h)$ as \emph{CLIC-Half}, and CLIC with circular desired action sets $\hat{\mathcal{A}}^C(\bm a^r,\bm a^h)$ as \emph{CLIC-Circular}.

\subsubsection{Convergence Conditions for the Overall Desired Action Set}
\label{sec:sub:sub:overall_desired_space}
For simplicity, we analyze the unimodal case at a fixed state $\bm s$. 
In Appendix \ref{appendix:proof_convergence_DesiredA}, we show that, for any state $\bm s$,
$  \lim_{k\to\infty}
  \hat{\mathcal A}^{\mathcal D_{\bm s}}_k
  \subseteq
  \mathcal A^*_{\bm s}$ under the following sufficient conditions:
\begin{enumerate}[label= (A\arabic*)]
  \item For all iterations $k$, the policy $\pi_{\bm\theta_k}(\cdot|\bm s)$ assigns nonzero probability mass to the current overall set $\hat{\mathcal{A}}^{\mathcal{D}_{\bm s}}_k$.
  \item For any sampled suboptimal action $\bm a^r_k \sim \pi_{\bm \theta_k}(\cdot | \bm s)$, there is a nonzero probability for the teacher to provide corrective feedback $\bm a^h_k$.
  \item For each teacher feedback $\bm a^h_k$, the desired action set $\hat{\mathcal{A}}(\bm a^r_k, \bm a^h_k)$ includes at least one optimal action.
\end{enumerate}
This analysis formalizes how aggregated corrective feedback progressively refines the overall desired action set at a state.
Condition (A2) is a standard assumption on teacher availability in human-gated IIL \cite{2019_Rodrigo_D_COACH, 2019_HG_DAgger}.
Condition (A3) corresponds to the assumption in Section~\ref{sec:sub:general_def_desiredA}; 
 in practice, satisfying it requires choosing a desired-set construction (e.g., polytope vs.\ circular) and hyperparameters that reflect the feedback modality and its uncertainty. We examine the effect of these choices in Sections~\ref{sec:exp:simulation} and~\ref{sec:exp:ablation}.
% in practice, we instantiate $\hat{\mathcal{A}}(\bm a^r,\bm a^h)$ by selecting an appropriate set construction (e.g., polytope vs.\ circular) and hyperparameters that reflect the feedback modality and its uncertainty.
% We analyze the sensitivity of these choices in Sections~\ref{sec:exp:simulation} and~\ref{sec:exp:ablation}. In practice, the desired-set construction should be chosen to reflect the feedback modality and its uncertainty; otherwise, the induced set may exclude optimal actions and degrade performance.
To satisfy condition (A1), we introduce a policy-shaping loss (Eq.~\eqref{eq:loss_desired_action_space_high_level}), which increases the probability mass that $\pi_{\bm\theta}(\cdot|\bm s)$ assigns to the desired action sets.
The details of this loss and its implementation are presented in Section~\ref{sec:Policy_shaping}.

\textbf{Practical scope of the analysis}:
The above analysis is stated for aggregated corrections at a fixed state. For continuous state-action spaces, it is unlikely to receive multiple corrective feedback at the exact same state. 
% In practice, CLIC therefore relies on function approximation to transfer the effect of a desired action set across similar states. As a result, the learned feasible action region is refined through the generalization capabilities of the policy network, rather than through explicit per-state set combination. 
In practice, CLIC therefore relies on the generalization capabilities of the policy network to transfer the effect of a desired action set across similar states.
% , thus refining the policy-induced action distribution more effectively.
Likewise, for multi-modal tasks, sufficiently expressive parameterizations (e.g., EBMs or diffusion models) can represent multiple optimal actions, allowing the learned policy to approximate multiple refined desired regions at a given state. We examine this behavior empirically in Sections~\ref{sec:exp:simulation} and~\ref{sec:exp:real_rotbo:insert-T}.

\section{Policy Shaping via Desired Action Sets}
% \section{Policy shaping}
\label{sec:Policy_shaping}

% In this section, we describe how to train a robot policy using desired action sets.
% First, in Section~\ref{sec:sub:implicit_policy_shaping}, we introduce the loss function to align the robot policy with a target policy.
% Next, in Section \ref{sec:target_policy}, we define this target policy via Bayes’ rule, specifying the observation model and prior.
% Then, Section~\ref{sec:sub:algorithm_implicit_policy_shaping} summarizes the resulting algorithm for implicit policy shaping.
% Finally, in Section~\ref{sec:sub:explicit_policy_shaping},  we provide a simplified formulation, which assumes a Gaussian-parameterized policy.

In this section, we describe how desired action sets are used to train the robot policy. Given a desired action set constructed from corrective feedback, we seek to update the policy so that its action distribution shifts toward actions aligned with that set. Section~\ref{sec:sub:implicit_policy_shaping} introduces the loss function for policy shaping, and Section~\ref{sec:target_policy} defines the target distribution used by this loss, including its observation model and prior. Section~\ref{sec:sub:algorithm_implicit_policy_shaping} summarizes the resulting policy shaping algorithm. Finally, Section~\ref{sec:sub:explicit_policy_shaping} presents a simplified formulation under a Gaussian policy assumption.

\subsection{Loss function for Policy Shaping}
\label{sec:sub:implicit_policy_shaping}

Here, we present the loss function for training the robot policy using the desired action sets. 
Our goal is to increase the probability of $\pi_{\bm \theta}$ selecting actions within the set $\hat{\mathcal{A}}(\bm{a}^r, \bm{a}^h)$ (Eq.~\eqref{eq:loss_desired_action_space_high_level}).
To do so, the policy $\pi_{\bm \theta}$ can be updated to align with a target distribution that assigns a high probability to actions within this set. 
We define this target as $\pi^{\text{target}}(\bm a| \bm s, \hat{ \mathcal{A}}{(\bm a^r, \bm a^h)} ) $, abbreviated as $\pi^{\text{target}}(\bm a \mid \bm s)$ when the set is clear from context.
% We will detail the definition of $\pi^{\text{target}}(\bm a| \bm s)$ in Section \ref{sec:target_policy}.
Section~\ref{sec:target_policy} formalizes $\pi^{\text{target}}$ and specifies how it is computed from $\hat{\mathcal A}(\bm a^r,\bm a^h)$.

To align the robot policy $\pi_{\bm \theta}(\bm a| \bm s )$ with $\pi^{\text{target}}(\bm a| \bm s ) $, $\pi_{\bm \theta}$ can be optimized by minimizing the KL divergence between the target and the policy distribution:
\begin{align}
    \ell_{KL}(\bm \theta) = \!\! \!\!\underset{(\bm{a}^{h}, \bm{a}^{r}, \bm{s}) \sim p_{\mathcal D}}{\mathbb{E}} \!\!\left[ \mathrm{KL}\left(\pi^{\text{target}}(\bm a| \bm s ) \big\| \pi_{\bm \theta}(\bm a | \bm s) \right) \right] 
    \label{eq:KL_loss_general}
\end{align}
To approximate Eq.~\eqref{eq:KL_loss_general}, 
we estimate both $\pi^{\text{target}}\!(\bm a| \bm s )\!$ and $\pi_{\bm \theta}(\bm a | \bm s)$ using a set of sampled actions, defined as follows:
\begin{equation}
    \mathbb A = \{ \bm a^h, \bm a^r\} \cup   \{\bm a_j\}_{j=1}^{N_a}, \label{eq:sampled_actions_EBM}
\end{equation}
where $ N_{\text{a}}$ samples $\{\bm a_j \}$ can be obtained by Langevin MCMC from the robot policy as described in Section \ref{sec:pre:policy}. 
We explicitly include $\{\bm a^h,\bm a^r\}$ so that both the human-corrected action and the robot action  are always included in
the sampled action set $\mathbb A $.
Given $\mathbb A $, the policy $\pi_{\bm \theta}$ evaluated at each sampled action can be approximated using Eq.~\eqref{eq:estiamtion_policy_EBM}.

\textbf{Connections to InfoNCE Loss:} 
The KL objective can be reformulated as a weighted sum of the InfoNCE loss defined in Eq.~\eqref{eq:ibc_info_NCE}. This is achieved by substituting Eq.~\eqref{eq:estiamtion_policy_EBM} into the KL divergence and neglecting the constant term $c$ ({Appendix \ref{appendix:connection_INFONCE}}).
\begin{align*}
    \ell_{\!KL}\!(\bm \theta) \!
   \simeq \!\!\!\!
    \!\!\!\! \underset{(\bm{a}^{h}\!, \bm{a}^{r}\!, \bm{s})  \sim p_{\mathcal D}}{\mathbb{E}} \!\! \sum_{\bm a \in \mathbb A} \!\! -\pi^{\text{target}}(\bm a| \bm s ) \ell_{\text{InfoNCE}}(\bm s, \bm a, \mathbb A \! \backslash  \! \{\bm a\}) \!\!+ \!c
\end{align*}
The key difference from standard InfoNCE loss in IBC is how the target policy is defined: 
BC corresponds to the limiting case with the desired set $\hat{\mathcal{A}}(\bm a^r,\bm a^h)=\{\bm a^h\}$, in which the target policy places all probability mass on the human action.
Our KL objective instead assigns each $\bm a\in\mathbb A$ a soft weight $\pi^{\text{target}}(\bm a| \bm s)$ based on compatibility with the desired action set.
 This distributes learning signal across multiple desired actions and reduces overfitting to a single (potentially noisy) action label.

\subsection{Target Distribution for Aligning with a Desired Action Set}
\label{sec:target_policy}
We now define the target $ \pi^{\text{target}}$ from a desired action set $\hat{\mathcal A}(\bm a^r,\bm a^h)$. 
Conceptually, we view the set $\hat{\mathcal A}(\bm a^r,\bm a^h)$ as an observation about which actions are acceptable at state $\bm s$. 
Unlike a pointwise label, this observation does not uniquely specify a single action; rather, it provides evidence over the action space that guides the policy toward actions consistent with the desired set and away from inconsistent ones.
We implement this idea by constructing $\pi^{\text{target}}$ as a posterior distribution: starting from a prior over actions at $\bm s$, we update this prior using a soft set-membership likelihood induced by $\hat{\mathcal A}(\bm a^r,\bm a^h)$. Formally, we define
\begin{align}
\!\!\pi^{\text{target}}(\bm a| \bm s, \hat{ \mathcal{A}}{(\bm a^r\!, \bm a^h)} ) \!=\!\frac{ p\big(\!\bm a \in\hat {\mathcal{A}} {(\bm a^r\!, \bm a^h)} | \bm a , \bm s \big) p(\bm a | \bm s)}{p (\bm a \in\hat {\mathcal{A}} {(\bm a^r\!, \bm a^h)}  | \bm s)}\!,\!\! 
    \label{eq:posterior_data}
\end{align}
where 
$p(\bm a | \bm s)$ is the prior and $p\big(\!\bm a \in\hat {\mathcal{A}} {(\bm a^r\!, \bm a^h)} | \bm a , \bm s \big)$ is the observation model that quantifies the degree to which action $\bm a$ is consistent with the desired action set at state $\bm s$.
The denominator is the normalizing constant.
% In practice, we compute this posterior over the sampled actions $\mathbb A$ (Eq.~\eqref{eq:sampled_actions_EBM}). 
Sections~\ref{section:sub:prob_desired_action_space} and~\ref{section:sub:prior_distri} specify the observation model and the prior, respectively.

\subsubsection{Observation Model of Desired Action Sets}
\label{section:sub:prob_desired_action_space}
We interpret the desired action set $\hat {\mathcal{A}} {(\bm a^r, \bm a^h)}$ as a noisy observation about which actions are consistent with the corrective feedback $(\bm a^r, \bm a^h)$ at state $\bm s$.
Accordingly, the observation model $p \big(\bm a \in \hat{\mathcal{A} }{(\bm a^r, \bm a^h)} | \bm a , \bm s\big) $ defines a soft set-membership likelihood: it assigns a higher probability to actions that belong to the desired set and a lower probability to actions outside of the set.
In the noise-free case, this reduces to an indicator function $\mathbb{I}[\bm a \in \hat {\mathcal{A}} {(\bm a^r, \bm a^h)}]$.
To account for noisy human feedback and possible mismatch in the set construction, we replace this hard membership boundary with a smooth sigmoid.  
% We detail this for both CLIC-Half and CLIC-Circular: 

\begin{figure}[t]
	\centering
	\includegraphics[width=0.48\textwidth]{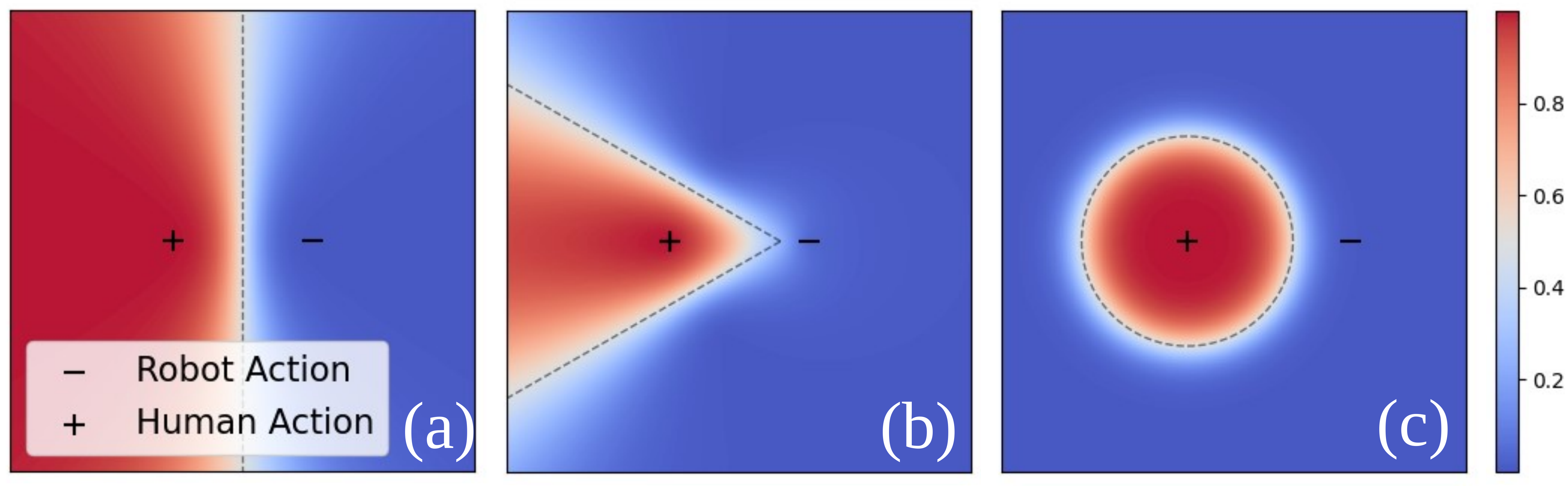}
    % \includesvg[width=0.49\textwidth, inkscapelatex=false]{figs/Fig2_illustration_2d_measurement_model.svg} 
	\caption{ Illustration of the observation model $\text{Pr} [\bm a \in \hat {\mathcal{A}} {(\bm a^r\!\!,\! \bm a^h)}| \bm a , \bm s]$ for all $\bm a \in \! \mathcal{ A}$. 
    In each figure, the state $\bm s$, human action $\bm a^h$, and robot action $\bm a^r$ are fixed, while the action $\bm a$ varies across the action space. The black dotted line denotes the boundary of $\hat {\mathcal{A}} {(\bm a^r\!\!, \!\bm a^h)}$.
    % , which is smoothed by a sigmoid function. 
    (a) Half-space desired action set; (b) Polytope desired action set; (c) Circular desired action set.}
	\label{fig:Fig2_illustration_2d_measurement_model}
\end{figure}

\textbf{Observation Model for CLIC-Half}:
Recall from Section~\ref{sec:sub:desired_halfspace} that a half-space $\mathcal{A}^H(\bm a^-, \bm a^+)$ contains actions closer to $\bm a^+$ than to $\bm a^-$. Its hard membership rule,
$\mathbb{I}[\mathbb{D}(\bm a, \bm a^-) - \mathbb{D}(\bm a, \bm a^+)\geq 0]$, can be smoothed as follows:
\begin{align}
   \!\! p(\bm a \in \!\mathcal{A}^H {(\bm a^-, \bm a^+)} | \bm a , \bm s) \!= \!  \sigma_T(\mathbb{D}(\bm a, \bm a^-) \!- \!\mathbb{D}(\bm a, \bm a^+)), \!\!
    \label{eq:observation_Model_half}
\end{align}
where $\sigma_T(x) = (1 + \exp (- x/T))^{-1}$ and $T > 0$ is a temperature parameter that controls how sharply the function transitions between 0 and 1. 
% As $T \rightarrow 0$, $\sigma_T(x)$ behaves more like a step function.
Fig.~\ref{fig:Fig2_illustration_2d_measurement_model}(a) shows a 2D example.

For a polytope desired set $\hat {\mathcal{A}} {(\bm a^r, \bm a^h)}$ defined as the intersection of half-spaces in Eq. \eqref{eq:def_desired_action_space_From_Halfspaces}, by leveraging conditional independence, we have
\begin{align*}
    p\big(\bm a \in\hat {\mathcal{A}} {(\bm a^r, \bm a^h)} | \bm a , \bm s \big) =  \!\prod\nolimits_{i=0}^{N_I} p\big(\bm a \in \mathcal{A}^H{ (\bm a^-_i, \bm a^{+}_i)}| \bm a, \bm s\big),
\end{align*}
where $(\bm a^-_i, \bm a^{+}_i)$ are the contrastive action pairs obtained from $(\bm a^r, \bm a^h)$.
Fig. \ref{fig:Fig2_illustration_2d_measurement_model}(b) illustrates this observation model for a polytope desired action set.

\textbf{Observation Model for CLIC-Circular}:
% In this case, similar to the way that we define Eq.~\eqref{eq:observation_Model_half}, the observation model for the circular desired action set is defined as: 
For the circular desired action set in Section~\ref{sec:sub:desired_action_space_absolute}, we similarly relax the hard membership rule into
\begin{align}
    p(\bm a \in \hat{\mathcal{A}} {(\bm a^r, \bm a^h)} | \bm a , \bm s) =   \sigma_T( r(\bm a^r, \bm a^h) \!-\! \mathbb{D}(\bm a, \bm a^h)), 
    \label{eq:observation_model_circular}
\end{align}
where $r(\bm a^r, \bm a^h)  =  (1-\varepsilon)\mathbb{D}(\bm a^r, \bm a^h)$ is the radius defined in Eq.~\eqref{eq:desired_circular_space}.
This assigns high probability to actions within distance $r$ of $\bm a^h$, and smoothly decreases outside the boundary.

\subsubsection{Prior choices (uniform vs. policy-weighted)}
\label{section:sub:prior_distri}
\begin{figure}
    \centering
    \includegraphics[width=0.48\textwidth]{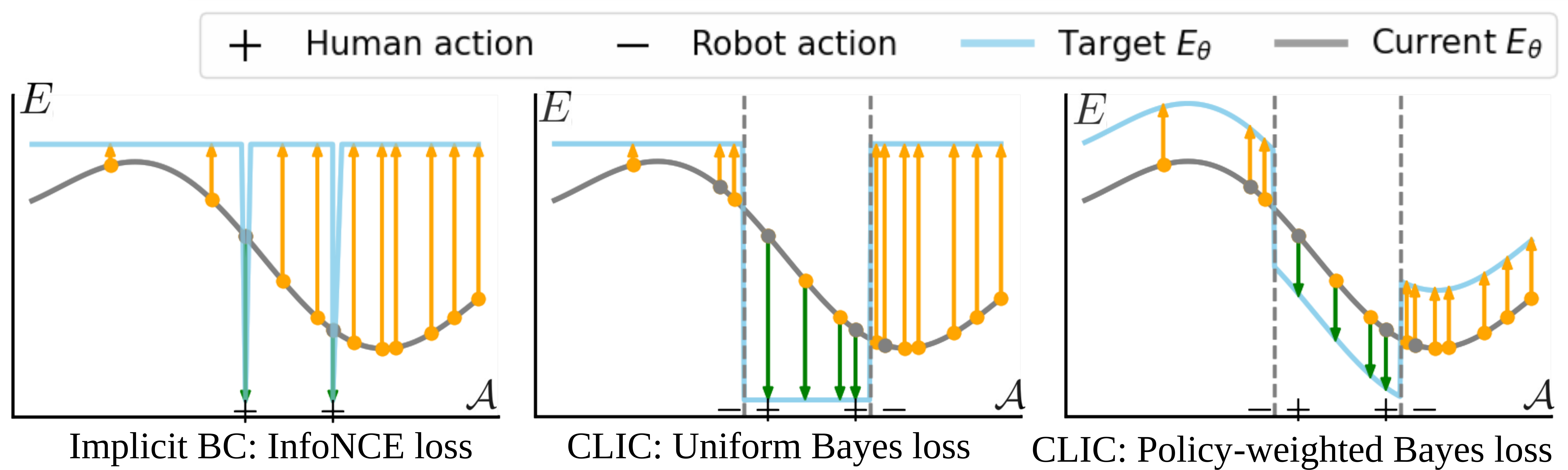}
    % \includesvg[width=0.49\textwidth, inkscapelatex=false]{figs/Fig6_illustration_loss_EBMs.svg} 
	\caption{Illustration of various loss functions for training an EBM policy in a 1D action space. Orange dots denote action samples, with orange arrows indicating increased energy values and green arrows showing decreased energy values.}
 \label{fig:Fig6_illustration_loss_EBMs}
\end{figure}

The prior $p(\bm a| \bm s)$ in Eq.~\eqref{eq:posterior_data} represents the initial belief over feasible actions before incorporating the desired-set observation.
A uniform prior yields a baseline target distribution
$\pi^{\text{target}}(\bm a| \bm s)\propto p(\bm a\in\hat{\mathcal A}(\bm a^r,\bm a^h)| \bm a,\bm s)$,
which weights all sampled actions inside the set equally. 
However, because $\hat{ \mathcal{A}}{(\bm a^r, \bm a^h)}$ may contain suboptimal actions, the uniform prior can overweight suboptimal regions of the set, potentially steering the policy in the wrong direction.
We instead use the current policy as the prior, $p(\bm a| \bm s)=\pi_{\bm\theta}(\bm a| \bm s)$.
This yields a conservative update. In addition, $\pi_{\bm\theta}(\cdot| \bm s)$ is progressively shaped by previously observed desired sets, and thus serves as an implicit approximation of the overall desired action set at $\bm s$.
The resulting policy-weighted target reweights the current policy by the set-membership likelihood:
\begin{align}
\pi^{\text{target}}(\bm a| \bm s)
~\propto~
p(\bm a\in\hat{\mathcal A}(\bm a^r,\bm a^h)| \bm a,\bm s)\,\pi_{\bm\theta}(\bm a| \bm s).
\label{eq:policy-weighted-bayes-loss}
\end{align}
In practice, we approximate $\pi^{\text{target}}$ on the sampled action set $\mathbb A$ (Eq.~\eqref{eq:sampled_actions_EBM}) using the energy-based estimate of $\pi_{\bm\theta}$ (Eq.~\eqref{eq:estiamtion_policy_EBM}).
We then minimize the KL loss in Eq.~\eqref{eq:KL_loss_general}.
We refer to this objective as the \emph{uniform Bayes loss} when the prior is uniform over $\mathbb A$, and as the \emph{policy-weighted Bayes loss} when the prior is the current policy.
The effects of these two loss variants are illustrated in Fig.~\ref{fig:Fig6_illustration_loss_EBMs}.
% Fig.~\ref{fig:Fig6_illustration_loss_EBMs} illustrates the effects of these two variants.
% Intuitively, the set-membership likelihood favors actions within the desired set, while the policy prior favors actions that already have high probability under $ \pi_{\bm \theta}$, reducing the risk of reinforcing suboptimal actions within the set.

 \begin{figure}
    \centering
    \includegraphics[width=0.48\textwidth]{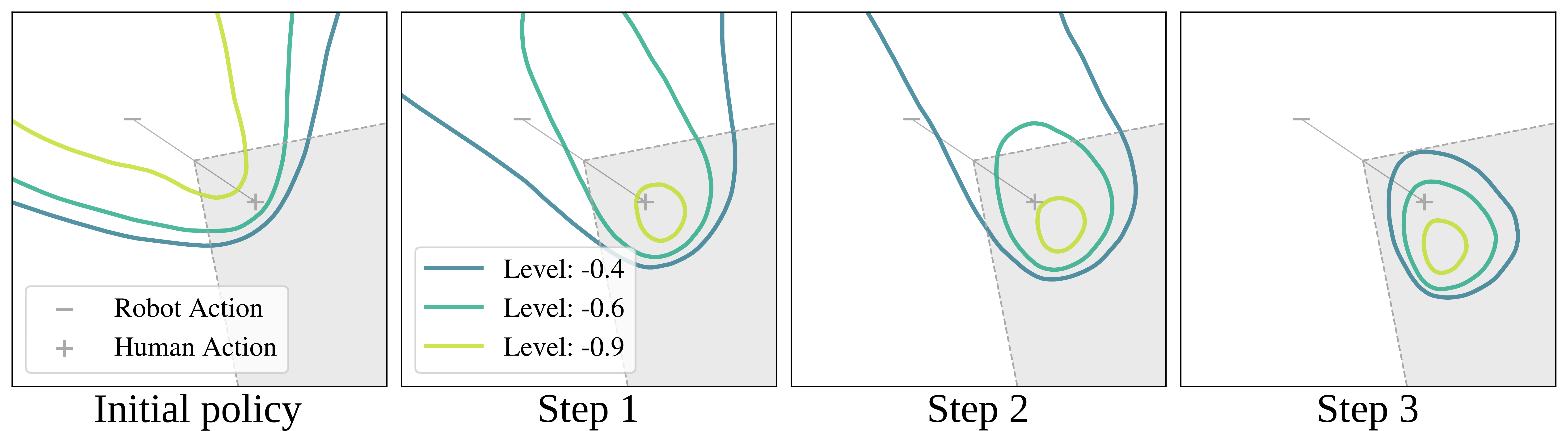}
    % \includesvg[width=0.49\textwidth, inkscapelatex=false]{figs/Fig9_illustration_implicit_policy_loss_effects.svg} 
    % \vspace*{-5mm}
	\caption{2D example of the policy shaping with Algorithm \ref{alg:CLIC_algorithm}. The batch size is 1, and the same corrective feedback is used over three steps. 
Initially, the EBM policy has the most density outside the desired action set. After the first update step, the peak density shifts toward the desired action set but still retains significant density outside it. 
With two additional update steps, the EBM policy is mostly inside the desired action set.  }
 \label{fig:Fig9_illustration_implicit_policy_loss_effects}
\end{figure}

\subsection{Algorithm for Policy Shaping}
\label{sec:sub:algorithm_implicit_policy_shaping}
Algorithm~\ref{alg:CLIC_algorithm} summarizes how desired action sets are incorporated into policy training. 
In each update step, action samples are drawn from the current policy  (line~\ref{alg:line:MCMC}). 
The observation model assigns each sampled action a score reflecting its consistency with the desired action set (line~\ref{alg:line:observation_model_cal}). 
These scores are combined with the estimated current policy to construct the target distribution over the sampled actions (lines~\ref{alg:line:estimate_policy}--\ref{alg:line:estimate_target}). 
Finally, the KL loss is evaluated to update the policy parameters $\bm\theta$ (line~\ref{alg:line:KL_loss}). 
Fig.~\ref{fig:Fig9_illustration_implicit_policy_loss_effects} illustrates how repeated updates shift more probability mass of the policy into the desired action set.

% An example of the training process is shown in Fig. \ref{fig:Fig9_illustration_implicit_policy_loss_effects}, which illustrates how the algorithm adjusts the density of an EBM and effectively aligns it with a desired action set.  

\subsection{Explicit Policy Shaping}
\label{sec:sub:explicit_policy_shaping}

Implicit policies can encode multi-modal feedback data but typically require longer training and inference times than explicit policies \cite{2023_diffusionpolicy, 2022_implicit_BC}.
Therefore, we introduce a simplified version of CLIC with the assumption that the policy follows a Gaussian distribution with fixed covariance, $\pi_{\bm \theta}(\bm a | \bm s) \sim\mathcal{N} (\mu_{\bm \theta}(\bm s), \bm \Sigma)$.
Under this assumption, the task is considered unimodal, and it can be represented using explicit models. In such cases, pointwise BC methods with absolute corrections are more robust to overfitting than in the implicit case, as multiple actions for the same state are averaged. In contrast, methods based on relative corrections are still prone to instability due to feedback becoming inconsistent as training progresses. We therefore focus on the latter in this section.

Given a corrective feedback $(\bm a^r, \bm a^h)$ at state $\bm s$,
% instead of increasing the probability of $\pi_{\bm \theta}$ selecting actions within $\hat{\mathcal{A}}(\bm{a}^r, \bm{a}^h)$
we enforce that the policy assigns more probability mass to the desired action set than to its complement:
\begin{equation}
    \pi_{\bm \theta}(\bm a \in \hat{\mathcal{A}}{(\bm a^r, \bm a^h)}| \bm{s}) \geq \pi_{\bm \theta}(\bm a \notin \hat{\mathcal{A}}{(\bm a^r, \bm a^h)}| \bm{s}).
    \label{eq:policy_improvement_distribution}
\end{equation}
%Explicit policies have been extensively used for absolute corrections in unimodal tasks \cite{2019_HG_DAgger, 2011_DAgger, 2023_Survey_LfD}. Here, we focus on relative corrections, which are less studied in this setting.
In this case,  Eq.~\eqref{eq:policy_improvement_distribution} can be satisfied by enforcing this simplified inequality (See Appendix \ref{apppendix:reduce_policy_improvemnt_inequality_Gaussian_assumption}):
\begin{align} 
    \pi_{\bm \theta}(\bm a^{-}_i|\bm s) \leq \pi_{\bm \theta}(\bm a^{+}_i|\bm s), i = 0, \dots, N_I,
    \label{eq:policy_improvement}
\end{align}
where $(\bm a^{-}_i, \bm a^{+}_i)$ are the contrastive action pairs obtained from $(\bm a^r, \bm a^h)$, as defined in Section~\ref{sec:Desired_action_space}.  
Intuitively, each action $\bm a^{+}_i$ should be assigned higher likelihood than its contrastive counterpart $\bm a^{-}_i$.
We enforce Eq.~\eqref{eq:policy_improvement} using the hinge loss:
\begin{align*}
\ell (\bm\theta ) = \!\! \!\!\!\! \underset{(\bm{a}^{h}, \bm{a}^{r}, \bm{s}) \sim p_{\mathcal D}}{\mathbb{E}} \sum_{i = 0}^{N_I} \max (0,  \log \pi_{\bm \theta}(\bm a^{-}_i|\bm s) - \log\pi_{\bm \theta}(\bm a^{+}_i|\bm s)  ). 
\label{eq:loss_policy_improvement}
\end{align*}
If Eq.~\eqref{eq:policy_improvement} is satisfied, the mean of the Gaussian lies inside the polytope desired action set defined by the correction, making the loss zero. Otherwise, the loss is used to update the parameters $\bm \theta$. We refer to this approach as \emph{CLIC-Explicit}.

\textbf{Connections to Learning from Preference Feedback}: 
Our set-supervision framework recovers standard pairwise preference objectives as a limiting case under the Gaussian policy assumption. In particular, when $\alpha = 180^\circ$ and $\varepsilon = 0.5$, the augmented contrastive pairs collapse to the original observed pair $(\bm a^r, \bm a^h)$, and the polytope desired action set reduces to a half-space. In this case, the CLIC-Explicit objective becomes the standard pairwise ranking loss:
\begin{align*}
\ell (\bm\theta ) =\!\! \underset{(\bm{a}^{h}, \bm{a}^{r}, \bm{s}) \sim p_{\mathcal D}}{\mathbb{E}} \max (0,  \log \pi_{\bm \theta}(\bm a^{r}|\bm s) - \log\pi_{\bm \theta}(\bm a^{h}|\bm s)  ), 
\end{align*}
% This form is closely related to losses used in Sequence Likelihood Calibration (SLiC) \cite{zhao2022calibrating, 2023_slic}, Direct Preference-based Policy optimization \cite{2023direct_preference_policy_optimization}, and  Contrastive Preference Learning \cite{2023_Contrastive_Prefence_Learning, 2025_PPL_intervention_preference}.
which is closely related to losses used in SLiC~\cite{zhao2022calibrating, 2023_slic} and direct preference-based policy optimization methods~\cite{2023direct_preference_policy_optimization, 2025_PPL_intervention_preference, 2023_Contrastive_Prefence_Learning}.
For Gaussian policies, such pairwise preference constraints directly shape the mean and thus provide a direct guarantee for Eq.~\eqref{eq:policy_improvement_distribution}. 
% Beyond this special case, CLIC-Explicit is not limited to pairwise ranking: it uses data augmentation to explicitly approximate \emph{polytope} desired action sets and optimizes the corresponding \emph{set-level} objective in Eq.~\eqref{eq:policy_improvement_distribution}. This distinction is consequential outside explicit Gaussian policies. 
However, this does not extend to expressive implicit policies: for EBMs or diffusion models, satisfying pairwise comparisons does not ensure that probability mass concentrates within the desired action set; we illustrate this failure mode in Section~\ref{sec:exp:toy_exp}.
% Tips: for each results section, when you write each paragraph, try to follow this:
% (1) motivation, why are you doing this?
% (2) preset result (pure data)
% (3) give interpretation

\begin{figure}[t]
	\centering
	\includegraphics[width=0.49\textwidth]{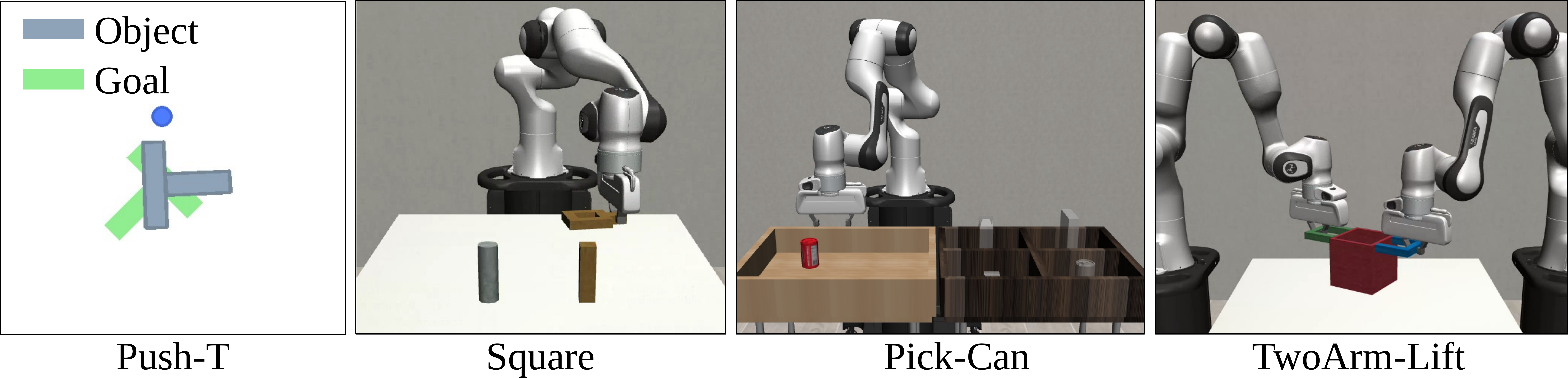}
    % \includesvg[width=0.49\textwidth, inkscapelatex=false]{figs/Fig5_exp_platforms.svg} 
	\caption{Tasks for the simulation experiments. Each task is tested with various feedback types, including accurate demonstrations, noisy demonstrations, and relative corrective feedback. For the TwoArm-Lift task, partial feedback is also tested by applying feedback only to one of the robots.}
	\label{fig:tasks}
\end{figure}

\begin{table*}[t!]
\footnotesize
\setlength\tabcolsep{ 5.2 pt}
\caption{Experimental results in simulation under accurate demonstration feedback. SR indicates the success rate, and CT represents the convergence timestep ($\times 10^3$). A ‘$\diagdown$’ symbol denotes that the algorithm did not converge. 
% For calculating CT, $\diagdown$ entries are replaced with the maximum allowable timestep.
}
\label{tab:sim_exp_accurate}
\begin{tabular}{lcccccccccc|cccc}
\Xhline{0.75pt}
Method & \multicolumn{2}{c}{CLIC-Half (ours) } &  \multicolumn{2}{c}{CLIC-Circular (ours)} & \multicolumn{2}{c}{Diffusion Policy} & \multicolumn{2}{c}{Implicit BC} & \multicolumn{2}{c}{PVP} & \multicolumn{2}{c}{CLIC-Explicit (ours)} & \multicolumn{2}{c}{HG-DAgger} \\
 & SR & CT & SR & CT & SR & CT & SR & CT & SR & CT &SR & CT & SR & CT \\ \hline
 %  \textbf{Accurate data}  && & & & & & & & & & && & \\
Push-T & 0.931 & 25.6  & \textbf{0.955} & 28.3 & 0.915 & 24.1 & 0.890 & 31.8 & 0.440 & 28.5 &  0.765 & 35.6 & 0.710 & 38.6  \\
Square & 0.930 & 43.0 & \textbf{0.960} & 55.5 & 0.953 & 48.4 & 0.732 & 54.6 &  0.000 & $\diagdown$   &0.634 & 65.9 & 0.420 & 67.0 \\
Pick-Can & 0.983 & 37.8 & \textbf{0.990} &  38.4 & 0.980 & 36.1 & 0.688 & 44.2 &  0.000 & $\diagdown$    & 0.995 & 42.5 & 0.990 & 35.7 \\
TwoArm-Lift & 0.970 & 18.5 &  \textbf{0.990} & 12.6 & \textbf{0.990} & 23.0 & 0.000 & $\diagdown$ &  0.000 & $\diagdown$    & 0.932 & 14.7 & 0.982 & 14.9  \\
\hline
Average & {0.954} & 31.2 & \textbf{0.974} & 33.7 &  0.960 & 32.9 & 0.578  & $\diagdown$  & 0.066 & $\diagdown$ & 0.836 & 39.7 & 0.776 & 39.1 \\
\Xhline{0.75pt}
\end{tabular}
\end{table*}

% \vspace{-20pt}

\section{Experiments}
\label{sec:experiments}

We demonstrate the effectiveness of CLIC through a series of simulations and real-world experiments. In Section \ref{sec:exp:simulation}, we compare CLIC with state-of-the-art methods under various types of feedback. 
% in the robot action space.
Section \ref{sec:exp:ablation} presents an ablation study, analyzing the impact of key parameters and design choices. Section~\ref{sec:exp:toy_exp} uses a 2D toy problem to visualize how different supervision objectives shape the learned energy landscape under noisy feedback. 
Finally, Section \ref{sec:exp:real_rotbo} validates the performance of CLIC in real-robot experiments.

\subsection{Simulation Experiments}
\label{sec:exp:simulation}
\textbf{Baselines} We compare CLIC with multiple baselines.
For explicit policies, we consider HG-DAgger \cite{2019_HG_DAgger} and D-COACH \cite{2019_Rodrigo_D_COACH}, which are IIL algorithms that learn from demonstrations and relative corrections, respectively. These two methods are refined for better performance, as reported in Appendix \ref{appendix:baselines}.
For implicit policies, the baselines include IBC \cite{2022_implicit_BC}, PVP \cite{2023_NIPS_PVP},  Diffusion Policy (DP) \cite{2023_diffusionpolicy}, and Ambient DP (ADP)  \cite{2023ambient}. 
As IBC, DP, and ADP are originally offline IL methods, we adapt them to the IIL framework for fair comparisons. 
Within this IIL framework, all methods share the same structure, differing only in how feedback is converted into supervision and used to update the policy.

Among these baselines, IBC, DP, ADP, HG-DAgger, and D-COACH primarily rely on pointwise supervision, while PVP uses a pairwise comparison objective by assigning high probability mass to human actions $\bm a^h$ and low probability mass to robot actions $\bm a^r$.
This pairwise structure is closely related to direct preference-based policy optimization methods \cite{2025_PPL_intervention_preference, 2023_Contrastive_Prefence_Learning}, differing mainly in policy parameterization (implicit EBM-based representation versus explicit Gaussian policies).
Since translating direct preference learning into EBMs results in a formulation closely related to PVP, we treat PVP as the representative baseline for pairwise-comparison-based objectives.
IBC, detailed in Section \ref{sec:Preliminaries}, and PVP are closely related to our method because they both train EBMs. 
DP is a counterpart to IBC when learning from demonstrations. It outperforms IBC because of the improved training stability offered by diffusion models. 
Based on Ambient Diffusion \cite{2023ambient}, ADP extends DP to learn clean action distributions from highly-corrupted action data.

\textbf{Tasks and metrics} We compared these methods across four simulated tasks, including a Push-T task introduced in \cite{2023_diffusionpolicy} and three manipulation tasks from the robosuite benchmark \citep{2020_robosuite} (see Fig. \ref{fig:tasks} and Appendix \ref{appendix:simulated_experiments_task details}).
We use a velocity control scheme across all methods. Specifically, each method outputs a velocity command for the robot's end-effector and, if applicable, the gripper as the action at each time step.
In our IIL framework, the agent learns from a \textbf{simulated teacher} to ensure repeatability. 
This teacher, a scripted expert policy, provides feedback every n=2 steps if the distance between its actions and the learner's exceeds a 0.2 threshold.
 Each method was run for 160 episodes in every experiment, with the procedure repeated 3 times to calculate average final success rates and convergence time steps. 
Specifically, for each individual experiment, we calculated the final success rate by averaging the success rates of the last 8 episodes, with each episode's success determined by 10 policy evaluations.
We defined the convergence time step as the earliest time step when the success rate exceeded 90\% of the final success rate.

\begin{table}[t!]
\footnotesize
\caption{Various feedback in the action space}
\centering
\setlength\tabcolsep{ 2.5 pt}
% \begin{tabular}{@{}ll@{}}
\begin{tabularx}{0.485\textwidth}{@{}lX@{}}
\toprule
\textbf{Type of Feedback Data}      & \textbf{Definition} \\ \midrule
Accurate absolute correction              & \( \bm a^h = \bm a^* \) \\
Gaussian noise             & \( \bm a^h = \bm a^* + \bm \omega, \bm \omega \sim \mathcal{N}(\bm0, \lambda||\bm a^* - \bm a^r||^2) \) \\
Partial feedback           & \( \bm a^h \in \{[\bm a^*_{r1}, \bm a_{r2}], [\bm a_{r1}, \bm a_{r2}^*]\} \) \\ 
\hline
Accurate relative correction              & \(\bm  a^h = \bm a^r + e\bm h^*, \bm h^*= \frac{\bm a^* - \bm a^r}{||\bm a^* - \bm a^r||}  \) \\
Direction noise            & \( \bm a^h \!= \bm a^r + e \bm h_r \), \!\( \angle (\bm h_r, \bm h^*) = \beta \in [0, 90^\circ\!) \!\) \\
\bottomrule
\end{tabularx}
\label{tab:feedback-definitions}
\end{table}

\textbf{Feedback types}
In addition to accurate absolute and relative corrections, Table \ref{tab:feedback-definitions} summarizes other common types of human feedback. These feedback types are also utilized in the simulation experiments. Here, $\bm a^*$ is defined as the original action taken by the simulated teacher. Partial feedback is utilized in the TwoArm-Lift task, where $\bm a_{r\,i}, i\in\{1, 2\}$ denotes each robot's action, and $\bm a_{ri}^*$ denotes its optimal action.

\begin{table*}[t!]
\footnotesize
\caption{Simulation results under noisy demonstration data. SR: success rate, CT: convergence timestep ($\times 10^3$). }
\label{tab:sim_exp_noise}
\setlength\tabcolsep{ 5.2 pt}
\begin{tabular}{lcccccccccccc|cccc}
\Xhline{0.75pt}
Method & \multicolumn{2}{c}{CLIC-Half  } & \multicolumn{2}{c}{CLIC-Circular } & \multicolumn{2}{c}{Diffusion Policy} & \multicolumn{2}{c}{Ambient DP} & \multicolumn{2}{c}{Implicit BC} & \multicolumn{2}{c}{PVP} & \multicolumn{2}{c}{CLIC-Explicit} & \multicolumn{2}{c}{HG-DAgger} \\
& SR & CT & SR & CT & SR & CT & SR & CT &SR & CT & SR & CT & SR & CT &SR & CT \\
%  \hline  \textbf{Accurate data}  && & & & & & & & && & \\
% PushT & \textbf{0.931} & 25.6 & 0.915 & 24.1 & 0.890 & 31.8 & 0.440 & 28.5 &  0.765 & 35.6 & 0.710 & 38.6  \\
% Square & 0.930 & 43.0 & \textbf{0.953} & 48.4 & 0.732 & 54.6 &  0.000 & $\diagdown$   &0.634 & 65.9 & 0.420 & 67.0 \\
% PickCan & 0.983 & 37.8 & 0.963 & 36.1 & 0.688 & 44.2 &  0.000 & $\diagdown$    & 0.995 & 42.5 & 0.990 & 35.7 \\
% TwoArmLift & 0.970 & 34.9 & 0.990 & 23.0 & 0.000 & 2.1 &  0.000 & $\diagdown$    & 0.902 & 14.0 & 0.982 & 14.9  \\
\hline
\textbf{Gaussian } && & & & & & & & & & && & & &\\ 
Push-T & 0.880 & 35.6 &  \textbf{0.960} & 29.2 & 0.893 & 49.0 &  0.840 &  54.0  & 0.735 & 42.1 & 0.155 & 45.8 & 0.663 & 41.4 & 0.598 & 41.0 \\
Square & \textbf{0.925} & 64.5 & 0.855 & 63.9 & 0.000 &  $\diagdown$  &  0.482 & 58.6 & 0.000 &  $\diagdown$ &   0.000 & $\diagdown$    & 0.238 & 71.0 & 0.060 & 77.2 \\
Pick-Can & 0.973 & 37.8 & \textbf{1.000} & 42.8 & 0.467 & 68.2  &  0.950 &  34.8  & 0.070 & 70.4 &   0.000 & $\diagdown$    & 0.800 & 69.5 & 0.028 & 23.1 \\
TwoArm-Lift & 0.847 & 47.0 & \textbf{0.945} &  19.2& 0.000 & $\diagdown$  &  0.907 & 19.3  &   0.000 & $\diagdown$    &  0.000 & $\diagdown$    & 0.433 & 39.3 & 0.008 & 63.8 \\
\hline
% \textbf{Partial feedback } && & & & & & & & && & \\
% TwoArmLift & \textbf{0.990} & 26.9 & 0.897 & 29.7 & 0.000 & $\diagdown$  & 0.000 & $\diagdown$ & 0.780 & 19.1 & 0.687 & 25.7 \\
% \hline
% \textbf{Relative correction} && & & & & & & & && &  \\
% PushT & \textbf{0.853} & 40.8 & 0.060 & 72.0 & 0.400 & 58.8 &   0.110	& 50.4   & 0.733 & 43.5 & 0.520 & 49.0 \\
% Square & \textbf{0.817} & 69.0 & 0.000 & 2.1 & 0.005 & 56.3 &   0.000 & $\diagdown$    & 0.065 & 66.1 & 0.243 & 79.7 \\
% PickCan & 0.870 & 40.8 & 0.000 & 2.0 & 0.310 & 81.7 &   0.000 & $\diagdown$    & \textbf{0.890} & 67.2 & 0.693 & 62.8 \\
% TwoArmLift & \textbf{0.860} & 31.1    &  0.000 & $\diagdown$   & 0.000  & $\diagdown$  &   0.000 & $\diagdown$   & 0.613 & 18.9 & 0.115 & 64.7 \\
% \hline
% \hline
Average & 0.906 & 46.2 & \textbf{0.933} &  42.0 &  0.340 & $\diagdown$  &  0.795 & 41.8   & 0.268  & $\diagdown$  & 0.039 & $\diagdown$ & 0.566 & 53.3 & 0.172& 35.7 \\
\hline
\textbf{Direction }   && & & & & & & & & & && & &   &     \\
Push-T & 0.700 & 48.1 &  \textbf{0.950} & 27.3  & 0.187 & 67.9  & 0.737  &  50.3   & 0.574 & 55.5 & 0.000  &  $\diagdown$ & 0.638 & 44.6 & 0.473 & 43.6 \\
Square & 0.870 & 63.6 &  \textbf{0.910} & 58.9 & 0.125 & 75.8  &  0.310 & 66.2  & 0.230 & 70.4 & 0.000  &  $\diagdown$ & 0.161 & 66.9 & 0.128 & 71.6 \\
Pick-Can & \textbf{1.000} & 43.1 & \textbf{1.000} &  39.0 &   0.000  & $\diagdown$   &  0.850 & 38.9    & 0.482 & 81.4 & 0.000  & $\diagdown$  & 0.867 & 55.4 & 0.342 & 72.0 \\
TwoArm-Lift & 0.965 & 18.7 & \textbf{0.980} & 16.5  & 0.885 & 46.6  &  0.957 & 17.4   &   0.000 & $\diagdown$   &   0.000  & $\diagdown$   & 0.807 & 21.0 & 0.157 & 31.4 \\
\hline
Average & 0.884 & 43.4 &    \textbf{0.960} & 35.4 & 0.399  & $\diagdown$  & 0.714  &  43.2   & 0.429 & $\diagdown$ & 0.000 & $\diagdown$ &  0.618& 47.0 & 0.275 &  54.7 \\
\Xhline{0.75pt}
\end{tabular}
\end{table*}

\subsubsection{Experiments with accurate feedback}
\label{sec:exp:accurate_feedback}

Table \ref{tab:sim_exp_accurate} shows the results when the teacher's absolute corrective feedback has no noise. 

\textbf{CLIC remains competitive with pointwise BC under accurate feedback}
The results in Table \ref{tab:sim_exp_accurate} show that CLIC-Half and CLIC-Circular both perform strongly under accurate feedback, consistently outperforming IBC and remaining competitive with DP; moreover, CLIC-Explicit outperforms HG-DAgger on the uni-modal tasks.
This result is notable because pointwise BC objectives remain valid with accurate demonstrations, yet CLIC's set-valued supervision does not suffer a performance disadvantage in this setting.
% Furthermore, IBC performance decreases as the action dimension of the task increases, with zero success rate in the TwoArm-Lift task. 
% CLIC-Half and CLIC-Circuler remain stable in this case and we 
Among the CLIC variants, CLIC-Circular performs best under accurate absolute feedback. This is expected, since its circular desired action sets provide more specific supervision than the polytope sets used by CLIC-Half when the teacher action is accurate. 
These results show that desired action sets can provide sufficiently informative supervision even when exact action targets are available.

\begin{table*}
\footnotesize
\setlength\tabcolsep{ 5.2 pt}
\caption{Simulation results under partial and relative feedback data. SR: success rate, CT: convergence timestep ($\times 10^3$). 
% A ‘$\diagdown$’ symbol denotes that the algorithm did not converge. For calculating CT, $\diagdown$ entries are replaced with the maximum allowable timestep.
}
\label{tab:sim_exp_relative_partial}
\begin{tabular}{lcccccccccccc|cccc}
\Xhline{0.75pt}
Method & \multicolumn{2}{c}{CLIC-Half } & \multicolumn{2}{c}{CLIC-Circular } & \multicolumn{2}{c}{Diffusion Policy} &\multicolumn{2}{c}{Ambient DP}& \multicolumn{2}{c}{Implicit BC} & \multicolumn{2}{c}{PVP} & \multicolumn{2}{c}{CLIC-Explicit} & \multicolumn{2}{c}{D-COACH} \\
 & SR & CT & SR & CT & SR & CT & SR & CT &SR & CT & SR & CT & SR & CT & SR & CT \\\hline
\textbf{Partial } && & & & & & & & && & & &\\
TwoArm-Lift & \textbf{0.990} & 26.9 & 0.920 &  17.8   & 0.897 & 29.7 &  \textbf{0.990} &  18.8 & 0.000 & $\diagdown$  & 0.000 & $\diagdown$ & 0.863 & 18.1 & 0.687 & 25.7 \\
\hline
\textbf{Relative } && & &  & & & & & & &   &    & && &  \\
Push-T & \textbf{0.853} & 40.8  & 0.000 & $\diagdown$    & 0.060 & 72.0  &   0.000 & $\diagdown$   & 0.400 & 58.8 &   0.110	& 50.4   & 0.733 & 43.5 & 0.520 & 49.0 \\
Square & \textbf{0.940} & 65.6 & 0.000 & $\diagdown$    & 0.000 & $\diagdown$  &   0.000 & $\diagdown$  & 0.005 & 56.3 &   0.000 & $\diagdown$    & 0.065 & 66.1 & 0.243 & 79.7 \\
Pick-Can & \textbf{0.983} & 41.9 &  0.000 & $\diagdown$    & 0.000 & $\diagdown$  &   0.000 & $\diagdown$   & 0.310 & 81.7 &   0.000 & $\diagdown$    & 0.890 & 67.2 & 0.693 & 62.8 \\
TwoArm-Lift & \textbf{0.955} & 25.3    &  0.000 & $\diagdown$    & 0.000 & $\diagdown$   &   0.000 & $\diagdown$    & 0.000  & $\diagdown$  &   0.000 & $\diagdown$   & 0.920 & 16.8 & 0.115 & 64.7 \\
\hline
Average & \textbf{0.933} & \textbf{43.4} &  0.000 & $\diagdown$    & 0.015 & $\diagdown$  &   0.000 & $\diagdown$   & 0.346  & $\diagdown$ & 0.066 & $\diagdown$ & 0.652 & 48.4 & 0.393 & 64.1 \\
\Xhline{0.75pt}
\end{tabular}
\end{table*}

\textbf{CLIC shapes implicit policies effectively, whereas pairwise comparison alone is too weak}
Table \ref{tab:sim_exp_accurate} shows a clear difference between set-valued and pairwise supervision when training expressive implicit policies. PVP, which applies pairwise comparison to an implicit EBM policy, performs poorly on the robosuite tasks, suggesting that pairwise comparison alone provides supervision that is too weak to effectively shape an implicit policy. In contrast, both CLIC-Half and CLIC-Circular achieve strong performance, indicating that desired action sets provide richer constraints over the action space and are therefore more suitable for training implicit policies. 

Meanwhile, CLIC-Explicit performs reasonably well with a simple Gaussian policy, which is consistent with the analysis in Section \ref{sec:sub:explicit_policy_shaping}: for explicit unimodal policies, pairwise comparison objective can already steer the policy mean toward the desired region. For implicit policies, however, it is often insufficient because it does not control how probability mass is distributed over the action space.
% This contrast is also reflected in the comparison between CLIC-Explicit and PVP. Although both objectives are defined only on $(\bm a^r, \bm a^h)$ and are closely related to pairwise comparison, CLIC-Explicit performs well with a simple Gaussian policy, whereas PVP fails with an implicit policy. This suggests that weak comparison-based supervision may be sufficient for simple explicit policies, but is often insufficient for shaping more expressive implicit policies. Overall, these results show that desired action sets provide a more informative supervision signal than pairwise comparison, especially when training implicit policies.

\subsubsection{Experiments with noisy demonstrations}
Noise is common when human teachers provide feedback to robots.
This can arise due to factors such as human fatigue or the limitations of teleoperation devices. 
To evaluate the ability of CLIC and baselines to learn from noisy feedback, we implemented two types of noise in simulation, as defined in Table \ref{tab:feedback-definitions}. 
For absolute corrections, we added Gaussian noise with $\lambda=0.5$ to the demonstrations. 
For relative corrections, the feedback is derived from absolute correction with a known magnitude. To introduce noise, we perturb the original direction signal by $\beta=45^\circ$ while maintaining its magnitude.

\textbf{CLIC remains robust while baselines degrade under noisy feedback}
The results in Table \ref{tab:sim_exp_noise} show that, as feedback transitions from accurate to noisy, CLIC-Half and CLIC-Circular experience much smaller performance drops compared to baselines like DP and IBC.
This can be observed by comparing Table \ref{tab:sim_exp_noise} with Table \ref{tab:sim_exp_accurate}.
 These baselines degrade because of their strict assumption of having accurate action labels.
 ADP outperforms DP but is worse than CLIC-Half and CLIC-Circular, as its assumption of linearly corrupted noise is violated here.
 % In comparison, CLIC allows adjusting the desired action set via hyperparameters, ensuring the optimal action remains within the desired space.
% This capability helps maintain robust performance under noisy conditions. 
In comparison, CLIC controls the desired action set through a small number of interpretable hyperparameters. In the noisy-feedback settings considered here, this flexibility allows the induced set to better match the uncertainty of the feedback signal and improves robustness to corrupted actions.

\subsubsection{Experiments with relative or partial feedback}
\label{sec:exp:simulation_relative_partial}  
When providing demonstrations is not possible, humans can provide feedback in more flexible ways.
One such scenario is partial feedback, where limitations in the control interface or a large action space make it challenging to provide complete demonstrations. We evaluated all methods on the TwoArm-Lift task, in which the teacher provides demonstrations to only one robot at a time.
Another scenario is relative correction, where the teacher provides an action that is an improvement over the robot's current action but is not necessarily optimal.
We tested this feedback type on four simulation tasks.

\textbf{CLIC-Half and CLIC-Explicit effectively learn from partial feedback}
The results are shown in Table \ref{tab:sim_exp_relative_partial}.
Here, human actions consist of  \textit{feedback dimensions} (where human feedback is provided) and  \textit{non-feedback dimensions} (which may be suboptimal).
While CLIC-Half maintains a high success rate, DP suffers from lower success rates and longer convergence times.  
This difference arises because the BC loss in DP attempts to imitate the entire teacher action, including suboptimal non-feedback dimensions.
In contrast, CLIC-Half imitates a desired action set rather than a single action label.
It focuses on improving actions on the feedback dimensions while leaving the non-feedback dimensions unconstrained. 
The same reasoning explains why CLIC-Explicit outperforms HG-DAgger.
ADP matches CLIC-Half because partial feedback aligns with its assumption of linear corruption.
On the other hand, CLIC-Circular's performance drops because its circular desired action set may fail to include the optimal action. 
This result highlights the importance of matching the desired-set construction to the feedback modality.

\textbf{CLIC-Half and CLIC-Explicit can learn from relative corrections}
Here, relative corrections indicate the direction of improvement from the robot action, but do not specify the correction magnitude or the distance to the optimal action.
CLIC-Half and CLIC-Explicit show only small performance drops compared to their results under absolute corrections in Table \ref{tab:sim_exp_accurate}.
However, all baselines based on the pointwise BC objective fail completely. 
This failure occurs because the BC loss can mislead policy updates, when the current policy’s actions are better than human actions in the dataset.
ADP also fails as all action dimensions are corrupted, violating its assumption. 
CLIC-Circular fails because its circular desired action set excludes the optimal action, in which case condition (A3) in Section~\ref{sec:overall_desiredA_space} does not hold.
In contrast, CLIC-Half and CLIC-Explicit construct polytope desired action sets, which provide less specific supervision than circular sets but remain compatible with the optimal action. As the policy improves, these sets do not conflict with improved actions and continue to provide useful supervision.
This result validates the ability of CLIC-Half and CLIC-Explicit to learn from relative corrections.
% This result shows that, for relative corrections, effective learning requires a set-valued formulation that preserves compatibility with the optimal action; polytope desired action sets satisfy this property, whereas circular sets may not.

\begin{figure}[t]
    \centering
    \includegraphics[width = 0.48\textwidth]{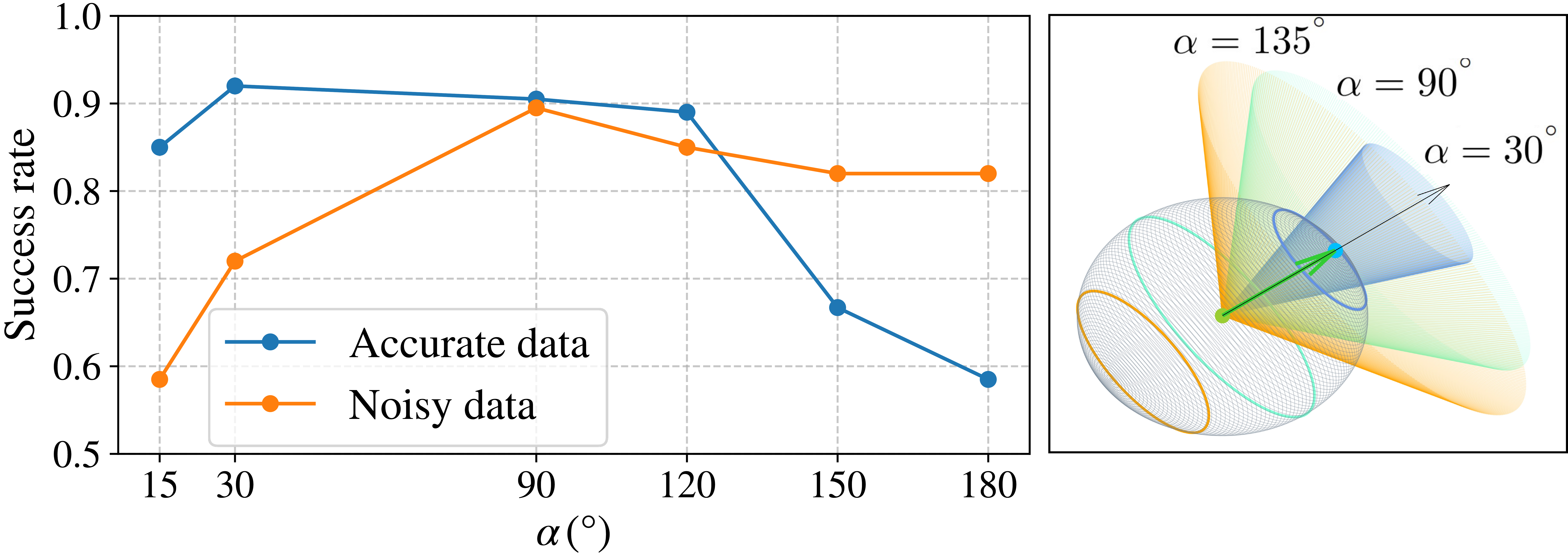}
    % \includesvg[width=0.49\textwidth, inkscapelatex=false]{figs/Fig10_effects_of_alpha.svg} 
	\caption{Hyperparameter analysis of the directional certainty parameter 
$\alpha$ for CLIC-Half. The right figure visualizes how different values of 
$\alpha$ adjust the desired action set in 3D.}
 \label{fig:Fig10_effects_of_alpha}
\end{figure}

\begin{figure}[t]
    \centering
    \includegraphics[width = 0.48\textwidth]{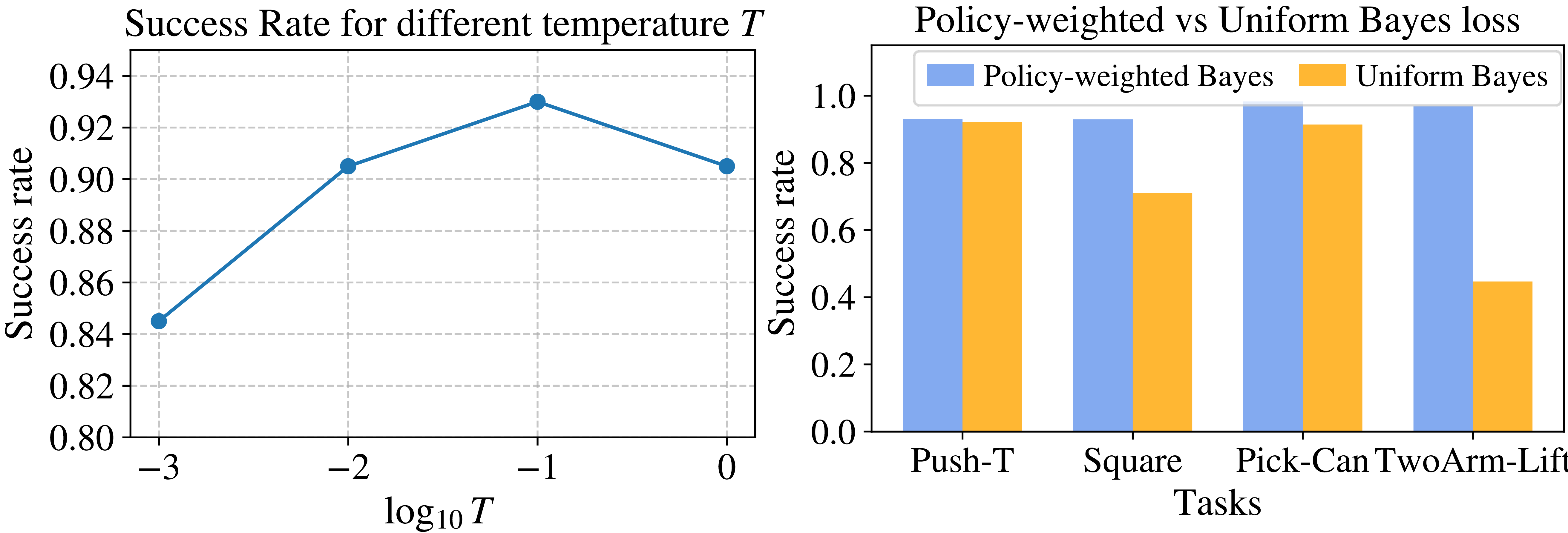}
    % \includesvg[width=0.49\textwidth, inkscapelatex=false]{figs/Fig13_exp_ablation_2.svg}
	\caption{Ablation study: (1) effects of the temperature parameter $T$. (2) Policy-weighted  vs uniform Bayes loss.}
 \label{fig:Fig13_exp_ablation_2}
\end{figure}

\subsection{Ablation Study}
\label{sec:exp:ablation}

\begin{figure*}[t]
	\centering
    \includegraphics[width=\textwidth]{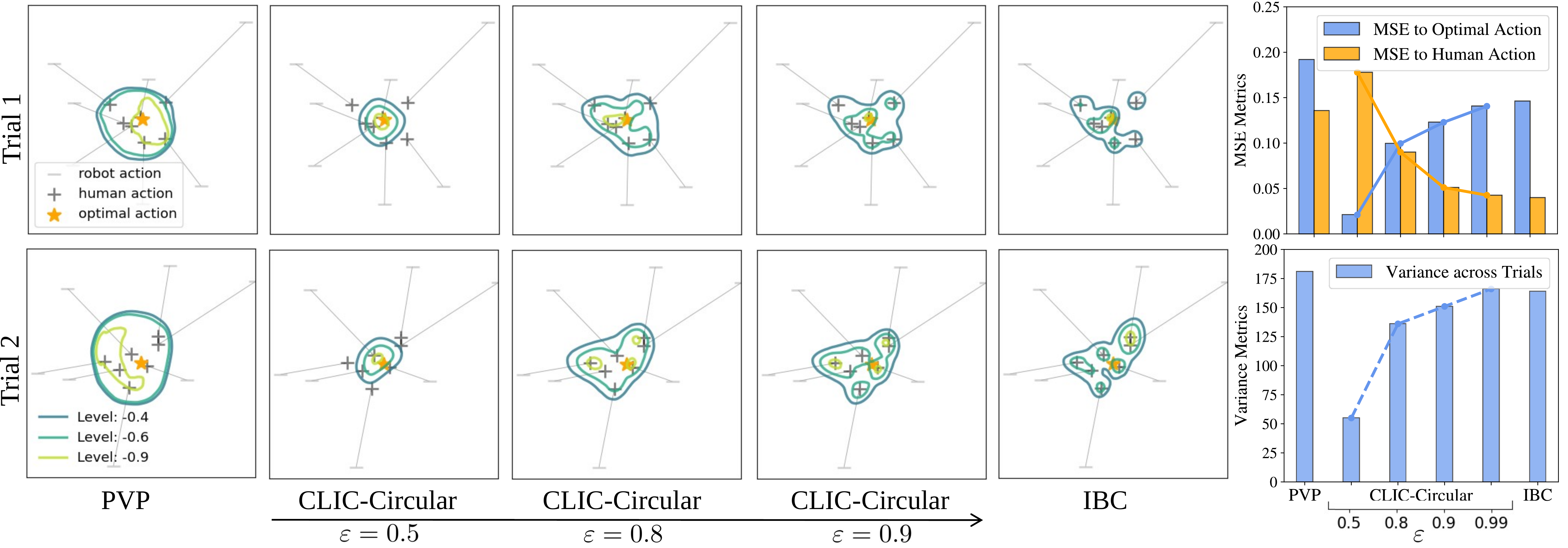}
    % \includesvg[width=\textwidth, inkscapelatex=false]{figs/Fig1_Illustration_generalization_across_state_3.svg} 
	\caption{ \textbf{Learned EBM landscapes across different trials}. The figure compares the energy landscapes learned by CLIC, PVP, and IBC after training in a 2D action space. Each row corresponds to the resulting EBMs of each trial. 
    In the middle part, we visualize the process of how CLIC-Circular reduces to IBC as $\varepsilon$ increases.
    CLIC-Circular ( with $\varepsilon=0.5$) effectively trains EBM across different trials, leading to consistent minima close to the true optimal action. In contrast, IBC overfits human actions and fails to estimate the true optimal action. Three evaluation metrics are shown in the right part of the figure.}
 \label{Fig1_Illustration_generalization_across_state}
\end{figure*}
We analyze the impact of various hyperparameters and loss design choices on the performance of CLIC: the directional certainty parameter $\alpha$, the temperature $T$ used in 
the observation model, and assumptions regarding the prior  $p(\bm{a}|\bm{s})$.
During these experiments, CLIC-Half is utilized.

\subsubsection{Effects of directional certainty $\alpha$}
The angle $\alpha$ controls the shape of the polytope desired action set.
We evaluate its effect in the Square task, with 
results reported in Fig. \ref{fig:Fig10_effects_of_alpha}.
For accurate feedback, the success rate decreases when $\alpha$ exceeds $120^\circ$. 
 This occurs because increasing $\alpha$ expands the desired action set to include more undesired actions, thereby providing less useful information for updating the policy.
For direction noise (noise angle $\beta=45^{\circ}$), the success rate decreases when $\alpha < 2 \beta = 90^\circ$. 
This is because when $\alpha < 2\beta$, direction noise can cause the polytope desired set to miss the optimal action, so the condition (A3) (Section~\ref{sec:overall_desiredA_space}) may not hold, leading to misleading updates.
Overall, CLIC performs well in a broad intermediate range of $\alpha$ that balances two effects: keeping the set informative while maintaining coverage of optimal actions under feedback noise.

\subsubsection{Effects of temperature $T$}
The temperature $T$ controls the sharpness of the observation model.
Four values of $T$ are tested in the Square task. 
The results are presented in the left side of Fig. \ref{fig:Fig13_exp_ablation_2}.
When $T$ is very small ($\log_{10} \!T \!\!=\! \!-3$), the success rate drops sharply. At this extreme, the observation model becomes binary (0/1), creating a sharp boundary that is difficult for the neural network to learn. Conversely, when $T$ is too large ($T = 1$), the success rate also declines. In this case, the probabilities of actions belonging to $\mathcal{A} {(\bm a^r, \bm a^h)}$ or not become nearly indistinguishable, offering limited information for policy improvement.
$T = 0.1$ proves to be a good balance and is selected across all experiments for CLIC-Half. 

\subsubsection{Policy-weighted Bayes Loss vs Uniform Bayes Loss}
We implement the uniform variant of CLIC and evaluate it with accurate demonstrations. The results, shown in the right side of Fig. \ref{fig:Fig13_exp_ablation_2}, demonstrate that the uniform Bayes loss leads to significantly poorer performance.
This highlights the importance of incremental policy updates. Since the desired action set may include some undesired actions, staying close to the current policy helps avoid imitating unintended behaviors, resulting in a more stable training process.

 \begin{figure*}[t]
    \centering
    \includegraphics[width = 1.0\textwidth]{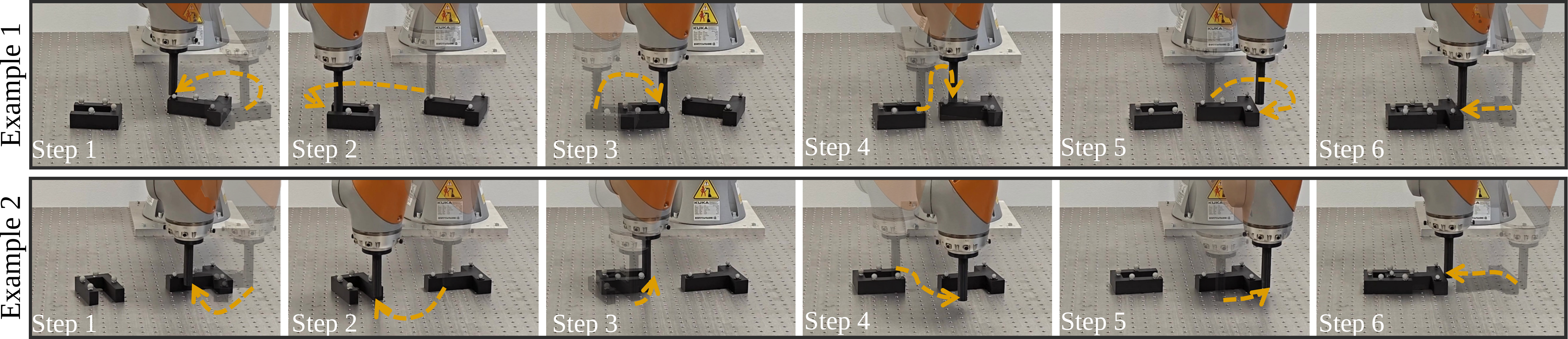}
    % \includesvg[width=1.0\textwidth, inkscapelatex=false]{figs/Fig7_InsertT_exp_results_traj_2.svg}
	\caption{Examples of CLIC-Half policy rollout for the Insert-T task after training. At each step, the transparent figure shows the initial state, and the orange arrow indicates the end-effector’s trajectory. The solid figure shows the end state.}
 \label{fig:Fig7_InsertT_exp_results_traj}
\end{figure*}

\subsection{Toy Experiments on Noisy Feedback}
\label{sec:exp:toy_exp}

Here, we present a toy task to illustrate the improved performance of CLIC over IBC and PVP. This task consists of a single constant state with a 2D action space, where the optimal action is $\bm{0}$ (see Fig. \ref{Fig1_Illustration_generalization_across_state}). 
The objective is to estimate the optimal action through multiple corrective feedback.
For each of 10 trials,
we generated a randomly sampled dataset consisting of 6 or 7 data points $(\bm s, \bm a^r, \bm a^h)$, with human actions drawn from a Gaussian distribution centered at the optimal action. 
Each method was trained offline for 1,000 steps, and we visualized the trained EBMs for the first two trials in Fig. \ref{Fig1_Illustration_generalization_across_state}.

To evaluate the methods, we introduced three metrics: (1) the mean square error (MSE) to optimal action: MSE between EBM's local minimum and the optimal action. (2) MSE to human action: the average MSE between each EBM's local minimum and its nearest human action. A smaller value indicates that the EBM is overfitting to the human action. (3) Variance across trials: the variance of the EBM values over the entire action space across ten trials.
These metrics are computed by averaging the results over the 10 trials and are reported in the right side of Fig. \ref{Fig1_Illustration_generalization_across_state}.

\begin{figure}[t]
    \centering
    \includegraphics[width = 0.48\textwidth]{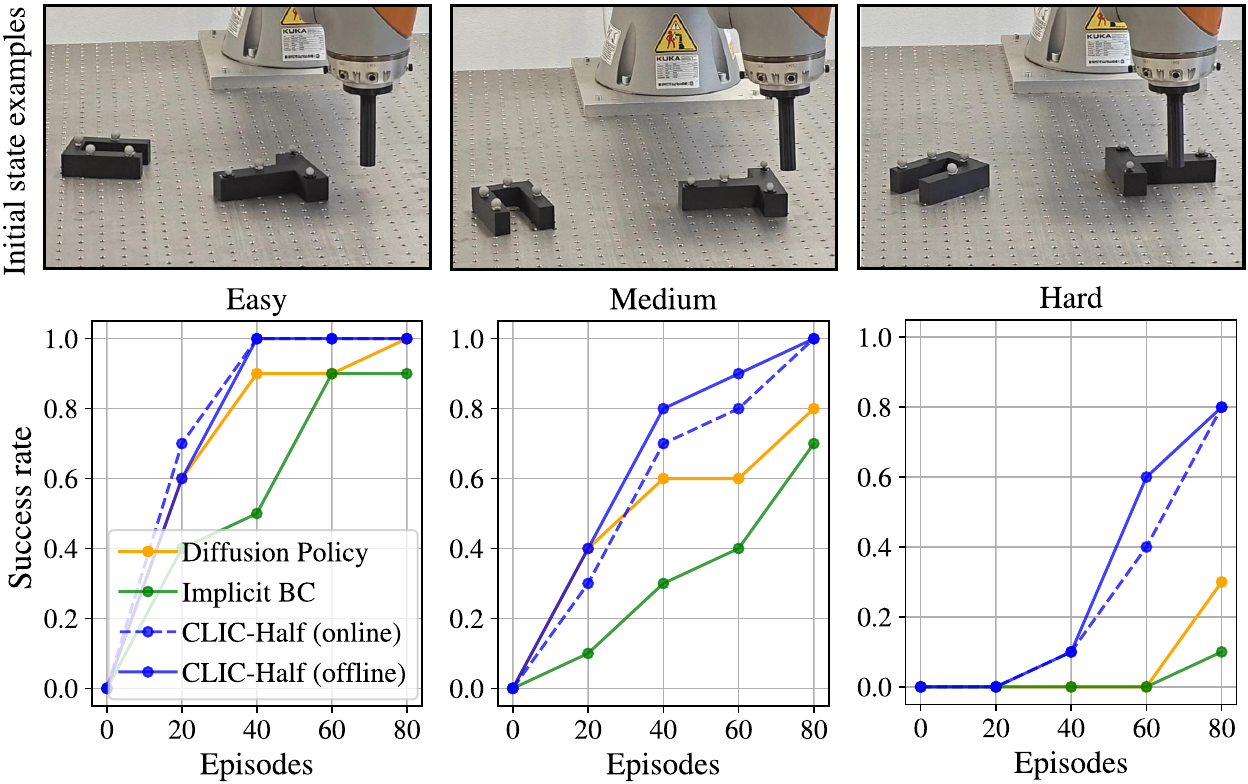}
    % \includesvg[width=0.49\textwidth, inkscapelatex=false]{figs/Fig7_InsertT_exp_results.svg}
	\caption{Experiment results for the Insert-T task, categorized by difficulty levels (easy, medium, and hard). Each column shows the performance metrics for a given difficulty level, along with examples of initial states for that level.
    ``CLIC-Half (offline)'' denotes results for CLIC-Half trained offline.}
 \label{fig:Fig7_InsertT_exp_results}
\end{figure}

\textbf{CLIC learns consistent EBM landscapes across different trials}
 The PVP-trained model assigns high probability mass to a large region of actions that are not present in the dataset, as shown on the left of Fig. \ref{Fig1_Illustration_generalization_across_state}.
 This behavior arises because its pairwise ranking loss constrains only relative ordering between observed human and robot action pairs, leaving the probability density of other actions weakly constrained. As a result, the learned landscape satisfies the pairwise comparisons while remaining ambiguous elsewhere.
 Conversely, IBC's loss function encourages the policy to match each human action label while discouraging all other actions, including those that are very similar to human actions. Consequently, the IBC-trained EBM overfits the data, creating minima at individual human actions. 
  This overfitting results in EBM landscapes with high variance across trials, as in the bottom-right of Fig. \ref{Fig1_Illustration_generalization_across_state}.
In contrast, CLIC learns from set-valued supervision rather than single action labels, producing consistent energy landscapes with minima close to the true optimal action and low variance across trials.
These results show that set-valued supervision yields more stable and accurate EBM landscapes than either pairwise comparisons or pointwise action targets, which helps explain the superior performance of CLIC over IBC and PVP in Section \ref{sec:exp:simulation}.
 % , as observed in Section \ref{sec:exp:accurate_feedback}.
 
\textbf{CLIC-Circular recovers the IBC in the limiting case}
We demonstrate that as the circular desired action set shrinks,  the behavior of CLIC-Circular reduces to that of IBC  (see the middle of Fig. \ref{Fig1_Illustration_generalization_across_state}).
By progressively increasing $\varepsilon$ to decrease the radius, the CLIC-trained EBM starts to split into several clusters.
This leads to overfitting, evidenced by a decrease in MSE to human actions and an increase in MSE to the optimal action, as shown in the right of Fig. \ref{Fig1_Illustration_generalization_across_state}.
% In the limit  $\varepsilon \rightarrow 1$,  the radius approaches zero, and the learned landscape closely resembles the one trained using IBC.
In the limit $\varepsilon \rightarrow 1$, the radius approaches zero, and CLIC-Circular effectively reduces to pointwise supervision, yielding a learned energy landscape similar to IBC.
This highlights the key distinction between the two methods: CLIC-Circular learns from set-valued action targets rather than pointwise action labels.
This distinction is crucial for stable learning under noisy feedback.

\subsection{Real-robot Validations}
\label{sec:exp:real_rotbo}

Here, we use three tasks to demonstrate the practical applicability of CLIC.
The experiments include a long-horizon multi-modal Insert-T task, a dynamic ball-catching task, and a water-pouring task that necessitates precise control of the robot's end effector position and orientation. 
For the Insert-T task, we employ CLIC-Half and compare its performance against IBC and DP.
For the ball-catching and water-pouring tasks, we use CLIC-Explicit because it performs well in uni-modal tasks, as demonstrated in Section \ref{sec:exp:simulation}, and is more time-efficient compared to CLIC with an EBM policy (Details in Appendix \ref{appendix:time_efficiency_comparision}.).
The experiments were carried out using a 7-DoF KUKA iiwa manipulator. When required, an underactuated robotic hand (1-dimensional action space) was attached to its end effector. A 6D space mouse was employed to provide feedback on the pose of the robot's end effector. Furthermore, in the ball-catching task, a keyboard provided feedback on the gripper's actuation. 
The setup of each task is detailed in Appendix \ref{appendix:real_robot_experiments_task details}, and the time durations used are reported in Appendix \ref{appendix:time_duration}.  
The experiment results are reported as follows:

\subsubsection{Insert-T—a comparison between state-of-the-art methods}
\label{sec:exp:real_rotbo:insert-T}
The Insert-T task requires the robot to insert a T-shaped object into a U-shaped object by pushing to adjust their positions and orientations.
Compared to the simulated Push-T task, Insert-T is more complex due to two factors: (1) it involves two objects, introducing multi-modal decisions about which object to manipulate first; and (2) it has an increased task horizon. 
We define three difficulty levels—easy ($<$1 contact changes), medium ($<$5), and hard ($\geq$5)—based on the required contact changes from a teacher policy.
Examples are shown in the upper part of Fig. \ref{fig:Fig7_InsertT_exp_results}.
During each evaluation, 10 different initial states are tested for each level; all methods are evaluated using the same set of initial states.
In the experiment, human-provided demonstration feedback is used to train CLIC-Half within the IIL framework. The collected data is also used to train baselines (DP and IBC) offline. For a fair comparison, CLIC is additionally trained offline on the same dataset as the baselines.

Results for different difficulty levels are shown in Fig. \ref{fig:Fig7_InsertT_exp_results}. For easy tasks, baseline methods perform similarly to CLIC but converge more slowly. For medium and hard tasks, CLIC achieves significantly higher success rates. This is particularly evident for hard tasks, where CLIC achieves 80$\%$ success compared to 30$\%$ for DP and 10$\%$ for IBC. 
The results demonstrate CLIC's ability to handle complex multi-modal tasks, thanks to the powerful encoding capabilities of EBMs and CLIC's stable EBM training. 
Furthermore, as the task difficulty increases, CLIC outperforms DP and IBC by a large margin. This suggests that, for training policies, using a desired action set is more robust and efficient in real-robot tasks than relying on a single action label.
 Examples of post-training policy rollouts for CLIC are shown in Fig. \ref{fig:Fig7_InsertT_exp_results_traj}.

Fig. \ref{fig:Fig7_InsertT_exp_results} also includes results for CLIC trained with offline data, showing a similar final success rate to the online version. This indicates that CLIC can be employed to learn from offline data as well.
While CLIC is primarily based on the IIL framework, the core ideas proposed here could also benefit offline methods. 
We believe exploring offline training of CLIC is a promising direction for future work.

\subsubsection{Ball catching—quick coordination and partial feedback}
 The ball-catching task is challenging because of its highly dynamic nature. 
This complexity makes it difficult to provide successful demonstrations to the robot, thus ruling out demonstration-based IIL methods for solving it\footnote{This limitation could be overcome with a highly reactive and precise teleoperation device. However, this also makes the solution more expensive.}.
Instead, relative corrective feedback is more intuitive and easier for this task as humans can provide direction signals occasionally to improve the robot's policy \cite{2020_DCOACH_temporal}.  
Besides, for a successful grasp, the robot must coordinate precisely the ball's motion, the end-effector's motion, along with the gripper's actuation.
This requirement makes it challenging to provide feedback on the complete action space at any given moment, and makes partial feedback suitable for this task.
With partial feedback, relative corrections can be independently provided for either the end-effector's motion or the gripper's actuation.

 \begin{figure}
    \centering
    \includegraphics[width = 0.455\textwidth]{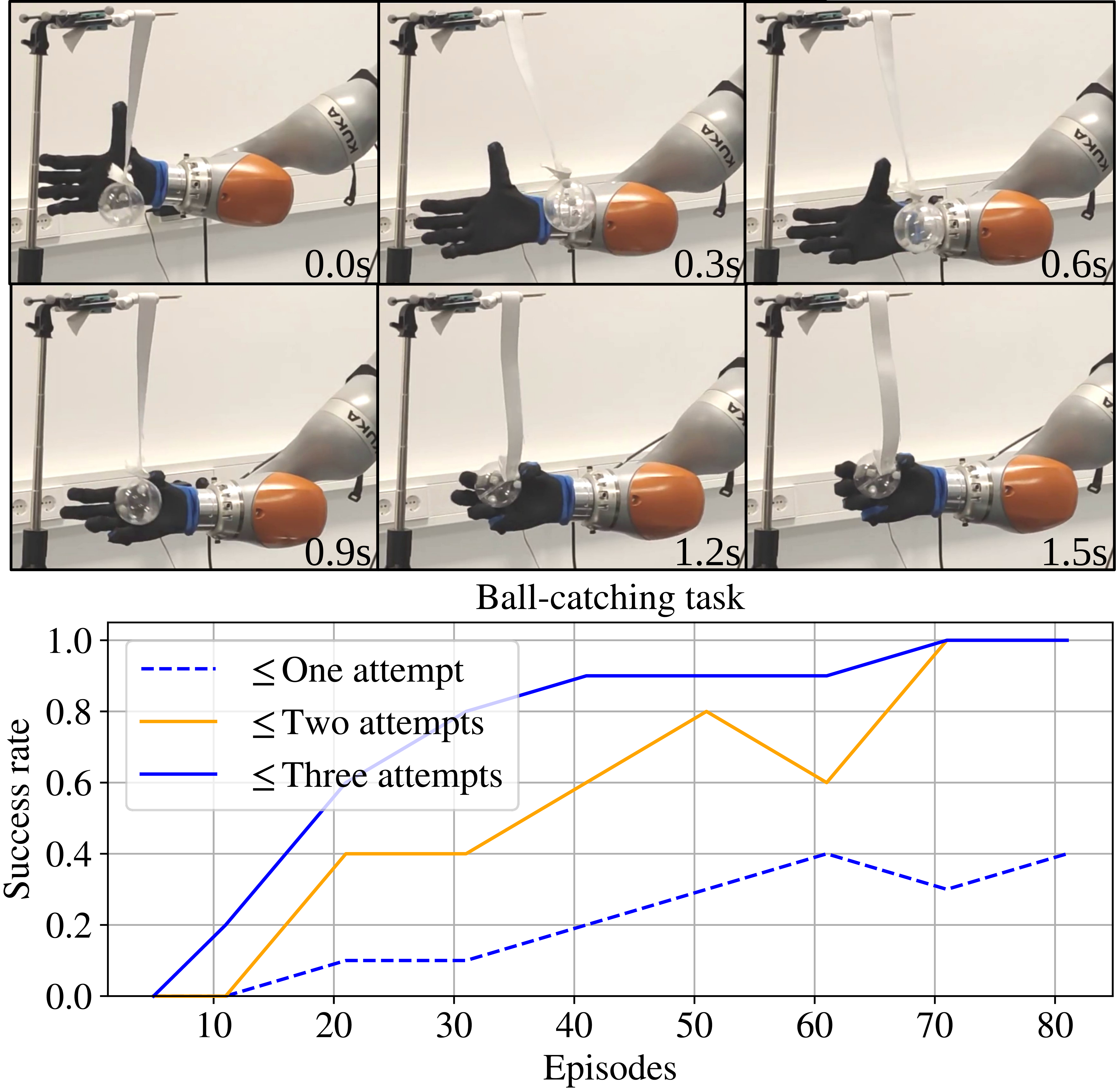}
    % \includesvg[width=0.49\textwidth, inkscapelatex=false]{figs/Fig8_CatchBall02.svg}
	\caption{Experiment results for the ball-catching task.}
 \label{fig:Fig8_CatchBall}
\end{figure}

\begin{figure}
    \centering
    \includegraphics[width = 0.45\textwidth]{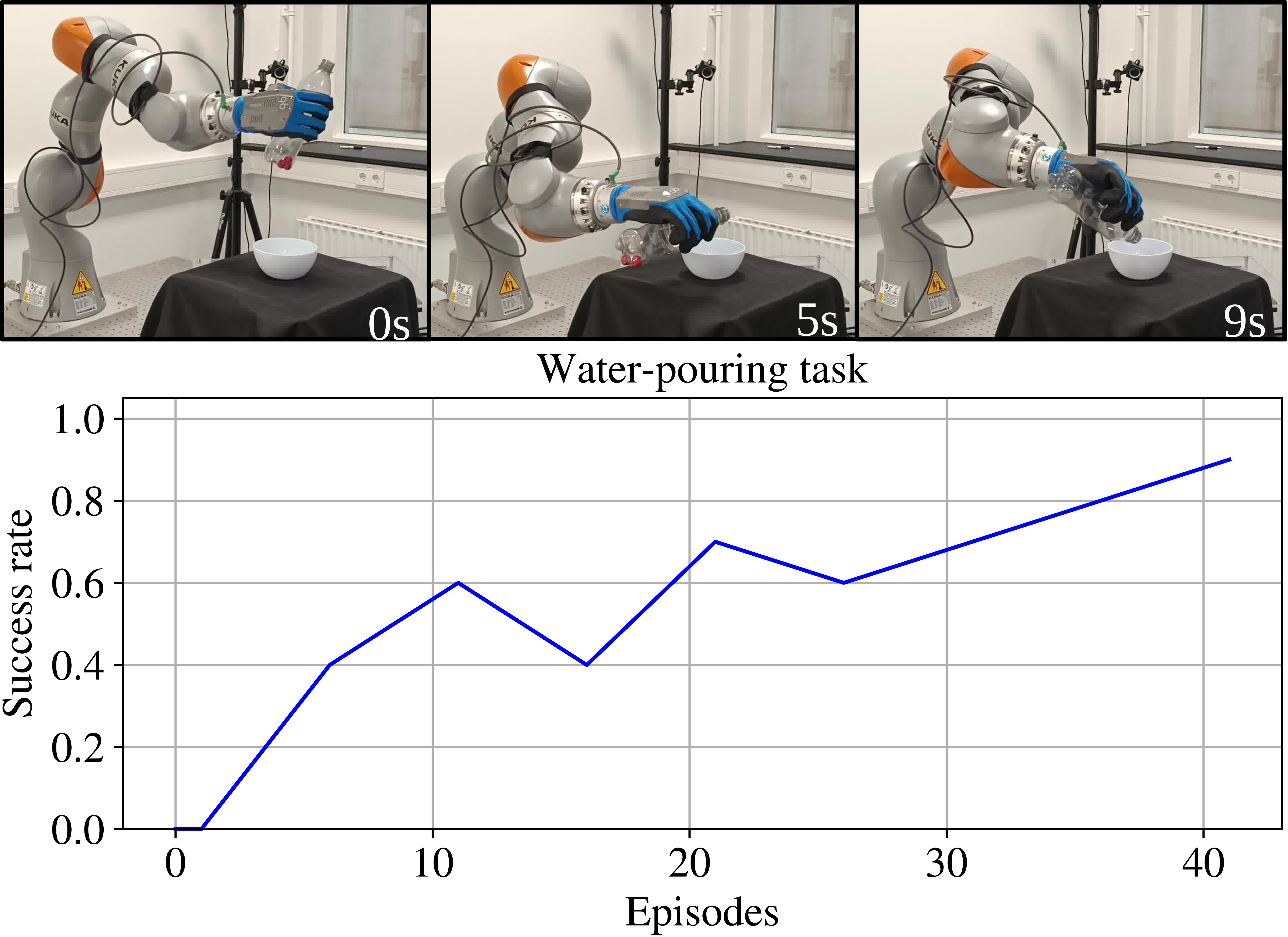}
    % \includesvg[width=0.49\textwidth, inkscapelatex=false]{figs/Fig9_waterpouring.svg}
	\caption{Experiment results for the water-pouring task.}
 \label{fig:Fig9_waterpouring}
\end{figure}

Fig. \ref{fig:Fig8_CatchBall} shows the experiment results of the ball-catching task, reporting the success rate of catching the ball within one, two, and three attempts. 
By the end of training, the robot achieves a 1.0 success rate for catching the ball within two attempts, and its first-attempt success rate continues to improve to 0.4.
One post-training policy rollout of a successful first-attempt catch is shown in Fig. \ref{fig:Fig8_CatchBall}, where the ball is caught within 1.5 seconds, an impressive result given the actuation delay of the robot hand.
This experiment demonstrates that CLIC can leverage both the relative corrective feedback and partial feedback effectively to learn challenging high-frequency tasks.

\subsubsection{Water pouring—learning full pose control with CLIC}
The water-pouring task requires the robot to control the pose of a bottle to precisely pour liquid (represented with marbles) into a small bowl. 
The human teacher can switch between absolute and relative corrections via keyboard. 
Initially, absolute feedback was preferred as the policy was learned from scratch, and it was easier to intervene in a 6D action space. As the policy improved, relative corrections made it easier to refine the policy in specific regions of the state space.

The experimental data is shown in Fig. \ref{fig:Fig9_waterpouring}. From Episode $1$ to Episode $16$, the teacher's feedback is provided in an absolute correction format. From Episode $16$ onward, the teacher's feedback is given as relative corrections to make small adjustments to the robot's policy. The success rate exhibits an overall improving trend, consistently increasing from 0.6 in Episode 26 to 0.9 by Episode 41. An example of the policy rollouts after training is illustrated in Fig. \ref{fig:Fig9_waterpouring}.
This experiment demonstrates the effectiveness of CLIC for learning precise control over position and orientation.

% \input{Discussion}
% \section{Limitations and Future work} 
% \label{sec:Limitation_future_work}
% Despite the demonstrated effectiveness of CLIC in simulations and real-world experiments, there are limitations that future work can improve.
% Firstly, our method shares a similar limitation with HG-DAgger in that it does not utilize data lacking corresponding human feedback. 
% The naive integration with approaches like those discussed in \cite{2022_Expert_Intervention_Learning}, \cite{2023_RSS_Robot_Learning_on_the_job} have been studied in Section~\ref{sec:experiments} but failed to result in significant improvement.
%  Secondly, as discussed in Section~\ref{sec:discussion}, our CLIC method fails to learn from multi-modal data because of the assumption of the Gaussian distribution policy.
% Thirdly, our method's reliance on learners' errors to provide 
%  corrective feedback poses a challenge: if the learner initially performs well, the method may not effectively enhance the robot's policy. To address this, evaluative feedback could be integrated to reinforce successful actions.
% Future research directions include 1) developing algorithms capable of utilizing data without human feedback; 2) creating algorithms for shaping policies suitable for multi-modal tasks; and 3) combining relative corrective feedback with evaluative or preference feedback to further improve the method's efficiency.

\section{Conclusion} 
\label{sec:conclusion}

% In this paper, we introduce CLIC, a novel approach to learning policies from interactive human corrections. 
% To achieve this, the desired action set concept and its probabilistic formulation are presented. 
% These are employed to design a novel loss function to align the robot's policy with desired action sets.

In this paper, we introduced set-valued supervision for learning from corrective feedback. Building on this perspective, we proposed CLIC, a practical interactive imitation learning framework that constructs desired action sets from human corrections and trains policies to align with these sets.
Across simulation and real-robot experiments, we showed that this formulation is effective across diverse feedback settings: it remains competitive under accurate feedback, is more robust under noisy, inconsistent, relative, and partial feedback, and supports multi-modal behaviors with implicit policies. These results show that set-valued supervision provides a useful middle ground between pointwise behavior cloning and pairwise comparison objectives for interactive policy learning.

Despite the demonstrated effectiveness of CLIC in simulations and real-world experiments, there are limitations that future work can improve.
Firstly, similar to HG-DAgger, our method does not utilize the robot's state-action data when the teacher provides no feedback \cite{2020_RSS_expert_interventio_learning}.
In future work, we plan to investigate how to incorporate these non-intervention data into CLIC. 
Secondly, 
CLIC relies on learner's errors to trigger corrective feedback. 
This can be a limitation because if the learner initially performs well, our method may not effectively improve the robot's policy. One potential solution is to incorporate evaluative feedback to reinforce learner's successful behaviors.
Beyond these limitations, several promising directions for future research include offline training of CLIC and applying the CLIC loss to train diffusion or flow-based models.

% Despite the demonstrated effectiveness of CLIC in simulations and real-world experiments, there are limitations that future work can improve.
% Firstly, our method shares a similar limitation with HG-DAgger in that it does not utilize data lacking corresponding human feedback. 
% The naive integration with approaches like those discussed in \cite{2022_Expert_Intervention_Learning}, \cite{2023_RSS_Robot_Learning_on_the_job} have been studied in Section~\ref{sec:experiments} but failed to result in significant improvement.
%  Secondly, as discussed in Section~\ref{sec:discussion}, our CLIC method fails to learn from multi-modal data because of the assumption of the Gaussian distribution policy.
% Thirdly, our method's reliance on learners' errors to provide 
%  corrective feedback poses a challenge: if the learner initially performs well, the method may not effectively enhance the robot's policy. To address this, evaluative feedback could be integrated to reinforce successful actions.
% Future research directions include 1) developing algorithms capable of utilizing data without human feedback; 2) creating algorithms for shaping policies suitable for multi-modal tasks; and 3) combining relative corrective feedback with evaluative or preference feedback to further improve the method's efficiency.

% \section*{Acknowledgments}

% \section*{Statements and Declaration}

% TODO

\section*{Author contributions}
Zhaoting Li led the method development, conducted the experiments, and wrote the manuscript. Rodrigo Pérez-Dattari contributed to the method design, experimental design, and manuscript revision. Cosimo Della Santina, Robert Babuska, and Jens Kober provided feedback throughout the project and contributed to revising the manuscript.

\section*{Declaration of conflicting interests}
The author(s) declared no potential conflicts of interest with respect to the research, authorship, and/or publication of this article.

\section*{Funding}
This project is made possible by a contribution from the National Growth Fund program NXTGEN Hightech.

%% Use plainnat to work nicely with natbib. 
\appendix
  
\section{Appendix}

% \subsection{Langevin MCMC for EBMs}
\subsection{Sampling actions from an EBM policy}
\label{Appendix:IBC}
To ensure the EBM learns an accurate data distribution, the negative samples should be close to the action label, avoiding overly obvious distinctions that hinder effective learning \cite{2020_flow_constrastive_estimation_EBM}. 
This can be achieved by generating negative samples from the current EBM using MCMC sampling with stochastic gradient Langevin dynamics \cite{2019_EBM_Du_Yilun, 2011_Langevin_dynamics}:
\begin{equation}
\tilde{\bm a}_j^i = \tilde{\bm a}_j^{i-1} - \lambda \nabla_{\bm a} E_\theta(\bm s, \tilde{\bm a}_j^{i-1}) + \sqrt{2\lambda}   \omega^i,    
\label{eq:mcmc_sampling}
\end{equation}
where $\{ \tilde{\bm a}_j^0 \}$ is initialized using the uniform distribution and $\omega^i$ is the standard normal distribution.
For each $\tilde{\bm a}_j^0 $, we run $N_{\text{MCMC}}$ steps of the MCMC chain, with $i = 0, \dots, N_{\text{MCMC}}$ denoting the step index. The step size $\lambda > 0$ can be adjusted using a polynomially decaying schedule.

During inference, the estimated optimal action \(\hat{\bm a}^*\) is obtained by minimizing the energy function, and can be approximated through Langevin MCMC:
\[
\hat{\bm a}^* = \underset{\bm a}{ \arg\min} E_\theta(\bm s, \bm a).
\]
 \subsection{Obtaining Contrastive Action Pairs from one Corrective Feedback}
\label{appendix:sub:CLIC_one_corrective_feedback}

\begin{figure}[t!]
	\centering
	\includegraphics[width=0.48\textwidth]{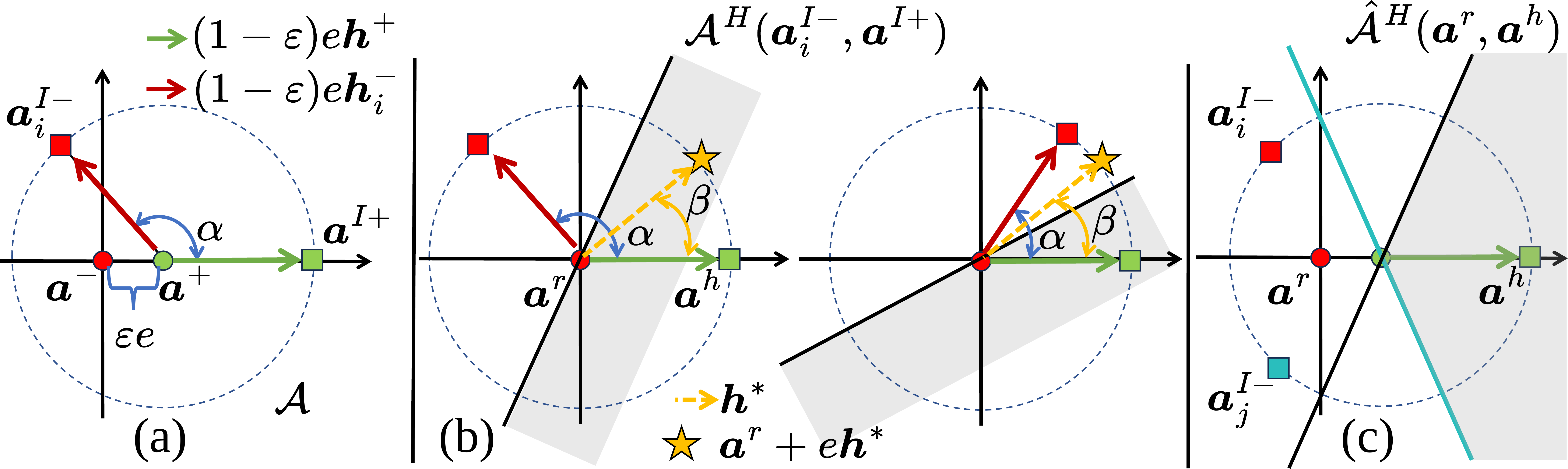}
	\caption{
 Illustration of polytope desired action set: (a) Generating contrastive action pairs from one corrective feedback $(\bm a^r, \bm a^h)$. 
 % Circles denote explicit actions, and
 Squares denote implicit information.
 (b) Examples of different $\alpha$ for the same $\bm h^*$. 
  % The action $\bm a+e\bm h^*$ is denoted by the orange star. 
  With $\varepsilon = 0$,  when $\alpha \geq 2 \beta$ (left), $\bm a+e\bm h^*$ is inside the desired action set. When $\alpha < 2 \beta$ (right), $\bm a+e\bm h^*$ is outside the desired action set. (c) Example of the intersection of the desired half-spaces.}
\label{fig:Fig8AndFig9_explain_feasible_space}
\end{figure}

Here, we detail how we generate contrastive action pairs $(\bm a^{-}, \bm a^{+})$ from one corrective feedback $(\bm a^r, \bm a^h)$. 
 For a given robot action $\bm a^r$ at state $\bm s$, the teacher provides the directional signal $\bm h$ to create the action pair  $(\bm a^r, \bm a^h)$ (Fig. \ref{fig:Fig8AndFig9_explain_feasible_space}a), where $\bm a^h = \bm a^r + e \bm h.$
We introduce the hyperparameter 
$\varepsilon \in [0, 1)$, which 
controls how much the robot action gets modified from $\bm h$.
Accordingly, we define one contrastive action pair
as $(\bm a^-, \bm a^+) = (\bm a^r, \bm a^r + \varepsilon e\bm h)$, which are denoted by the red and green circle in Fig. \ref{fig:Fig8AndFig9_explain_feasible_space}a, respectively.

Furthermore, the \emph{implicit information} of directional signal $\bm h$ is also utilized, as a way of data augmentation, to exclude some undesired actions in $\mathcal{A}^H{ (\bm a^r, \bm a^r \!\!+ \varepsilon e \bm h)}$. 
We define the positive correction as $\bm h^{+} \!\!\!=\! \bm h$. Additionally, we define the implicit negative correction $\bm h^{-}$ as a unit vector that points in a different direction from $\bm h^{+}$.
$\bm h^{-}$ is sampled from the set: 
% $\mathcal{H}^{-}$:
\begin{equation}
 \mathcal{H}^{-}(\bm h^{+}, \alpha) = \{ \bm h^{-} \in \mathcal{H} | \, \, \| \bm h^{-} \| = 1,  \angle(\bm h^{-}, \bm h^{+}) = \alpha \}, 
 \label{eq:assumption_negative_correction}
\end{equation}
where angle $\alpha \in (0^\circ, 180^\circ]$ is a hyperparameter that indicates how much certainty we have on the directional information provided by the human. 
% There are infinitely many possible $\bm h^{-}$ if the dimension of $\mathcal{A}$ is greater than 2. 
% In this case, we can sample $\bm h^{-}$ by generating a unit vector orthogonal to \(\bm h^{+}\) and combining it with \(\bm h^{+}\) at the specified angle \(\alpha\).
With sampled $\bm h^{-}_i \in \mathcal{H}^{-}$, we can obtain corresponding action pairs $(\bm a^{-}_i, \bm a^{+}_i), i=1,\dots, N_I$, where $N_I$ represents the total number of implicit actions.
% $N_I$ is set to 128
In Fig. \ref{fig:Fig8AndFig9_explain_feasible_space}a, the green arrow denotes $\bm h^{+}$ and the red arrow denotes one sample of $\bm h^{-}$.
The action pairs ${(\bm a^{-}_i, \bm a^{+})}$ are defined as follows, illustrated by the squares in Fig. \ref{fig:Fig8AndFig9_explain_feasible_space}a:
\begin{equation}
    \bm a^{+} = \bm a^h,
     \bm a^{-}_i = \bm a^{r} + \varepsilon e  \bm h^{+} + (1-\varepsilon) e \bm h^{-}_i.
     \label{eq:implicit_action_pairs}
\end{equation}
 One example of a polytope desired action set is shown in Fig. \ref{fig:Fig8AndFig9_explain_feasible_space}c.
% To exclude some undesired actions in $\mathcal{A}^H { (\bm a^r, \bm a^r + \varepsilon e \bm h)}$ while ensuring $\bm{a}^*$ remains within $\hat{\mathcal{A}}^H(\bm a^r, \bm a^h)$, 
The hyperparameters $\alpha$ and $\varepsilon$ control the geometry of the desired action set and should be selected to match the uncertainty of the feedback signal, as illustrated in Fig.~\ref{fig:Fig8AndFig9_explain_feasible_space}b.
The ablations in Section \ref{sec:exp:ablation} show that CLIC remains effective over a reasonably broad range around these choices.

\subsection{Proof for the Convergence of Overall Desired Action Set for the Unimodal Case}
\label{appendix:proof_convergence_DesiredA}
% Proposition \ref{proposition:convergence}:
%    For any state $\bm s$, assume a trained policy $\pi_{\bm \theta}$ always select actions from $\hat{\mathcal{A}}^{\mathcal{D}_{\bm s}}_k$,
%    and there are a finite number of optimal actions.
%    Assume that there is a nonzero probability for the teacher to provide feedback to each suboptimal action $\bm a^r_{k+1} \sim \pi_{\bm \theta}(\cdot | \bm s)$. 
%     Then, as $k \rightarrow \infty$, $\hat{\mathcal{A}}^{\mathcal{D}_{\bm s}}_k$ converges to $\mathcal{A}^*(\bm s)$ or a subset of $\mathcal{A}^*(\bm s)$.

Here, we prove that given conditions (A1), (A2), and (A3), for any state $\bm s$, $ \lim_{k\rightarrow \infty} \mathcal{A}^{\mathcal{D}_{\bm s}}_k  \subset \mathcal{A}^*_{\bm s}.$

\begin{proof}
    For any state $\bm s$, the combined desired action set after receiving $k$ feedback is denoted as $\mathcal{A}^{\mathcal{D}_{\bm s}}_k$.
When feedback is received at iteration $k{+}1$, the set is updated by intersection:
\begin{align}
\mathcal{A}^{\mathcal{D}_{\bm s}}_{k+1}
=
\mathcal{A}^{\mathcal{D}_{\bm s}}_{k}
\cap
\hat{\mathcal{A}}(\bm a^r_{k+1}, \bm a^h_{k+1}),
\label{eq:update_intersection_new}
\end{align}
where $\bm a^r_{k+1} \sim \pi_{\bm\theta_k}(\cdot \mid \bm s)$ is the learner's action and
$\bm a^h_{k+1}$ is the teacher feedback.
Thus, 
\begin{align}
\mathcal{A}^{\mathcal{D}_{\bm s}}_{k+1} \subseteq \mathcal{A}^{\mathcal{D}_{\bm s}}_{k}, \quad \forall k,
\label{eq:nested_sets_new}
\end{align}
i.e., $\{\mathcal{A}^{\mathcal{D}_{\bm s}}_{k}\}_{k\ge 0}$ is a volume non-increasing sequence.

 By (A3), $\mathcal{A}^{\mathcal{D}_{\bm s}}_{k+1}  \neq  \varnothing$.  The combined desired action set remains non-empty and contains at least one optimal action.
%  We show that $\lim_{k\rightarrow \infty} \mathcal{A}^{\mathcal{D}_{\bm s}}_k  \subset \mathcal{A}^*_{\bm s}$ by contradiction.
%  Assume for contradiction that there exists a suboptimal action  $\bar{\bm{a}} \in  \mathcal{A}^{\mathcal{D}_{\bm s}}_k \setminus \mathcal{A}^*_{\bm s}$.
% Since $\bar{\bm{a}} \in \mathcal A^{\mathcal D_{\bm s}}_k$, Assumption (A1) implies that the policy continues to assign nonzero probability mass to $\bar{\bm{a}} $ 
% throughout the iterative learning process. 
% Therefore, at any iteration, there remains a nonzero probability that this action 
% $\bar{\bm{a}} $ is sampled.
% Whenever $\bm a^r_{k+1}=\bar{\bm{a}} $, it is suboptimal by construction.
% By (A2), with nonzero probability, the teacher provides corrective feedback, producing a desired set
% $\hat{\mathcal{A}}(\bar{\bm{a}}, \bm a^h_{k+1})$ that excludes at least  $\bar{\bm{a}}$. Therefore, with nonzero
% probability we have
% \begin{align}
% \bar{\bm{a}} \not\in \mathcal{A}^{\mathcal{D}_{\bm s}}_{k+1}
% =
% \mathcal{A}^{\mathcal{D}_{\bm s}}_{k} \cap \hat{\mathcal{A}}(\bar{\bm{a}}, \bm a^h_{k+1}),
% \end{align}
% Therefore, as $k \rightarrow \infty$, the probability that a suboptimal action  $\bar{\bm{a}}$ remains in $\mathcal{A}^{\mathcal{D}_{\bm s}}_{k} $ is zero, contradicting $\bar{\bm{a}} \in  \mathcal{A}^{\mathcal{D}_{\bm s}}_k$.
We show that $\lim_{k\rightarrow \infty} \mathcal{A}^{\mathcal{D}_{\bm s}}_k  \subset \mathcal{A}^*_{\bm s}$ by contradiction.
Assume for contradiction that there exists a suboptimal set
$
\mathcal{B} \subseteq \mathcal{A}^{\mathcal{D}_{\bm s}}_k \setminus \mathcal{A}^*_{\bm s}
$
that remains in the overall desired action set.
Since $\mathcal{B} \subseteq \mathcal A^{\mathcal D_{\bm s}}_k$, Assumption~(A1) implies that the policy can sample actions from this set with nonzero probability.
Therefore, at any iteration, there remains a nonzero probability that an action
$\bm a^r_{k+1} \in \mathcal{B}$ is sampled.
Whenever $\bm a^r_{k+1} \in \mathcal{B}$, it is suboptimal by construction.
By (A2), with nonzero probability, the teacher provides corrective feedback, producing a desired set
$\hat{\mathcal{A}}(\bm a^r_{k+1}, \bm a^h_{k+1})$ that excludes at least the corrected robot action $\bm a^r_{k+1}$.
Therefore, with nonzero probability we have
\begin{align}
\bm a^r_{k+1} \not\in \mathcal{A}^{\mathcal{D}_{\bm s}}_{k+1}
=
\mathcal{A}^{\mathcal{D}_{\bm s}}_{k} \cap \hat{\mathcal{A}}(\bm a^r_{k+1}, \bm a^h_{k+1}).
\end{align}
Thus, as $k \rightarrow \infty$, actions from suboptimal sets that remain in $\mathcal{A}^{\mathcal{D}_{\bm s}}_{k}$ can be sampled and corrected, allowing $\mathcal{A}^{\mathcal{D}_{\bm s}}_{k}$ to be refined accordingly.
This contradicts the assumption that such a suboptimal set remains in the limiting overall desired action set.
\end{proof}

\subsection{Connection between our KL Loss and InfoNCE Loss}
\label{appendix:connection_INFONCE}
In this section, we detail the connection between our KL Loss and the InfoNCE loss:
\begin{align*}
    &\mathrm{KL}\left(\pi^{\text{target}}(\bm a| \bm s ) \big\| \pi_{\theta}(\bm a | \bm s)\right)  \! \\ &= -\underset{\bm a \sim \pi^{\text{target}}(\bm a| \bm s )}{\mathbb{E}} \!\! \! \!  \log \pi_{\theta}(\bm a | \bm s) + \! \! \! \! \! \! \! \underset{\bm a \sim \pi^{\text{target}}(\bm a| \bm s )}{\mathbb{E}} \! \!\! \!  \! \! \! \!  \log \pi^{\text{target}}(\bm a| \bm s )  \\
    &= -\underset{\bm a \sim \pi^{\text{target}}(\bm a| \bm s )}{\mathbb{E}} \!\! \! \!  \log \pi_{\theta}(\bm a | \bm s) + c\\
    &\simeq \sum_{\bm a \in \mathbb A} -\pi^{\text{target}}(\bm a| \bm s ) \log \pi_{\theta}(\bm a | \bm s) + c \\
    &\overset{Eq. \eqref{eq:estiamtion_policy_EBM}}{\simeq} \sum_{\bm a \in \mathbb A} -\pi^{\text{target}}(\bm a| \bm s ) \log \frac{e^{-E_{\bm \theta}(\bm s, \bm a)}}{\sum_{\bm a' \in \mathbb A} e^{-E_{\theta}(\bm s, \bm a')}} + c\\
    &\overset{Eq. \eqref{eq:ibc_info_NCE}}{=} \sum_{\bm a \in \mathbb A} -\pi^{\text{target}}(\bm a| \bm s ) \ell_{\text{InfoNCE}}(\bm s, \bm a, \mathbb A \backslash  \{\bm a\}) +c,
\end{align*}
where $c$ denotes the constant that does not depend on $\bm \theta$.
By plugging the above equation into Eq.~\eqref{eq:KL_loss_general},
the KL loss becomes 
\begin{align*}
    \ell_{\!KL}\!(\bm \theta) \!
    % &= \sum_{\bm a } -p (\bm a|\bm a^{+},  \bm a^{-}) \log \frac{e^{-E_{\bm \theta}(\bm s, \bm a)}}{\displaystyle\sum_{\bm a' \in \mathcal{A}^{-}} e^{-E_{\theta}(\bm s, \bm a'_j)} + \sum_{\bm a' \notin \mathcal{A}^{+}} e^{-E_{\theta}(\bm s, \bm a'_j)}} \\
    % \simeq 
   \simeq \!\!\!\!
    \!\!\!\! \underset{(\bm{a}^{h}\!, \bm{a}^{r}\!, \bm{s})  \sim p_{\mathcal D}}{\mathbb{E}}\sum_{\bm a \in \mathbb A}  \pi^{\text{target}}(\bm a| \bm s ) \ell_{\text{InfoNCE}}(\bm s, \bm a, \mathbb A \! \backslash  \!\{\bm a\}) + \!c
\end{align*}

\subsection{Proof of the Simplified CLIC-Explicit Objective}
\label{apppendix:reduce_policy_improvemnt_inequality_Gaussian_assumption}

 We want to prove that, with the Gaussian distribution assumption and a polytope desired action set, Eq. \eqref{eq:policy_improvement_distribution} can be satisfied by satisfying Eq. \eqref{eq:policy_improvement}.
\begin{proof}
As the probability of sampling actions within $\hat{\mathcal{A}}{(\bm a^r, \bm a^h)}$ using $\pi_{\bm \theta}$ can be defined as  
\begin{align*}
    \pi_{\bm \theta}(\bm a \in \hat{\mathcal{A}}{(\bm a^r, \bm a^h)}| \bm{s}) 
    &=  \!\!\int_{\bm a \in \mathcal{A}}  \!\!\!\!   \!\!\!\!\pi_{\bm \theta}(\bm a | \bm s)  \text{Pr} [\bm a \in \hat {\mathcal{A}} {(\bm a^r, \bm a^h)}| \bm a , \bm s] d \bm a  \\
    &\overset{T\rightarrow0}{=} \int_{\bm a \in \mathcal{A}}  \pi_{\bm \theta}(\bm a | \bm s)   \cdot \mathbf{1}_{\bm a \in \hat {\mathcal{A}} {(\bm a^r, \bm a^h)}} d \bm a,
\end{align*}
Then Eq.~\eqref{eq:policy_improvement_distribution} in equivalent to 
\begin{align}
    \int_{\bm a \in \mathcal{A}}  \pi_{\bm \theta}(\bm a | \bm s)  \left( 2 \cdot \mathbf{1}_{\bm a \in \hat {\mathcal{A}} {(\bm a^r, \bm a^h)}} - 1\right) d \bm a\geq 0 
    \label{eq:proof_goal_simplifed_inequality}
\end{align}
% We can prove the following:
% % Despite this simplification, by constraining the policy with Gaussian distribution assumption to satisfy  Eq. \eqref{eq:policy_improvement}, it also satisfies Eq. \eqref{eq:policy_improvement_general_case}, which is 
% \begin{align}
% &\pi_{\bm \theta}( \bm a^{-}_i | \bm s) < \pi_{\bm \theta}( \bm a^{+}_i | \bm s) , i = 1, \dots, N_I,
% \\&\pi_{\bm \theta}(\bm a | \bm s) \sim\mathcal{N} (\mu_{\bm \theta}(\bm s), \bm \Sigma) \\
% \Rightarrow
%     &\int_{\bm a \in \mathcal{A}}  \pi_{\bm \theta}(\bm a | \bm s)  \left( 2 \cdot \mathbf{1}_{\bm a \in \hat {\mathcal{A}} {(\bm a^r, \bm a^h)}} - 1\right) d \bm a\geq 0 .
% \end{align}

% First step-> mean of gaussian is inside the desired action set
% Second step-> assume a very small variance, then
    As $\pi_{\bm \theta}(\bm a | \bm s)  \sim\mathcal{N} (\bm \mu_{\bm \theta}(\bm s), \bm \Sigma)$, we have 
\begin{align*}
\pi_{\bm \theta}(\bm a | \bm s)  \propto \exp \left( {-\frac{1}{2}( \bm a-\bm \mu_{\bm \theta}(\bm s))^{\mathsf{T}} {\bm \Sigma^{-1}}( \bm a-\bm \mu_{\bm \theta}(\bm s))} \right), 
\label{eq:gaussian}
\end{align*}
Without loss of generality, we set $\bm \Sigma = \sigma \bm I$ proportional to the identity matrix. 
Then, from $\pi_{\bm \theta}( \bm a^{-}_i | \bm s) < \pi_{\bm \theta}( \bm a^{+}_i | \bm s)$, we have 
\begin{equation*} 
\| \bm a^{+}_i-\bm  \mu_{\bm \theta}(\bm s) \|^2 < \| \bm a^{-}_i-\bm \mu_{\bm \theta}(\bm s) \|^2,  
\end{equation*}
which means that $\bm \mu_{\bm \theta}(\bm s) \in \hat {\mathcal{A}} {(\bm a^r, \bm a^h)} $. 
For $\alpha = 180^\circ$, $\hat {\mathcal{A}} {(\bm a^r, \bm a^h)}$ is a half-space and $2 \cdot \mathbf{1}_{\bm a \in \hat {\mathcal{A}} {(\bm a^r, \bm a^h)}} - 1 > 0$ always holds, thus Eq.~\eqref{eq:proof_goal_simplifed_inequality} always holds.
For $\alpha \in (0^\circ, 180^\circ)$, extra assumption needs to be made regarding the variance $\sigma$ to make Eq.~\eqref{eq:proof_goal_simplifed_inequality} holds.
For the extreme case, we can choose a $\sigma \rightarrow 0$ such that the policy outputs only the mean. 
In this case, Eq.~\eqref{eq:proof_goal_simplifed_inequality} holds.
% Now, we consider actions with the same distance to $\bm \mu^*(\bm s)$, which is $\bm a \in \mathcal{B}(p)= \{  \bm a \in \mathcal{A} | \|\bm a -\bm \mu^*(\bm s) \|^2 = p \}$.
% The number of actions on this ball that belong to the desired action set is larger than that that do not belong to the desired space. 
% Note that actions on the ball also have the same value of $\pi^*( \bm a | \bm s) $ by definition.
% Therefore, we have
% $\int_{\bm a \in \mathcal{B}(p) } \pi^*( \bm a | \bm s)\left( 2 \cdot \mathbf{1}_{\mathbb{D}(\bm a, \bm a^-) \geq \mathbb{D}(\bm a, \bm a^+)} - 1\right) d \bm a\geq 0$
% Thus, 
% \begin{align*}
%     \int_{\bm a \in \mathcal{A}}  \pi^*( \bm a | \bm s) \left( 2 \cdot \mathbf{1}_{\mathbb{D}(\bm a, \bm a^-) \geq \mathbb{D}(\bm a, \bm a^+)} - 1\right) d \bm a \\
%     = \int_p \int_{\bm a \in \mathcal{B}(p) } \pi^*( \bm a | \bm s) \left( 2 \cdot \mathbf{1}_{\mathbb{D}(\bm a, \bm a^-) \geq \mathbb{D}(\bm a, \bm a^+)} - 1\right) d \bm a \\
%     \geq 0
% \end{align*}
\end{proof}

% \begin{align*}
%     \ell_{KL}(\bm \theta) &\Leftrightarrow
%     \sum_{\bm a \in \mathbb A} -\pi^{\text{target}}(\bm a| \bm s ) \log \pi_{\theta}(\bm a | \bm s) \\
%     &\overset{Eq. \eqref{eq:estiamtion_policy_EBM}}{\simeq} \sum_{\bm a } -\pi^{\text{target}}(\bm a| \bm s ) \log \frac{e^{-E_{\bm \theta}(\bm s, \bm a)}}{\sum_{\bm a' \in \mathbb A} e^{-E_{\theta}(\bm s, \bm a')}}\\
%     &\overset{Eq. \eqref{eq:ibc_info_NCE}}{=} \sum_{\bm a \in \mathbb A} -\pi^{\text{target}}(\bm a| \bm s ) \ell_{\text{InfoNCE}}(\bm s, \bm a, \mathbb A \backslash  \bm a)
% \end{align*}

\subsection{Details of the Implementation of Baselines}
\label{appendix:baselines}
\subsubsection{D-COACH}
In \mbox{D-COACH}, the teacher shapes policies $\pi(\bm a|\bm s) \!\sim \! \mathcal{N} (\bm \mu_{\bm \theta}(\bm s), \bm \Sigma)$ by giving occasional relative corrective feedback $\bm h$. 
% Note that the robot action $\bm a$ is parameterized as $\bm \mu_{\bm \theta}(\bm s)$.%, specifically with the parameters $\bm \theta$ as they were at the moment the feedback was given.
% This signal $\bm h$ is used to create a corrected action $\bm a^{+}$ whose relative magnitude with respect to $\bm a$ is defined as the hyperparameter $e$, where  $e$ is a smaller value compared with the difference between $\bm a$ and the optimal action $\bm a^*$.
The human action $\bm a^{h} = \bm a^r  + e  \bm h$ is employed to update the policy parameters $\bm \theta$ in a behavior cloning manner:
\begin{equation}
    %\label{eq:dcoach_update}
    \ell^{\text{COACH}}_{\pi}(\bm s) = \min_{\bm \theta}{ \| \bm \mu_{\bm \theta}(\bm s) - \bm a^{h} \|^2}.
    \label{eq:dcoach}
\end{equation}
However, when learning from past experiences by using a replay buffer, as $\bm a^{h} \neq \bm a^*$, old feedback can lead the policy in the wrong direction. 
% More specifically, if $\bm \mu_{\bm \theta}(\bm s)$ is already closer than $\bm a^{h}$ to $\bm a^*$, i.e., $ \| \bm \mu_{\bm \theta}(\bm s) - \bm a^{*}  \|^2 <  \| \bm a^{h}- \bm a^{*} \|^2$, employing Eq. \ref{eq:dcoach_update} become harmful.
Consequently, to avoid this, \mbox{D-COACH} keeps a small data buffer $\mathcal{D}$, using only recent feedback for policy updates.
This approach leads to overfitting to recent trajectories and reduced learning efficiency.
% \subsubsection{BD-COACH}

To address this issue, batched D-COACH \cite{2021_BDCOACH}, \mbox{BD-COACH} for short, was proposed, which is used as the baseline in Section~\ref{sec:exp:simulation_relative_partial}. Here, besides learning a policy,
\mbox{BD-COACH} learns a human model $H_{\bm \phi}(\bm a, \bm s) = \bm h$. This model estimates the human's relative corrective feedback $\bm h$ given the robot's action $\bm a^r$ and state $\bm s$.
%All the history data $[\bm s, \bm a, \bm h]$ is saved to the data buffer $\mathcal{D}$.
Then, the human model is trained by minimizing the loss 
$
    \ell_H = \min_{\bm \phi}{ \| H_{\bm \phi}(\bm a^r, \bm s) - \bm h \|^2}
$,
 and the policy model is trained by minimizing the loss 
\begin{align*}
\ell_{\pi}^{\text{BD-COACH}}(\bm s) = \min_{\bm \theta}{ \| \bm \mu_{\bm \theta}(\bm s) - \hat {\bm a}^{h} \|^2},
\end{align*}
where $\hat{\bm a}^{h} = \bm a^r+ e \cdot  H_{\phi}(\bm a^r , \bm s)$. 
Therefore, since $\hat{\bm a}^{h}$ is estimated in relation to the current robot's policy, the correction data is no longer outdated. 
% However, this approach depends on an accurate estimation of $\bm h$, which is learned simultaneously to $\bm \mu_{\bm \theta}$. 
% Consequently, the learning process becomes slower because the policy can only improve after $H_{\bm \phi}$ generates coherent results. Additionally, precise hyperparameter tuning is essential for a stable learning process, given that two models are being optimized simultaneously.
\subsubsection{HG-DAgger}
\label{appendix:HGDAgger}
HG-DAgger is an intervention-based IIL algorithm that aggregates human demonstrations into a dataset and updates its behavior at the end of each episode by minimizing the distance between its current policy and the actions stored in the dataset.
In contrast, methods like CLIC and D-COACH update the policy continuously during training episodes, as shown in Algorithm \ref{alg:CLIC_algorithm} (line \ref{alg:line:update_policy_once}).
To make a fair comparison and make the IIL framework consistent across different methods, we introduce a slight modification to HG-DAgger,  allowing it to update its policy during each training episode, similar to CLIC. 
This modification enables HG-DAgger to converge faster than its original version. 

HG-DAgger assumes an explicit Gaussian policy and updates it using behavior cloning loss (see Eq.~\eqref{eq:dcoach}). For other offline BC baselines, such as DP and IBC, we adopt the HG-DAgger framework while replacing only the policy update step with their respective methods.

\subsubsection{Ambient DP}
ADP assumes access to corrupted training data $\bm a^h = \bm A_{corrupt} \bm a^*$, where $\bm A_{corrupt}$ is a known corruption matrix during training.
It introduces $\tilde{\bm A}$ to further corrupt the data $\tilde{\bm a}^h =  \tilde{\bm  A}\bm A_{corrupt}\bm a^*$. 
The diffusion model is then optimized to recover $\bm a^h$ from a noisy version of $\tilde{\bm a}^h $ at diffusion step $t$.
In simulated experiments, $\bm A_{corrupt}$ is unknown except in the partial feedback case. To instantiate ADP nonetheless, we give it access of $\bm a^*$ to construct this matrix, which is an extra privilege not afforded to the other methods.

% The loss
% \begin{equation}
%     \ell_{\pi_\theta}(\bm{s}) = 
%     - \| A\big( \tilde{A} \pi_{\bm \theta} (\bm s)  - \bm a^h \big) \|
%     \label{eq:ADP_Loss}
% \end{equation}

\subsection{The Setup of Experiments}
% \label{appendix:real_robot_experiments_task details}

In all the experiments, we used state-based observation as input for the policy across all methods. 
The summary of the tasks in both the simulation and the real world is reported in Table \ref{tab:appendix:setup_exp}.
% The term `multi-modal' denotes whether the task has multiple solutions inside the dataset. 

\begin{table}[]
\footnotesize
\caption{Tasks summary }
\label{tab:appendix:setup_exp}
\begin{center}
\begin{tabular}{cccccc}
\Xhline{0.75pt}
\multicolumn{1}{c}{\multirow{2}{*}{{Tasks}}} & \multicolumn{1}{c}{\multirow{2}{*}{{\begin{tabular}[c]{@{}c@{}} State\\ dim \end{tabular}}}} & \multicolumn{1}{c}{\multirow{2}{*}{{\begin{tabular}[c]{@{}c@{}} Action\\ dim\end{tabular}}}} & \multicolumn{1}{c}{\multirow{2}{*}{\begin{tabular}[c]{@{}c@{}}Multi-\\ modal\end{tabular}}} & \multicolumn{1}{c}{\multirow{2}{*}{\begin{tabular}[c]{@{}c@{}} Contact\\ Rich\end{tabular}}} & \multicolumn{1}{c}{\multirow{2}{*}{\begin{tabular}[c]{@{}c@{}} High\\ freq\end{tabular}}}  \\
\multicolumn{1}{c}{}                       & \multicolumn{1}{c}{}                           & \multicolumn{1}{c}{}                           & \multicolumn{1}{c}{}  & \multicolumn{1}{c}{}  & \multicolumn{1}{c}{} \\
\hline
Push-T         &   24         & 2         &        \ding{51}  & \ding{51}  &  \ding{55}  \\
Square        &     48      & 7         &         \ding{51}    & \ding{55} & \ding{55} \\
Pick-Can       &     40      & 7         &          \ding{55}  & \ding{55} & \ding{55} \\
TwoArm-Lift    &     48     & 14        &         \ding{55}    & \ding{55}  & \ding{55}\\
Ball-catching &      12     & 3         &         \ding{55}   & \ding{55} &  \ding{51} \\
Water-pouring &      7     & 6         &          \ding{55}  & \ding{55} & \ding{55} \\
Insert-T       &    52       & 2         & \ding{51} &  \ding{51}& \ding{55}\\
\Xhline{0.75pt}           
\end{tabular}
\end{center}
\end{table}

\subsubsection{Simulated tasks}
\label{appendix:simulated_experiments_task details}
\label{}
The task descriptions are as follows:
\begin{enumerate}[label=\roman*), leftmargin=0.6cm]
  \item \textbf{Push-T:} This task, introduced by \cite{2023_diffusionpolicy}, involves the robot pushing a T-shape object to a fixed target using its circular end effector. 
 \item \textbf{Square:} The robot must place a square-shaped nut onto a fixed square peg. 
   \item \textbf{Pick-Can:} The objective is to pick up a can object and transport it to a fixed target bin. 
    \item \textbf{TwoArm-Lift:} This task involves two robots working together to lift a shared object. 
\end{enumerate}
For each task, the object's position is randomly initialized at the beginning of each episode.
\subsubsection{Real-world tasks}
\label{appendix:real_robot_experiments_task details}
The learned policy is evaluated every 5 episodes for the water-pouring task, every 10 episodes for the ball-catching task, and every 20 episodes for the Insert-T task. The details of each real-world task are detailed as follows:
\begin{enumerate}[label=\roman*), leftmargin=0.6cm]
  \item \textbf{Ball catching:} This task involves the robot catching a ball that's swinging on a string attached to a fixed point. 
  % The challenge lies in catching the ball while it is still moving. 
  The robot's end effector operates within the same plane where the ball swings, maintaining a fixed orientation. The {action space} consists of the robot's end effector linear velocity in the specified plane and a one-dimensional continuous actuation command for controlling the robot's gripper, where 0 represents fully closed and 1 fully open. The {state space} includes the end-effector's relative position and velocity with respect to the ball, the angle between the gravity vector and the ball's string along with its corresponding angular velocity, and the poses of both the ball and the fixed point.
  The poses of the ball and the fixed point are measured using an OptiTrack system.
  % The CLIC-Relative is utilized in this experiment.

  \item \textbf{Water pouring:} The robot controls the pose of a bottle to precisely pour liquid (represented with marbles) into a small bowl. The {action space} is 6D, consisting of the robot's end-effector linear and angular velocities. The {state space} is defined by the robot's end-effector pose, which consists of its Cartesian position and its orientation, represented with a unit quaternion. The initial pose of the robot is randomized at the start of each episode within certain position and orientation limits to ensure safety.
  % Both the CLIC-Relative and CLIC-Absolute are utilized in this experiment.

  \item \textbf{Insert-T:} The robot must insert a T-shape object into a U-shape object by pushing them. 
  % It doesn't matter which object it manipulates first. 
  The {action space} is defined as the linear velocity of the end-effector in a 2D plane over the table.
  The robot's end-effector orientation and z-axis position (the one aligned with the table's normal vector) are fixed throughout the task.
  The positions of the objects are measured by an OptiTrack system.
  The state space consists of the positions of 10 key points on a T-shape object, the positions of 10 key points on a U-shape object, and the position and velocity of the end-effector.
\end{enumerate}

\subsubsection{Time duration}
\label{appendix:time_duration}
The total time duration of the real-world experiments within the IIL framework is reported in Table \ref{tab:appendix:Time_duration_real_world}, excluding the time spent resetting the robot or performing evaluations.

\begin{table}[h]
\footnotesize
\caption{Total time duration of real-world experiments}
\label{tab:appendix:Time_duration_real_world}
\begin{tabular}{cccc}
\Xhline{0.75pt}
                    & Ball-catching & Water-pouring & Insert-T \\ \hline
Time duration (minutes) & 74        & 40            & 140            \\
\Xhline{0.75pt}
\end{tabular}
\end{table}

% In Fig. \ref{fig:real_exp_figs_combined_all}, the {total time} is used as the x-axis, which refers to the entire experiment duration, excluding resetting time. 

\subsection{Time Efficiency Comparison}
\begin{table}[h]
% \small\sf\centering
% \setlength\tabcolsep{ 4 pt}
\footnotesize
\caption{Time efficiency comparison per step}
\label{tab:appendix:Time_efficience_comparison}
\begin{tabular}{cccc}
\Xhline{0.75pt}
                    & CLIC-Half & CLIC-Circular & CLIC-Explicit \\ \hline
Inference time (ms) & 28.61        & 28.52            & 1.13             \\
Training time (ms)  &    201.32        &    176.64          &      12.66           \\ \Xhline{0.75pt}
\end{tabular}
\end{table}
\label{appendix:time_efficiency_comparision}
For all CLIC methods with a batch size of 32, the inference and training times per step on the Square task were recorded and averaged.
As reported in Table \ref{tab:appendix:Time_efficience_comparison}, although implicit models have better encoding capability, they require more time for both training and inference compared to the explicit model used in CLIC-Explicit.
This presents a trade-off when selecting an algorithm for practical use.  For uni-modal tasks, CLIC-Explicit can be used for its time efficiency. For multi-modal tasks, CLIC-Half and CLIC-Circular should be used instead. 
Additionally, Table \ref{tab:appendix:Time_efficience_comparison} shows that CLIC-Half has a slightly longer training time per step than CLIC-Circular, as it involves an additional step of sampling implicit negative actions.

\subsection{Implementation Details}
% \textcolor{red}{Update this information}
\subsubsection{Network structure}
For implicit policies, CLIC, IBC, and PVP use the same neural network structure for the EBM. 
The neural network consists of five fully connected layers with [512, 512, 512, 256, 1] units, respectively.
The ReLU activation function is applied between all layers except for the last layer, which has no activation function.
The input is the concatenation of the state and action vectors, and the output is a scalar. 
For DP, the neural network follows a similar structure, except that the last layer has a number of units equal to the dimensionality of the action space.

For the explicit policy $\pi_{\bm \theta}(\bm s,\bm a) \sim \mathcal{N} (\bm \mu_{\bm \theta}(\bm s), \bm \Sigma)$, the mean $\bm \mu_{\bm \theta}(\bm s)$ is parameterized by a neural network consisting of five fully connected layers. The layer units are the same as those of the DP's model; except that the final layer applies a sigmoid activation, followed by scaling and shifting to produce values between -1 and 1. This neural network takes a state vector as input and maps it to an action.
This action is obtained via $\bm a_t = \bm \mu_{\bm \theta}(\bm s_t)$.
The covariance matrix $\bm \Sigma$ can be chosen to control the variability of actions sampled from the Gaussian distribution. In our implementation, it is ignored entirely, in which case actions are always taken as $\bm \mu_{\bm \theta}(\bm s_t)$.
% The values of other hyperparameters for different methods are shown in Table \ref{table:appendx_Hyperparameters}.

\subsubsection{Gradient Penalty}
As described in Appendix B.3.1 of IBC \cite{2022_implicit_BC}, we incorporate a gradient penalty loss to improve the training stability of CLIC. This penalty is computed using only the action samples from the final MCMC step. In practice, we found that the gradient penalty used in IBC is more effective than the L2 norm penalty.

\subsubsection{Hyperparameters for Training}
At each training episode, the network parameters will be updated with an update frequency of $b=5$ for all methods, as shown in line \ref{alg:line:update_feq_b} of Algorithm \ref{alg:CLIC_algorithm}.
The batch size is set to 32 for accurate, relative, and partial feedback, and 10 for noisy feedback. 
The learning rate $\eta$ is 0.0003. 
The optimizer for the neural network is Adam with $\beta_1 = 0.1, \beta_2 = 0.999$ and $\epsilon=1e-7$.
The number of training steps at the end of each episode, $N_{training}$, is set to 500 for all methods except DP, for which it is set to 1000.

For CLIC-Circular, $\varepsilon$ is set to $0.5$ for accurate and relative feedback, and set to $1.0$ for partial and noisy feedback, and the temperature $T$ is set to 0.05 for all tasks.
For CLIC-Half, the hyperparameters $\alpha$ and $\varepsilon$ are set to $\alpha=30^\circ$ and $\varepsilon = 0.3$ for accurate and relative feedback; and set to $100^\circ$ and $\varepsilon = 0.1$  for partial and noisy feedback. 
For the Insert-T task, $\alpha=30^\circ$ and $\varepsilon = 0.1$.
The number of contrastive action pairs, $N_I$, is 128. 
The magnitude hyperparameter is $e = 0.2$.
The number of sampled actions from EBM is $N_{\text{a}} = 512$.
$N_{\text{MCMC}} = 25$ during training and $N_{\text{MCMC}} = 50$ during inference.  

For IBC, $N_{\text{neg}}=512$, $N_{\text{MCMC}} = 25$ during training and $N_{\text{MCMC}} = 50$ during inference.  

% \begin{table}[]
% \caption{Hyperparameters for the algorithms}
% \label{table:appendx_Hyperparameters}
% \begin{center}
% \begin{tabular}{l|l|c}
% \hline
% Algorithm                           & \multicolumn{1}{c|}{Hyperparameters}        & Value  \\ \hline
% \multirow{3}{*}{CLIC-Correction}    & Magnitude parameter $e$  & 0.02   \\& Learning rate & 0.001  \\
%                                     & End-of-episode training iterations & 1000   \\ \hline
% \multirow{2}{*}{CLIC-Demonstration} & Learning rate                               & 0.0003 \\
%                                     & End-of-episode training iterations & 1000   \\ \hline
% \multirow{3}{*}{BD-COACH}           & Magnitude parameter $e$                       & 0.02   \\
%                                     & Learning rate                               & 0.0003 \\
%                                     & End-of-episode training iterations & 1000   \\ \hline
% \multirow{2}{*}{Modified HG-DAgger} & Learning rate                               & 0.0003 \\
%                                     & End-of-episode training iterations & 1000   \\ \hline
% \multirow{2}{*}{Original HG-DAgger} & Learning rate                               & 0.0003 \\
%                                     & End-of-episode training iterations & 2000   \\ \hline
% \end{tabular}
% \end{center}
% \end{table}

% % \bibliographystyle{plainnat}
% \bibliographystyle{unsrt} % Ensures citations appear in order
% \bibliographystyle{jabbrv_ieeetr}

\bibliographystyle{SageH}
\bibliography{main}

% \bibliographystyle{./IEEEtranBST/IEEEtran}
% \bibliography{./IEEEtranBST/IEEEabrv, references}

% \input{rebuttal}

\end{document}